\def\swordtitleruleplacement{1}
\def\swordtitlebrandvariant{2}
\def\swordtitlefontvariant{1}
\documentclass[oneside, a4paper, onecolumn, 10pt]{article}

\usepackage{booktabs}
\usepackage{array}
\usepackage{amssymb}
\usepackage{tabularx}

\title{Move by Move: Measuring and Steering\\How LLMs Conduct Psychotherapy}

\author{%
{\normalsize
\textbf{Afonso Baldo}\textsuperscript{1,\,3,\,*},\enspace \textbf{Hugo Pitorro}\textsuperscript{1,\,3,\,*},\enspace \textbf{Areti Vassilopoulos}\textsuperscript{1,\,2},\enspace \textbf{Anabela C. Areias}\textsuperscript{1},\enspace \textbf{Maya D'Eon}\textsuperscript{1}}\\[0.8mm]
{\normalsize
\textbf{Fabíola Costa}\textsuperscript{1},\enspace \textbf{Ricardo Rei}\textsuperscript{1},\enspace \textbf{Nuno M. Guerreiro}\textsuperscript{1}}\\[1.1mm]
{\small \textsuperscript{1}Sword Health \qquad \textsuperscript{2}Yale University \qquad \textsuperscript{3}Instituto Superior Técnico}\\[0.45mm]
{\small \href{mailto:ai.research@sword.com}{\textcolor{black}{{\footnotesize\faEnvelope}\hspace{0.45em}ai.research@sword.com}}}
}
\date{}
\usepackage[dvipsnames]{xcolor}

\definecolor{linkblue}{RGB}{0,82,155}
\definecolor{citegreen}{RGB}{0,110,70}
\definecolor{urlviolet}{RGB}{100,40,130}

\definecolor{swordcoral}{RGB}{194,64,47}
\definecolor{swordblue}{RGB}{58,84,176}
\definecolor{swordviolet}{RGB}{105,74,158}

\definecolor{lstbg}{RGB}{250,246,239}
\definecolor{lstframe}{RGB}{223,220,213}
\definecolor{lstcmt}{RGB}{0,110,70}
\definecolor{lstrole}{RGB}{0,82,155}
\definecolor{lststr}{RGB}{60,60,60}
\definecolor{lstkw}{RGB}{140,30,120}
\definecolor{hlcol}{RGB}{250,246,239}

\definecolor{tokfg}{HTML}{374151}
\definecolor{tokbg}{HTML}{F3F4F6}

\definecolor{bordo}{RGB}{128,0,32}

\definecolor{panelPrompt}{HTML}{FCF3EA}
\definecolor{panelReason}{HTML}{F4F4F4}
\definecolor{panelAnswer}{HTML}{E8E8E4}
\definecolor{panelBorder}{HTML}{B9B9B9}
\definecolor{panelInk}{HTML}{27272A}
\definecolor{panelInkStrong}{HTML}{141414}
\definecolor{panelAccent}{HTML}{C43C45}
\definecolor{panelAccentBG}{HTML}{F3DEDE}

\usepackage{fullpage}
\usepackage[left=2.5cm,top=1.7cm,bottom=2.2cm,right=2.5cm]{geometry}
\usepackage[utf8]{inputenc}
\usepackage[T2A,T1]{fontenc}
\usepackage{iftex}
\ifPDFTeX
  \usepackage{CJKutf8}
  \DeclareUnicodeCharacter{211D}{\ensuremath{\mathbb{R}}}
  \DeclareUnicodeCharacter{2202}{\ensuremath{\partial}}
  \DeclareUnicodeCharacter{2212}{\ensuremath{-}}
\fi

\usepackage{amsmath}
\usepackage{amssymb}
\usepackage{bbm}
\usepackage{bm}

\usepackage{booktabs}
\usepackage{multirow}
\usepackage{colortbl}

\usepackage{graphicx}
\usepackage{wrapfig}
\usepackage{needspace}
\usepackage{float}
\usepackage[most]{tcolorbox}

\usepackage{pifont}
\usepackage{fontawesome5}
\usepackage{soul}
\usepackage{eurosym}
\usepackage{xspace}
\usepackage{seqsplit}
\usepackage{comment}
\usepackage{enumitem}

\usepackage{tgpagella}
\usepackage{roboto}
\usepackage{nopageno}

\usepackage[authoryear,round]{natbib}

\usepackage[
  colorlinks=true,
  linkcolor=swordblue,
  citecolor=swordblue,
  urlcolor=swordblue,
  unicode
]{hyperref}
\usepackage{xurl}
\usepackage{footnotehyper}
\makesavenoteenv{figure}

\usepackage{etoc}
\usepackage[nameinlink,capitalise,noabbrev]{cleveref}

\usepackage{siunitx}

\usepackage{tikz}
\usetikzlibrary{positioning, arrows.meta, calc, decorations.pathreplacing, fit, backgrounds, patterns}
\usepackage{pgfplots}
\usepgfplotslibrary{groupplots}
\pgfplotsset{compat=1.16}
\usepackage{fancyvrb}
\usepackage{fvextra}
\fvset{breaklines=true,breakanywhere=true}
\usepackage{algorithm}
\usepackage{algpseudocode}
\usepackage{listings}

\usepackage{iftex}
\ifPDFTeX
  \newcommand{\bodyfont}{\fontfamily{XCharter-TLF}\selectfont}
  \newcommand{\labelfont}{\fontfamily{phv}\selectfont}
  \newcommand{\paperromanfont}{\normalfont}
\else
  \usepackage{fontspec}
  \newfontfamily\bodyfont{Spectral}[
    Path=fonts/, Extension=.ttf,
    UprightFont=*-Regular, ItalicFont=*-Italic,
    BoldFont=*-Bold, BoldItalicFont=*-BoldItalic]
  \newfontfamily\labelfont{LibreFranklin}[
    Path=fonts/, Extension=.ttf,
    UprightFont=*-SemiBold, BoldFont=*-Bold]
  \newfontfamily\paperromanfont{lmroman10-regular.otf}[Ligatures=TeX]
\fi

\ifPDFTeX
  
\else
  \newfontfamily\intlfont{NotoSansMonoCJKsc-Regular.otf}[Path=fonts/, Scale=MatchLowercase]
  
  \IfFontExistsTF{Noto Sans Devanagari}
    {\newfontfamily\devfont{Noto Sans Devanagari}[Scale=MatchLowercase]}{\let\devfont\intlfont}
  \IfFontExistsTF{Noto Sans Georgian}
    {\newfontfamily\geofont{Noto Sans Georgian}[Scale=MatchLowercase]}{\let\geofont\intlfont}

\fi
\ifPDFTeX \fi

\lstdefinestyle{trace}{
  basicstyle=\ttfamily\scriptsize,
  backgroundcolor=\color{lstbg},
  frame=single, rulecolor=\color{lstframe}, framerule=0.6pt,
  breaklines=true, breakindent=0pt, postbreak=\mbox{\textcolor{lstframe}{$\hookrightarrow$}\space},
  columns=fullflexible, keepspaces=true,
  commentstyle=\color{lstcmt}\itshape,
  keywordstyle=\color{lstrole}\bfseries,
  emphstyle=\color{lstrole}\bfseries,
  morekeywords={messages,user,assistant,system,thinking,text,signature,primer,decoder,reconciler},
  morecomment=[l]{\#},
  aboveskip=2pt, belowskip=2pt,
  xleftmargin=5pt, xrightmargin=3pt, framexleftmargin=5pt,
}

\lstdefinestyle{pyclean}{
  language=Python,
  basicstyle=\ttfamily\scriptsize,
  backgroundcolor=\color{lstbg},
  frame=single, rulecolor=\color{lstframe}, framerule=0.6pt,
  breaklines=true, breakindent=0pt,
  postbreak=\mbox{\textcolor{lstframe}{$\hookrightarrow$}\space},
  columns=fullflexible, keepspaces=true,
  showstringspaces=false,
  commentstyle=\color{lstcmt}\itshape,
  keywordstyle=\color{lstkw}\bfseries,
  stringstyle=\color{lstrole},
  emphstyle=\color{lstrole}\bfseries,
  emph={client,messages,create,model,max_tokens,thinking,content,role,signature},
  aboveskip=6pt, belowskip=6pt,
  xleftmargin=5pt, xrightmargin=3pt, framexleftmargin=5pt,
}

\lstdefinestyle{extracttemplate}{
  basicstyle=\ttfamily\footnotesize,
  frame=none,
  backgroundcolor={},
  breaklines=true,
  breakindent=12pt,
  columns=fullflexible,
  keepspaces=true,
  showstringspaces=false,
  alsoletter={@},
  morekeywords={@thought},
  keywordstyle=\color{panelAccent}\bfseries,
  commentstyle=\color{panelAccent}\itshape,
  morecomment=[l]{//},
  xleftmargin=0pt,
  xrightmargin=0pt,
  aboveskip=0pt,
  belowskip=0pt,
}

\lstdefinestyle{extractcall}{
  language=Python,
  basicstyle=\ttfamily\scriptsize,
  frame=none,
  backgroundcolor={},
  breaklines=true,
  breakindent=12pt,
  columns=fullflexible,
  keepspaces=true,
  showstringspaces=false,
  commentstyle=\color{lstcmt}\itshape,
  keywordstyle=\color{lstkw}\bfseries,
  stringstyle=\color{lstrole},
  emphstyle=\color{panelAccent}\bfseries,
  emph={client,messages,create,generate_content,responses,models,Content,Part},
  xleftmargin=0pt,
  xrightmargin=0pt,
  aboveskip=0pt,
  belowskip=0pt,
}

\lstdefinestyle{judgeprompt}{
  basicstyle=\ttfamily\scriptsize,
  frame=none,
  backgroundcolor={},
  breaklines=true,
  breakindent=12pt,
  columns=fullflexible,
  keepspaces=true,
  showstringspaces=false,
  alsoletter={_-},
  emph={META,SUB,REAL,NOT-ARTIFACT,RESOLUTION,Reasoning trace},
  emphstyle=\color{panelAccent}\bfseries,
  xleftmargin=0pt,
  xrightmargin=0pt,
  aboveskip=0pt,
  belowskip=0pt,
}
\usepackage{xcolor}
\definecolor{accent}{HTML}{C43C45}

\usepackage[font=small,labelfont=bf,skip=4pt,format=plain]{caption}
\usepackage{fancyhdr}
\crefname{appendix}{Appendix}{Appendices}
\Crefname{appendix}{Appendix}{Appendices}
\crefname{app}{Appendix}{Appendices}
\Crefname{app}{Appendix}{Appendices}

\setitemize{itemsep=4pt,topsep=-3pt,parsep=0pt,partopsep=0pt,leftmargin=18pt}

\makeatletter
\renewcommand{\paragraph}{%
  \@startsection{paragraph}{4}%
  {\z@}{0.5ex \@plus 1ex \@minus .2ex}{-1em}%
  {\normalfont\normalsize\bfseries}%
}
\makeatother

\makeatletter
\let\origsection\section
\renewcommand\section{\@ifstar{\starsection}{\nostarsection}}
\newcommand\nostarsection[1]
{\sectionprelude\origsection{#1}\sectionpostlude}
\newcommand\starsection[1]
{\sectionprelude\origsection*{#1}\sectionpostlude}
\newcommand\sectionprelude{\vspace{-3mm}}
\newcommand\sectionpostlude{\vspace{-2.5mm}}
\makeatother

\makeatletter
\let\origsubsection\subsection
\renewcommand\subsection{\@ifstar{\starsubsection}{\nostarsubsection}}
\newcommand\nostarsubsection[1]
{\subsectionprelude\origsubsection{#1}\subsectionpostlude}
\newcommand\starsubsection[1]
{\subsectionprelude\origsubsection*{#1}\subsectionpostlude}
\newcommand\subsectionprelude{\vspace{-2mm}}
\newcommand\subsectionpostlude{\vspace{-2.5mm}}
\makeatother

\graphicspath{ {figures/} }

\sethlcolor{tokbg}

\tcbset{
  decodedpanel/.style={
    enhanced jigsaw,
    breakable,
    boxrule=0pt,
    arc=3.2pt,
    boxsep=0pt,
    left=6pt,
    right=6pt,
    top=5pt,
    bottom=6pt,
    before skip=6pt,
    after skip=6pt,
    coltext=panelInk,
  },
}

\newtcolorbox{decodedpromptpanel}{
  decodedpanel,
  colback=panelPrompt,
  fontupper=\sffamily\small\linespread{1.15}\selectfont,
}
\newtcolorbox{decodedreasoningpanel}{
  decodedpanel,
  colback=panelReason,
  fontupper=\ttfamily\scriptsize\linespread{1.15}\selectfont,
  before upper={\raggedright\sloppy},
}
\newtcolorbox{decodedanswerpanel}{
  decodedpanel,
  colback=panelAnswer,
  fontupper=\sffamily\small\linespread{1.15}\selectfont,
}

\newtcblisting{extractiontemplate}{
  decodedpanel,
  colback=panelPrompt,
  listing only,
  listing options={style=extracttemplate},
}
\newtcblisting{extractioncall}{
  decodedpanel,
  colback=panelReason,
  listing only,
  listing options={style=extractcall},
}
\newtcblisting{judgepromptpanel}{
  decodedpanel,
  colback=panelPrompt,
  listing only,
  listing options={style=judgeprompt},
}

\newcommand{\mvu}{\_\allowbreak}

\newenvironment{ontolist}
  {\begin{list}{$\bullet$}{%
     \setlength{\leftmargin}{1.1em}\setlength{\labelwidth}{0.7em}%
     \setlength{\labelsep}{0.4em}\setlength{\itemsep}{1pt}%
     \setlength{\parsep}{0pt}\setlength{\topsep}{2pt}%
     \setlength{\partopsep}{0pt}}}
  {\end{list}}

\newenvironment{ontoenum}
  {\begin{list}{}{%
     \setlength{\leftmargin}{1.6em}\setlength{\labelwidth}{1.2em}%
     \setlength{\labelsep}{0.4em}\setlength{\itemsep}{1pt}%
     \setlength{\parsep}{0pt}\setlength{\topsep}{2pt}%
     \setlength{\partopsep}{0pt}}}
  {\end{list}}

\usepackage{arydshln}
\makeatletter
\def\adl@drawiv#1#2#3{%
        \hskip.5\tabcolsep
        \xleaders#3{#2.5\@tempdimb #1{1}#2.5\@tempdimb}%
                #2\z@ plus1fil minus1fil\relax
        \hskip.5\tabcolsep}
\newcommand{\cdashlinelr}[1]{%
  \noalign{\vskip 2pt
           \global\let\@dashdrawstore\adl@draw
           \global\let\adl@draw\adl@drawiv}
  \cdashline{#1}[.4pt/2pt]
  \noalign{\global\let\adl@draw\@dashdrawstore
           \vskip 2pt}}
\makeatother

\providecommand{\swordbodyfontvariant}{0}
\ifnum\swordbodyfontvariant=1\relax
  \let\paperromanfont\rmfamily
\fi

\providecommand{\swordtitlefontvariant}{0}
\ifnum\swordtitlefontvariant=1\relax
  \ifPDFTeX
    \newcommand{\swordtitleface}{\fontfamily{XCharter-TLF}\selectfont}
  \else
    \newfontfamily\swordtitleface{XCharter}[
      UprightFont    = XCharter-Roman,
      ItalicFont     = XCharter-Italic,
      BoldFont       = XCharter-Bold,
      BoldItalicFont = XCharter-BoldItalic
    ]
  \fi
\else
  \newcommand{\swordtitleface}{}
\fi

\providecommand{\swordtitleruleplacement}{1}
\providecommand{\swordtitlebrandvariant}{0}
\newcommand{\swordtitlelogo}{%
  \ifnum\swordtitlebrandvariant=1\relax
    \includegraphics[width=2.25cm]{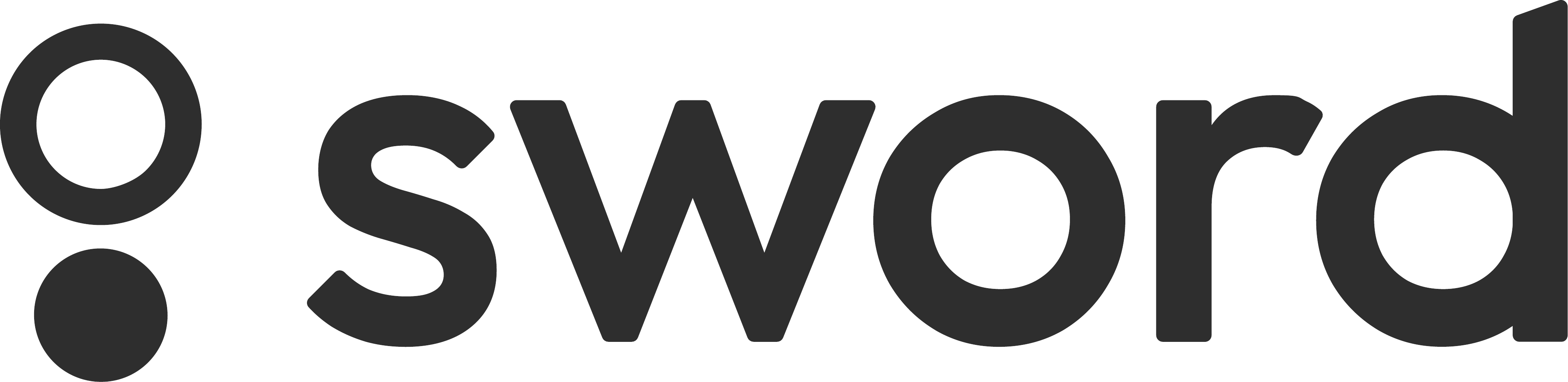}%
  \else
    \includegraphics[width=2.25cm]{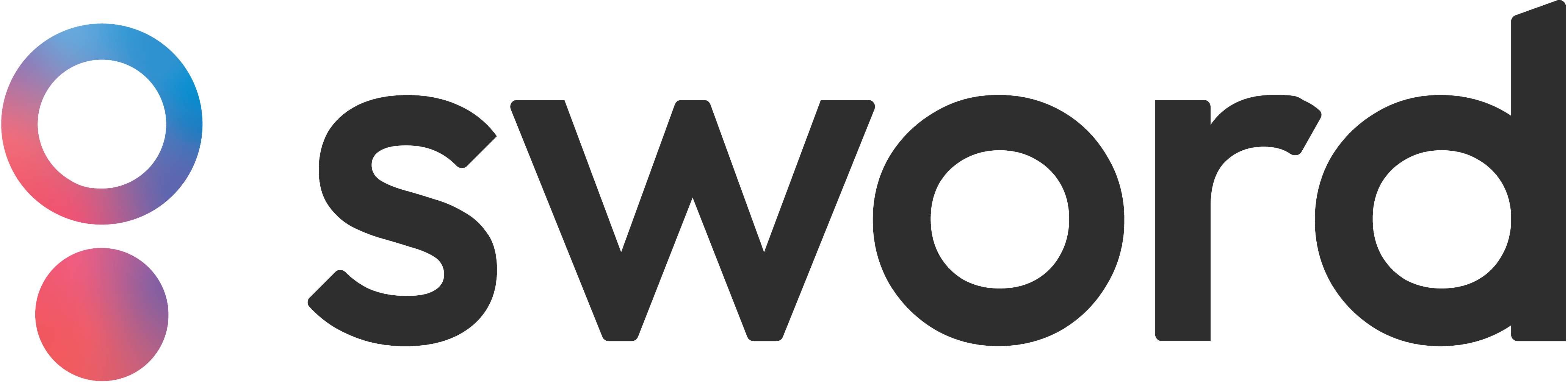}%
  \fi
}
\newcommand{\swordgradientrule}{%
  \begingroup
    \setlength{\parskip}{0pt}%
    \noindent
    \begin{tikzpicture}[baseline=0pt]
      \ifnum\swordtitlebrandvariant=0\relax
        \shade[
          shading=axis,
          left color=swordcoral,
          middle color=swordviolet,
          right color=swordblue,
          shading angle=90,
          rounded corners=0.7pt
        ] (0,0) rectangle (\linewidth,1.4pt);
      \else
        \fill[black,rounded corners=0.7pt]
          (0,0) rectangle (\linewidth,1.4pt);
      \fi
    \end{tikzpicture}\par
  \endgroup
}
\makeatletter
\renewcommand{\@maketitle}{%
  \newpage
  \null
  \vspace{-0.9cm}%
  \begingroup
    \setlength{\parindent}{0pt}%
    \raggedright
    \ifnum\swordtitleruleplacement=0\relax
      \swordgradientrule
      \vspace{0.22cm}%
    \fi
    \swordtitlelogo\par
    \ifnum\swordtitleruleplacement=1\relax
      \vspace{-0.08cm}%
      \swordgradientrule
      \vspace{0.22cm}%
    \else
      \vspace{0.25cm}%
    \fi
    {\swordtitleface\LARGE\bfseries\@title\par}
    \vspace{0.22cm}%
    {\@author\par}
  \endgroup
}
\makeatother

\begin{document}
\maketitle
\begingroup
  \renewcommand{\thefootnote}{}
  \footnotetext{\textsuperscript{*}Equal contribution.}
\endgroup
\vspace{0.22cm}
{\labelfont\fontsize{10.2}{12}\selectfont Abstract\par}
\vspace{0.04cm}
{\small As the demand for mental health services outpaces clinical workforce capacity, users are increasingly turning to large language models (LLMs) for accessible and private support.
Although recent evaluations suggest LLM care can reduce clinical symptoms, how models actually conduct therapeutic interactions compared to trained human clinicians remains largely unknown.
Drawing on established clinical psychology frameworks that conceptualize therapy as a sequential decision-making process, we can systematically analyze the conversational interventions used in AI-driven sessions.
To achieve this, we introduce an original, clinically validated ontology of ten distinct therapeutic ``moves'' to descriptively categorize strategies LLMs favor or neglect.
Applying this ontology at scale reveals that models drastically over-use probing, disregard common CBT-favored techniques and are strongly context-anchored. Moreover, we propose exposing these categories as callable tools, bringing models significantly closer to the human clinical baseline.
\par}
\vspace{0.35cm}

\section{Introduction}
Mental-health disorders have risen sharply worldwide, with the global burden of anxiety and depressive disorders roughly doubling since the 1990s \citep{gbd2023mentalDisorderCollaborators2026updated}. At the same time, mental-health services and the clinical workforce have not kept up with demand \citep{10.1001/jamapsychiatry.2023.1253}.
Driven by lower cost, immediate availability, and privacy preferences, users increasingly turn to large language models (LLMs) for emotional support, counseling-adjacent chat, and advice on interpersonal relationships \citep{anthropic2025affective, rousmaniere2026large}.

Despite their known issues (such as sycophancy \citep{fanous2025syceval}), stemming from their training as harmless and helpful assistants \citep{ouyang2022training, bai2022traininghelpfulharmlessassistant}, recent evaluations report LLM care as competitive to humans, with controlled trials exhibiting general symptom reduction \citep{rollwage2026cognitive, heinz2024evaluating}.
Yet these outcome-level results leave a more basic question unanswered: we know little about \emph{how} LLMs actually conduct a psychotherapy interaction, which interventions they favor, which they neglect, and how their conduct of a session compares with that of a trained clinician.

Clinical psychology offers a natural lens for this question. 
Psychotherapy can be conceptualized as a sequential clinical decision-making process where the therapist iteratively selects among alternative interventions as the conversation unfolds \citep{murphy2003, goldfried1980}. 
Established coding frameworks, such as the Motivational Interviewing Skill Code \citep{miller2003manual} and Hill's Helping Skills \citep{helpingskillstrainning_2014}, operationalize this view by mapping each therapist utterance onto a discrete set of strategies, or \emph{skills} (e.g., open questions, reflections, interpretations, challenges). 
These frameworks underpin therapist training and supervision, letting experts systematically analyze which skills trainees deploy and provide precise, objective feedback \citep{hill2007training, hill2008helping, rosengren2017building}.

Building on this insight, we introduce ``moves'': an original, clinically developed ontology comprising ten compact, function-based therapeutic categories.
These categories are grounded in the MULTI-60 inventory \citep{mccarthy2009multi}, drawing on clinical skills used across various treatment modalities, and were validated through an annotation campaign with five licensed clinical psychologists.
Through this ontology, we are able to assess how an LLM conducts a psychotherapy interaction: which moves it relies on, which it rarely uses, and where its conduct departs from that of a human clinician.
Like \citet{transluce2026mentalhealth}, we take a \textit{descriptive} rather than \textit{normative} approach: we surface these differences and track them across models and settings, describing how LLMs conduct such interactions without attempting to judge how \textit{good} or \textit{bad} that conduct is.\footnote{What counts as a clinically appropriate response is often not clear-cut, and even expert clinicians can disagree on it~\citep{jafari2026expert}. In the same spirit, although some of our categories (e.g., \textit{Psychoeducation}) are core components of evidence-based second- and third-wave cognitive behavioral therapies (Cognitive Behavioral Therapy~\citep{beck2020cognitive}, Acceptance and Commitment Therapy~\citep{hayes2006acceptance}, Dialectical Behavioral Therapy~\citep{linehan2025dbt}) that are directly linked to better patient outcomes~\citep{van2021systematic,kambeitz2022systematic}, we do not claim these connections explicitly: our analysis focuses on the patterns of, and alignment between, human and LLM-provided care.}

To apply the ontology at scale, we first introduce an automatic classifier that labels every clinician turn with its corresponding moves, and find that it matches the agreement among expert annotators.
Then, we move to characterizing the move distributions between human clinicians and a panel of three frontier models, leveraging both human transcripts \citep{alexanderstreettranscripts} as an anchor and fully synthetic conversations (Figure~\ref{fig:therapy_move_comparison}).
Anchoring on human context places the model in a similar position as the clinician, so that their moves can be compared directly.
On the other hand, fully synthetic conversations, in which the model leads the session from the start, better reflect how models are used \textit{in the wild}, where the model has no clinician turns to take its cues from on which moves to make, in what style, and when.
Finally, exploiting the extensive agentic training of current models \citep{glm5team2026glm5vibecodingagentic, team2025kimi}, we propose a way for models to use the ontology themselves, in which each move is exposed as a tool that the model calls before responding, showing that this brings their move distributions closer to those of human clinicians without any fine-tuning.

Our contributions are as follows:
\begin{itemize}
    \item We develop and validate an ontology of therapeutic moves, and propose an automatic judge classifier that achieves reliability similar to that of human annotators (\S\ref{sec:ontology}).
    \item We provide a quantitative analysis of how human clinicians and LLMs conduct psychotherapy, highlighting patterns such as models' large tendency, compared to humans, for continued probing, and the markedly different distributions they produce when human context is provided compared to when they lead the session themselves from the start~(\S\ref{sec:gen_analysis}).
    \item We propose exposing the moves as model tools and show that it steers models toward the human distribution: it roughly halves the mean deviation without any fine-tuning and raises turn-level agreement with the clinician's move by 7--9 percentage points, whereas placing the same ontology in the system prompt has little effect (\S\ref{sec:moves_framework}).
\end{itemize}

We will publicly release the full ontology, the classifier setup and tool definitions, our code, and the fully synthetic conversations together with their expert annotations.\footnote{We will share these resources upon acceptance. The human transcripts are licensed and cannot be redistributed.}

\begin{figure*}[t]
    \centering
\providecolor{swordblue}{RGB}{58,84,176}%
\providecolor{swordcoral}{RGB}{194,64,47}%
\providecolor{tznoir}{HTML}{20222c}%
\providecolor{tzline}{HTML}{e3e5e8}%
\providecolor{tzgray}{HTML}{f4f5f7}%
\providecolor{tzmuted}{HTML}{8a8f98}%
\providecolor{tzink}{HTML}{2a2c36}%
\providecolor{tzdark}{HTML}{55575e}%
\providecolor{tzsky}{HTML}{3F8FC1}%
\providecolor{tzrose}{HTML}{D94F67}%
\newcommand{\tzPaperRoman}{\paperromanfont}%
\newcommand{\tzBody}{\tzPaperRoman\scriptsize
  \hyphenpenalty=10000\exhyphenpenalty=10000\color{tzink}}%
\newcommand{\tzLabelSize}{\fontsize{6}{7}\selectfont}%
\newcommand{\tzSrc}[1]{\labelfont\tzLabelSize\color{#1}}%
\newcommand{\tzHead}{\labelfont\small}%
\newcommand{\tzPill}{\labelfont\tzLabelSize}%
\newcommand{\tzTag}{\ttfamily\tzLabelSize\color{tzsky}}%
\newcommand{\tzVia}{\labelfont\tzLabelSize\color{tzdark}}%
\begin{tikzpicture}[
  bubble/.style={execute at begin node={\hyphenpenalty=10000\exhyphenpenalty=10000\spaceskip=0pt\relax}, rounded corners=5pt, inner xsep=5.5pt, inner ysep=4pt,
                 text width=4.78cm, align=left, draw=tzline, line width=.5pt, fill=white},
  clientb/.style={bubble, fill=tzgray, draw=tzgray},
  ghostb/.style={rounded corners=5pt, inner xsep=6pt, inner ysep=3.6pt, align=left,
                 draw=tznoir!60, dash pattern=on 2pt off 1.6pt, line width=.6pt, fill=white},
  card/.style={execute at begin node={\hyphenpenalty=10000\exhyphenpenalty=10000\spaceskip=0pt\relax}, rounded corners=4pt, inner xsep=5.5pt, inner ysep=3.8pt,
               text width=3.6cm, align=left, line width=.5pt, fill=white},
  cardH/.style={card, draw=tznoir, line width=.8pt},
  cardM/.style={card, draw=tzline},
  cardN/.style={card, draw=tzline},
  pill/.style={rounded corners=4.5pt, inner xsep=4pt, inner ysep=2.4pt, fill=tznoir, text=white},
  pillM/.style={pill, fill=tzsky},
  pillN/.style={pill, fill=tzrose},
  judgetag/.style={rounded corners=3pt, inner xsep=4pt, inner ysep=2.2pt,
                   fill=white, draw=tzmuted!55, line width=.5pt, text=tzdark,
                   font=\labelfont\tzLabelSize},
  badge/.style={circle, fill=tznoir, text=white, inner sep=0pt, minimum size=10pt,
                font=\labelfont\tzLabelSize},
  branch connector/.style={draw=tzmuted!48, line width=.85pt,
                           dash pattern=on .55pt off 1.35pt, line cap=round,
                           rounded corners=2.5pt},
]
\providecommand{\tzFigureOneSetup}{}%
\tzFigureOneSetup

\node[anchor=north west] (Lhead) at (0,0)
  {\tzHead\color{tzink}Human session \textcolor{tzmuted}{· Transcript}};

\node[bubble, anchor=north west] at ($(Lhead.south west)+(0,-0.22)$) (t1)
  {\tzSrc{tzmuted}Clinician\\[1.2pt]\tzBody When you have to set the new rates at the plant, what goes through your mind?};
\node[clientb, anchor=north west] at ($(t1.south west)+(0.5,-0.13)$) (c1)
  {\tzSrc{tzmuted}Patient\\[1.2pt]\tzBody That the operators won't accept them. I get anxious just thinking about it.};
\node[anchor=north] at (c1.south -| 2.84,0) (e1) {\tzBody\color{tzmuted}•\,\,•\,\,•};

\node[bubble, anchor=north west] at ($(c1.south west)+(-0.5,-0.34)$) (t2)
  {\tzSrc{tzmuted}Clinician\\[1.2pt]\tzBody And then what — what's the worst that could happen?};
\node[clientb, anchor=north west] at ($(t2.south west)+(0.5,-0.13)$) (c2)
  {\tzSrc{tzmuted}Patient\\[1.2pt]\tzBody They'd quit. And I'd need two new people to replace one experienced person. That's about as bad as it gets.};
\node[ghostb, anchor=north west] at ($(c2.south west)+(-0.5,-0.13)$) (g1)
  {\tzSrc{tzmuted}Clinician — turn A\hspace{14pt}};
\node[badge] at (g1.east) (b1) {A};
\node[anchor=north] at (g1.south -| 2.84,0) (e2) {\tzBody\color{tzmuted}•\,\,•\,\,•};

\node[bubble, anchor=north west] at ($(g1.south west)+(0,-0.36)$) (t3)
  {\tzSrc{tzmuted}Clinician\\[1.2pt]\tzBody You keep asking yourself: ``Why didn't you perform better, Mark?''};
\node[clientb, anchor=north west] at ($(t3.south west)+(0.5,-0.13)$) (c3)
  {\tzSrc{tzmuted}Patient\\[1.2pt]\tzBody Right. And is there even an answer to that?};
\node[ghostb, anchor=north west] at ($(c3.south west)+(-0.5,-0.13)$) (g2)
  {\tzSrc{tzmuted}Clinician — turn B\hspace{14pt}};
\node[badge] at (g2.east) (b2) {B};

\def\sax{6.12}    
\def\sbx{11.07}  

\node[anchor=north west] (Mhead) at (\sax,0)
  {\tzHead\color{tzink}Next-turn continuations};

\node[anchor=north west] at (\sax+0.26,-0.8) (s1title)
  {\tzSrc{tzmuted}Turn A};

\node[cardH, anchor=north west] at ($(s1title.south west)+(0,-0.11)$) (s1h)
  {\tzSrc{tznoir}Clinician (gold)\\[5pt]\tzBody So that's rock bottom. Looking at it squarely — how awful is it, really, to have two people doing the work of one?};
\node[pill, anchor=north east] at ($(s1h.north east)+(-0.13,-0.1)$) {\tzPill Challenge};

\node[cardM, anchor=north west] at ($(s1h.south west)+(0,-0.14)$) (s1m)
  {\tzSrc{tzsky}LLM · With Moves\\[5pt]
   {\tzVia via tool}\hspace{3pt}{\tzTag do\_challenge()}\\[1.6pt]
   \tzBody Losing an operator would be genuinely inconvenient — but does it actually threaten you, or your job, in any serious way?};
\node[pillM, anchor=north east] at ($(s1m.north east)+(-0.13,-0.1)$) {\tzPill Challenge};

\node[cardN, anchor=north west] at ($(s1m.south west)+(0,-0.14)$) (s1n)
  {\tzSrc{tzrose}LLM · No Moves\\[5pt]\tzBody How often has that actually happened when you've set a rate someone disagreed with?};
\node[pillN, anchor=north east] at ($(s1n.north east)+(-0.13,-0.1)$) {\tzPill Inquiry};

\begin{scope}[on background layer]
\node[rounded corners=5.5pt, draw=tzline, line width=.5pt, fill=white,
      fit={(s1title)(s1h)(s1n)}, inner xsep=6.5pt, inner ysep=6pt] (stack1) {};
\end{scope}
\node[badge] at (stack1.west |- s1title.west) (sb1) {A};

\node[anchor=north west] at (\sbx+0.26,-0.8) (s2title)
  {\tzSrc{tzmuted}Turn B};

\node[cardH, anchor=north west] at ($(s2title.south west)+(0,-0.11)$) (s2h)
  {\tzSrc{tznoir}Clinician (gold)\\[5pt]\tzBody Let's try it. I'll be Mark: ``Mark, why don't you respond better?'' — now you answer back.};
\node[pill, anchor=north east] at ($(s2h.north east)+(-0.13,-0.1)$) {\tzPill Skill building};

\node[cardM, anchor=north west] at ($(s2h.south west)+(0,-0.14)$) (s2m)
  {\tzSrc{tzsky}LLM · With Moves\\[5pt]
   {\tzVia via tool}\hspace{3pt}{\tzTag do\_skill\_building()}\\[1.6pt]
   \tzBody What would you say to a friend who came to you with that exact question — why didn't I perform better?};
\node[pillM, anchor=north east] at ($(s2m.north east)+(-0.13,-0.1)$) {\tzPill Skill building};

\node[cardN, anchor=north west] at ($(s2m.south west)+(0,-0.14)$) (s2n)
  {\tzSrc{tzrose}LLM · No Moves\\[5pt]\tzBody What do you think the honest answer actually is?};
\node[pillN, anchor=north east] at ($(s2n.north east)+(-0.13,-0.1)$) {\tzPill Inquiry};

\begin{scope}[on background layer]
\node[rounded corners=5.5pt, draw=tzline, line width=.5pt, fill=white,
      fit={(s2title)(s2h)(s2n)}, inner xsep=6.5pt, inner ysep=6pt] (stack2) {};
\end{scope}
\node[badge] at (stack2.west |- s2title.west) (sb2) {B};
\node[judgetag, anchor=east] at (stack2.east |- Mhead.center) {All moves assigned by an LLM judge};

\begin{scope}[on background layer]
  \draw[branch connector] (b1.east) -- (b1.east -| stack1.west);
  \draw[branch connector] (b2.east) -| (stack2.south);
\end{scope}

\end{tikzpicture}%
    \vspace{5pt}
    \caption{Schematic of one of our experimental designs.
    Given the rolling prefix of a human therapy session, a panel of LLMs generates the next clinician turn, either freely (No-Moves) or with the ontology exposed as tools (With-Moves).
    An LLM judge assigns each continuation a move, which we compare against the human clinician's gold turn.
    Utterances shown are illustrative paraphrases of a real cognitive-therapy session.
    }
    \label{fig:therapy_move_comparison}
\end{figure*}

\section{Background}
\begin{table*}[t]
\centering
\footnotesize
\setlength{\tabcolsep}{4pt}
\renewcommand{\arraystretch}{1.12}
\begin{tabularx}{\textwidth}{
  >{\raggedright\arraybackslash}p{0.23\textwidth}
  >{\raggedright\arraybackslash}p{0.38\textwidth}
  >{\raggedright\arraybackslash}X}
\toprule
\textbf{Move} &
\textbf{Operational function} &
\textbf{Example therapist turn} \\
{\scriptsize\itshape MULTI-60 grounding} & & \\
\midrule

\texttt{do\_process\_alignment}\newline
{\scriptsize\textcolor{gray}{28, 38}} &
Calibrate collaboration, pacing, fit, or repair of the therapeutic process. &
``I think I moved too quickly there. How did that land for you?'' \\

\texttt{do\_goal\_setting}\newline
{\scriptsize\textcolor{gray}{1, 9, 28, 42}} &
Identify, clarify, or prioritize what the client wants to work toward. &
``Let us choose which of these two goals would be most useful to focus on today.'' \\

\texttt{do\_inquiry}\newline
{\scriptsize\textcolor{gray}{40, 46, 47}} &
Seek information or deepen understanding of meaning, sequence, context, or detail. &
``When the panic started, what did you notice first in your body?'' \\

\texttt{do\_shared\_understanding}\newline
{\scriptsize\textcolor{gray}{10, 18, 31}} &
Reflect, validate, normalize, paraphrase, or synthesize without adding a new explanatory claim. &
``You felt trapped and overwhelmed, and then you shut down.'' \\

\texttt{do\_support\_change}\newline
{\scriptsize\textcolor{gray}{7, 23, 25, 42, 52, 56}} &
Strengthen or explore hope, agency, readiness, confidence, ambivalence, or client-owned reasons for change. &
``What feels most important to you about making this change now?'' \\

\texttt{do\_interpret}\newline
{\scriptsize\textcolor{gray}{2, 19, 20, 27}} &
Offer a therapist-generated hypothesis about meaning, function, cause, or pattern. &
``I wonder if withdrawing protects you from the rejection you expect.'' \\

\texttt{do\_challenge}\newline
{\scriptsize\textcolor{gray}{13, 21, 37, 39, 49}} &
Examine a discrepancy, assumption, rigid belief, or coping pattern that may maintain the problem. &
``What evidence supports that idea, and what evidence pushes against it?'' \\

\texttt{do\_psychoeducation}\newline
{\scriptsize\textcolor{gray}{32, 58, 59}} &
Explain relevant information, theory, mechanism, or intervention rationale. &
``Avoidance lowers distress quickly, but that relief can keep fear going over time.'' \\

\texttt{do\_action\_planning}\newline
{\scriptsize\textcolor{gray}{9, 16, 17, 35, 51}} &
Translate a goal or insight into a concrete next step that can later be reviewed. &
``Would a five-minute walk after dinner be a realistic first step this week?'' \\

\texttt{do\_skill\_building}\newline
{\scriptsize\textcolor{gray}{15, 16, 47}} &
Introduce, teach, guide, or rehearse an identifiable skill within the interaction. &
``Let us try one slow breathing cycle together; first, breathe out fully.'' \\
\bottomrule
\end{tabularx}

\caption{The ten therapeutic moves. MULTI-60 item numbers provide provenance and may overlap across moves; complete definitions, positive and negative examples, and disambiguation rules appear in Appendix~\ref{app:ontology}.}
\label{tab:ontology}
\end{table*}

\subsection{Coding Psychotherapy}

Psychotherapy process research has long treated clinician language as a sequence of observable actions. 
Early \emph{microcounseling} work decomposed interviewing into trainable behaviors such as attending and questioning \citep{ivey1968microcounseling}. 
Hill's Counselor Verbal Response Category System similarly coded counselor responses into mutually exclusive categories and was later organized into the helping stages of exploration, insight, and action \citep{hill1978development,hill2014helping}. 
Related response-mode taxonomies classified what an utterance does---for example, questioning, reflection, interpretation, or advisement---rather than its topic \citep{stiles1979verbal,elliott1987primary}.
These traditions provide useful turn-local descriptions, but their inventories and unitization rules were designed primarily for human training and process research.
For example, the Cognitive Therapy Rating Scale assesses competence in cognitive therapy \citep{young1980cognitive}, while Motivation Interviewing Skills Code and Motivational Interviewing Treatment Integrity code adherence to Motivational Interviewing \citep{miller2003misc,moyers2016miti}. 
Such instruments offer clinically specific distinctions, but their labels are not directly applicable across therapeutic modalities.

The Multitheoretical List of Therapeutic Interventions (MULTI) was developed to bridge this divide. 
MULTI-60 describes therapist behavior using 60 jargon-reduced items grouped under eight orientations: psychodynamic, process-experiential, cognitive, interpersonal, behavioral, dialectical behavioral, person-centered and common factors \citep{mccarthy2009multi,graham2020manual}.
Importantly, MULTI is a descriptive measure of interventions rather than a measure of adherence or competence, and its standard forms summarize how characteristic each item is of a session.
Prior work has adapted MULTI to talk-turn classification, illustrating both its value as a cross-theoretical vocabulary and the difficulty of applying a session-oriented inventory densely at the turn level \citep{mehta2022multi}.
Our work directly builds on top of MULTI by further filtering and grouping items into a compact ontology that employs the most common and high impact clinical strategies applicable to evidence-based psychotherapy.
This shorter list is better suited for models both by reducing the breadth of options they must reason over, but also limiting the number of high-confusion categories.

\subsection{Large Language Models}

LLMs acquire broad capabilities through pretraining on heterogeneous text and code, after which instruction tuning and preference optimization shape them into general-purpose conversational assistants \citep{brown2020language,wei2022finetuned,ouyang2022training,bai2022traininghelpfulharmlessassistant, zhang2025interplaypretrainingmidtrainingrl}. 
Moreover, recent model development has evolved beyond generating text to conducting structured actions \citep{10.1145/3704435} (tools) and pursuing goals over multiple steps by interacting with environments such as repositories, terminals, browsers, and retrieval systems \citep{team2025kimi,glm5team2026glm5vibecodingagentic}.
This broadening of scope has led to models becoming more and more vital for everyday tasks and outcomes \citep{zuo2026qwen}, including mental-health related topics \citep{wu2026wisemind}.  
In this work, we both investigate specificities in how models' conduct psychotherapy, but also steer the model by configuring our ontology as a set of tools (\S\ref{sec:moves_framework})

\section{Ontology}

\label{sec:ontology}

Our ontology maps each therapist turn to one or more of ten \emph{moves}: compact, function-based categories intended to be distinguishable from the local turn textual form.
The inventory draws on the counseling-skills and response-mode traditions above. 
Each move is grounded in one or more MULTI-60 items (Table~\ref{tab:ontology}) providing important clinical traceability, and were co-defined with licensed clinical psychologists to ensure clinical grounding. This collaboration extended beyond the high-level definition of the moves to how the ontology is put to use in this work \textit{operationally}: the annotation guidelines and decision rules (Appendix~\ref{app:ontology}), the training of the annotators (Appendix~\ref{app:anotation_procedure}), the judge prompt (Figure~\ref{fig:judge-prompt}), and the tool descriptions and responses through which the moves are exposed to models (Appendix~\ref{app:tools}).

We made our ontology suitable for transcript annotation and LLM classification through three design choices. 
First, coding follows therapeutic \emph{function} rather than syntax (e.g. a question that tests a rigid belief is a marked as \textit{Challenge}, not only an \textit{Inquiry}). 
Second, labels are multi-label because one turn may both reflect a patient's experience while asking for clarification. 
Third, the ontology guidelines specifies contrastive boundaries for commonly confused pairs, including inquiry versus challenge, shared understanding versus interpretation, and action planning versus in-session skill building. 
We additionally use \texttt{no\_defined\_move} as an exclusive control label for administrative, social, or procedural turns.

\subsection{Human Validation}
\label{sec:human_validation} 

In addition to developing the ontology with licensed psychologists, we also assessed whether the ontology's definitions were sufficiently operational for clinicians who had not participated in its design. 
Five doctoral-level (PhD/PsyD) US-based licensed psychologists with more than six years of independent clinical practice, \emph{independently} labeled every therapist turn across three corpora.  
Annotation was multi-label: annotators assigned one or more moves whenever a turn explicitly performed multiple therapeutic functions.
Annotators received the complete training material in Appendix~\ref{app:ontology}, which provides an operational definition, positive examples, contraindications, and contrastive disambiguation guidance for each move. 

\paragraph{Setup.}
As for validation data, we include both the human therapy and LLM-generated transcripts, allowing us to test whether the same move definitions could be recognized across naturally occurring and synthetic interactions. 
The human corpus was drawn from the Alexander Street \emph{Counseling and Psychotherapy Transcripts} collection, which contains therapist--client sessions as well as demonstrations produced for clinical training \citep{alexanderstreettranscripts}.
We retained transcripts from Cognitive Behavioral Therapy (CBT) and closely related modalities using the criteria described in \S\ref{app:data_filtering}. The resulting validation set comprised 19 human transcripts (868 therapist turns) and 18 synthetic transcripts generated with Claude Sonnet 4.6. Additional details on the annotation procedure and corpus composition are provided in Appendix~\ref{app:anotation_procedure}.
For measuring agreement, as a turn may perform more than one move, each turn receives a \emph{set} of labels rather than a single category. 
We therefore measure agreement with Krippendorff's $\alpha$ \citep{krippendorff2018content} computed under Jaccard distance on the annotators' move sets, which compares the two sets directly and credits partial overlap.

\paragraph{Results.}
Table~\ref{tab:human_iaa} reports pairwise agreement between annotators on human transcripts.\footnote{Table~\ref{tab:agreement} reports the agreement levels for the remaining annotated corpus.}
Among the ten pairs of annotators, the mean agreement score was 0.627, with pairwise scores ranging from 0.593 to 0.653. 
We also find that the aggregate agreement result was not driven by a single annotator pair. Indeed, although we find limited prior work on the task of multi-label therapist coding, these values are in line with those reported for therapist skill coding~\citep{hammerfald2026leveraging} and for other annotation tasks in mental health support~\citep{arnaiz2026between, thomas2025large, szoke2026automated, cai2025exploring}. Overall, the numbers reflect the complexity of the task~---~multi-label with 11 possible classes --- and the ambiguity associated with mental health.\footnote{The label distribution is also highly imbalanced (Table~\ref{tab:agreement_per_move_deepseek}), which is known to lower Krippendorff's $\alpha$ \citep{feinstein1990high,gwet2008computing}.}

\begin{table}[t]
\centering
\begin{minipage}[t]{0.48\linewidth}
\vspace{0pt}
\centering
\scriptsize
\captionsetup{font=footnotesize}
\setlength{\tabcolsep}{2.5pt}
\renewcommand{\arraystretch}{1.0}
\begin{tabular}{lrrrrrr}
\toprule
 & \textbf{A1} & \textbf{A2} & \textbf{A3} & \textbf{A4} &
\textbf{A5} & \textbf{Mean} \\
\midrule
\textbf{A1} &       &       &       &       &       & 0.622 \\
\textbf{A2} & 0.593 &       &       &       &       & 0.613 \\
\textbf{A3} & 0.631 & 0.608 &       &       &       & 0.632 \\
\textbf{A4} & 0.604 & 0.623 & 0.653 &       &       & 0.629 \\
\textbf{A5} & 0.658 & 0.627 & 0.636 & 0.635 &       & 0.639 \\
\cdashlinelr{1-7}
\textbf{Classifier} & 0.635 & 0.611 & 0.691 & 0.664 & 0.618 & 0.644 \\
\bottomrule
\end{tabular}
\caption{Pairwise Krippendorff's $\alpha$ under Jaccard distance between
annotators on the human therapy transcripts.
\textit{Mean} is the average of an annotator's four pairwise coefficients.
\textit{Classifier} is our LLM judge classifier.}
\label{tab:human_iaa}
\end{minipage}\hfill
\begin{minipage}[t]{0.48\linewidth}
\vspace{0pt}
\scriptsize
\centering
\captionsetup{font=footnotesize}
\setlength{\tabcolsep}{3pt}
\renewcommand{\arraystretch}{1.0}
\begin{tabular}{lrcc}
\toprule
Move & $n$ & Annotators & Classifier \\
\midrule
Inquiry & 1472 & 0.710 & 0.761 \\
Shared understanding & 1286 & 0.652 & 0.674 \\
Interpret & 176 & 0.454 & 0.539 \\
Action planning & 142 & 0.590 & 0.657 \\
Skill building & 79 & 0.592 & 0.670 \\
Challenge & 66 & 0.335 & 0.467 \\
Goal setting & 60 & 0.578 & 0.647 \\
Support change & 47 & 0.185 & 0.326 \\
Psychoeducation & 34 & 0.584 & 0.540 \\
Process alignment & 9 & 0.329 & 0.306 \\
\bottomrule
\end{tabular}
\caption{Krippendorff's $\alpha$ agreement per move on the annotated transcripts. }
\label{tab:agreement_per_move_deepseek}
\end{minipage}
\end{table}

Looking at the agreements for each move, Table~\ref{tab:agreement_per_move_deepseek}, shows that some moves have low agreement levels: \textit{Process Alignment}, \textit{Support Change}, and \textit{Challenge}. These moves are harder to identify reliably; therefore, results involving them should be interpreted with caution. 
\paragraph{Judge Classifier.}
We develop a generative move classifier based on Deepseek V4.1 Flash \citep{deepseekai2026deepseekv41flash} to scale our experiments adequately and efficiently.
The judge receives the ontology (Figure~\ref{fig:judge-prompt} shows the exact prompt) and classifies each turn five times. The final classification is selected by majority vote.

We validate the judge against human annotations. As shown in Tables~\ref{tab:human_iaa} and~\ref{tab:agreement_per_move_deepseek}, the judge generally achieves higher agreement than human annotators, indicating it identifies therapist moves with at least human-level reliability. Appendix~\ref{app:judge_validation} further examines the judge's reliability, demonstrating that the judge's predictions exhibit very low variability across repeated classifications.

\section{Experiments}
With the ontology verified and in place, we move to the central point of our analysis: classifying a set of mental health counseling transcripts, human ( Appendix~\ref{app:data_filtering}) and synthetic, and uncovering statistical differences as a proxy to overall similarity or distance to human counseling.
This practice identifies discrepancies between the two but, \emph{does not measure outcomes or intend to set norms or any notion of correct clinical practice}.

Instead, our experiments aim to answer the following research questions with a descriptive point-of-view: 
\begin{itemize}
    \item How does clinician and LLM-based moves' structure compare under a human-induced therapy context? (\S\ref{sec:gen_analysis}) 
    \item Are our conclusions the same when there is no human grounding on therapy structure, and we have fully synthetic sessions? (\S\ref{sec:gen_analysis})
    \item Is the models' clinical approach similar to humans throughout the duration of each session? (\S\ref{sec:moves_over_time})
    \item Can we influence models towards a more human-like move distribution? (\S\ref{sec:moves_framework})

\end{itemize}

\begin{figure*}[ht]
    \centering
\begingroup
\definecolor{distributionHuman}{HTML}{0B0B0B}
\definecolor{distributionSonnet}{HTML}{D97757}
\definecolor{distributionSonnetPale}{HTML}{F0EEE6}
\definecolor{distributionGLM}{HTML}{3F8FC1}
\definecolor{distributionGPT}{HTML}{3A9D78}

\pgfplotsset{
  distribution human/.style={
    ybar, bar width=5.2pt, fill=black!5,
    draw=distributionHuman, line width=0.45pt,
    pattern=crosshatch dots, pattern color=distributionHuman!82,
    mark=none,
  },
  distribution sonnet human led/.style={
    ybar, bar width=3.5pt, fill=distributionSonnet,
    draw=distributionSonnet!80!black, line width=0.3pt, mark=none,
  },
  distribution sonnet model led/.style={
    ybar, bar width=3.5pt, fill=distributionSonnetPale,
    draw=distributionSonnet!75!black, line width=0.4pt,
    pattern=north east lines, pattern color=distributionSonnet!72!black,
    mark=none,
  },
  distribution glm human led/.style={
    ybar, bar width=3.5pt, fill=distributionGLM,
    draw=distributionGLM!80!black, line width=0.3pt, mark=none,
  },
  distribution glm model led/.style={
    ybar, bar width=3.5pt, fill=distributionGLM!14,
    draw=distributionGLM!75!black, line width=0.4pt,
    pattern=north east lines, pattern color=distributionGLM!72!black,
    mark=none,
  },
  distribution gpt human led/.style={
    ybar, bar width=3.5pt, fill=distributionGPT,
    draw=distributionGPT!80!black, line width=0.3pt, mark=none,
  },
  distribution gpt model led/.style={
    ybar, bar width=3.5pt, fill=distributionGPT!14,
    draw=distributionGPT!75!black, line width=0.4pt,
    pattern=north east lines, pattern color=distributionGPT!72!black,
    mark=none,
  },
}
\tikzset{led ci/.style={draw=black!70, line width=0.35pt}}

\newcommand{\ledbars}[7]{%
  \addplot[distribution human, bar shift=0pt, forget plot]
    coordinates {(0,#1)};
  \addplot[distribution sonnet human led, bar shift=-1.9pt, forget plot]
    coordinates {(1,#2)};
  \addplot[distribution sonnet model led, bar shift=1.9pt, forget plot]
    coordinates {(1,#3)};
  \addplot[distribution glm human led, bar shift=-1.9pt, forget plot]
    coordinates {(2,#4)};
  \addplot[distribution glm model led, bar shift=1.9pt, forget plot]
    coordinates {(2,#5)};
  \addplot[distribution gpt human led, bar shift=-1.9pt, forget plot]
    coordinates {(3,#6)};
  \addplot[distribution gpt model led, bar shift=1.9pt, forget plot]
    coordinates {(3,#7)};
}

\newif\ifledci \ledcitrue
\newcommand{\ledwhisker}[4]{%
  \ifdim #3pt=#4pt\relax\else
    \draw[led ci] ([xshift=#2]axis cs:#1,#3) -- ([xshift=#2]axis cs:#1,#4);
    \draw[led ci] ([xshift=#2-0.8pt]axis cs:#1,#3) -- ([xshift=#2+0.8pt]axis cs:#1,#3);
    \draw[led ci] ([xshift=#2-0.8pt]axis cs:#1,#4) -- ([xshift=#2+0.8pt]axis cs:#1,#4);
  \fi}
\newcommand{\ledci}[5]{%
  \ifledci
    \ledwhisker{#1}{-1.9pt}{#2}{#3}%
    \ledwhisker{#1}{1.9pt}{#4}{#5}%
  \fi}

\newcommand{\ledlegend}{%
  \addlegendimage{area legend, fill=black!5, draw=distributionHuman,
    line width=0.45pt, pattern=crosshatch dots,
    pattern color=distributionHuman!82}
  \addlegendentry{Human}
  \addlegendimage{area legend, fill=black!72, draw=black!72}
  \addlegendentry{Human-led}
  \addlegendimage{area legend, fill=black!10, draw=black!60,
    pattern=north east lines, pattern color=black!60}
  \addlegendentry{Model-led}
}
\newcommand{\ledtitle}[1]{{\labelfont #1}}

\begin{tikzpicture}[font={\paperromanfont\footnotesize}, text=black]
\begin{groupplot}[
  group style={
    group size=5 by 2,
    horizontal sep=0.55cm,
    vertical sep=1.20cm,
  },
  scale only axis,
  width=0.145\linewidth,
  height=1.78cm,
  xmin=-0.28, xmax=3.28,
  xtick={0,1,2,3},
  xticklabels=\empty,
  tick align=outside,
  tick style={draw=black!45, line width=0.35pt},
  tick label style={font={\paperromanfont\fontsize{5.8}{6.7}\selectfont}, text=black},
  xticklabel style={font={\paperromanfont\fontsize{5.3}{6.1}\selectfont}, text=black},
  ylabel style={font={\paperromanfont\fontsize{6.1}{7.0}\selectfont}, text=black},
  axis line style={draw=black!45, line width=0.45pt},
  axis x line*=bottom,
  axis y line*=left,
  ymajorgrids,
  grid style={draw=black!11, line width=0.35pt},
  scaled ticks=false,
  title style={
    font=\fontsize{6.3}{7.2}\selectfont,
    align=center, text=black,
    at={(axis description cs:0.5,1.005)}, anchor=south,
  },
  clip mode=individual,
]

\nextgroupplot[
  title={\ledtitle{Shared understanding}},
  ylabel={\% of moves},
  ymin=0, ymax=51, ytick={0,15,30,45},
  legend to name=ledLegend,
  legend columns=3,
  legend style={
    draw=none, fill=none,
    font={\paperromanfont\fontsize{6.3}{7.2}\selectfont},
    /tikz/every even column/.append style={column sep=8pt},
  },
]
\ledlegend
\ledbars{46.744499}{10.413553}{31.385006}{30.118694}{37.075718}{9.984152}{34.353918}
\ledci{1}{9.658925}{11.168180}{29.970429}{32.799583}  
\ledci{2}{28.427340}{31.810048}{35.209464}{38.941972}  
\ledci{3}{6.717810}{13.250494}{33.723702}{34.984134}  

\nextgroupplot[
  title={\ledtitle{Inquiry}},
  ymin=0, ymax=77, ytick={0,20,40,60},
]
\ledbars{22.631343}{69.257598}{50.868276}{45.202770}{41.166232}{66.191231}{30.254281}
\ledci{1}{68.203027}{70.312170}{49.676048}{52.060504}  
\ledci{2}{43.039273}{47.366266}{39.449999}{42.882464}  
\ledci{3}{61.209047}{71.173414}{28.640879}{31.867684}  

\nextgroupplot[
  title={\ledtitle{Skill building}},
  ymin=0, ymax=8.9, ytick={0,3,6},
]
\ledbars{6.825326}{7.224714}{0.000000}{7.171118}{0.174064}{6.867406}{0.311365}
\ledci{1}{6.614671}{7.834756}{0.000000}{0.000000}  
\ledci{2}{6.109748}{8.232487}{0.000000}{0.390779}  
\ledci{3}{5.904174}{7.830638}{0.121524}{0.501206}  

\nextgroupplot[
  title={\ledtitle{Psychoeducation}},
  ymin=0, ymax=4.9, ytick={0,2,4},
]
\ledbars{4.445442}{0.448430}{0.423549}{0.890208}{0.522193}{0.792393}{3.580695}
\ledci{1}{0.112311}{0.784550}{0.000000}{0.871849}  
\ledci{2}{0.114812}{1.665604}{0.342100}{0.702286}  
\ledci{3}{0.017140}{1.567646}{2.678727}{4.482664}  

\nextgroupplot[
  title={\ledtitle{Action planning}},
  ymin=0, ymax=13, ytick={0,4,8,12},
]
\ledbars{3.547373}{3.886398}{2.414231}{4.005935}{5.221932}{5.969361}{10.638298}
\ledci{1}{2.583212}{5.189583}{2.062854}{2.765609}  
\ledci{2}{2.511835}{5.500035}{4.632648}{5.811216}  
\ledci{3}{4.688667}{7.250055}{9.881747}{11.394849}  

\nextgroupplot[
  title={\ledtitle{Challenge}},
  ylabel={\% of moves},
  xticklabels={Human,Sonnet,GLM,GPT},
  ymin=0, ymax=7, ytick={0,2,4,6},
]
\ledbars{3.322856}{4.882910}{4.489623}{5.044510}{4.090513}{3.909139}{2.231448}
\ledci{1}{3.613413}{6.152407}{3.083424}{5.895822}  
\ledci{2}{3.682552}{6.406469}{3.489623}{4.691404}  
\ledci{3}{3.167235}{4.651042}{1.636418}{2.826477}  

\nextgroupplot[
  title={\ledtitle{Interpret}},
  xticklabels={Human,Sonnet,GLM,GPT},
  ymin=0, ymax=15, ytick={0,4,8,12},
]
\ledbars{2.828918}{0.597907}{7.073274}{3.066271}{8.093995}{0.739567}{11.987545}
\ledci{1}{0.192842}{1.002972}{5.740055}{8.406493}  
\ledci{2}{2.445148}{3.687394}{7.458435}{8.729555}  
\ledci{3}{0.102853}{1.376281}{10.692967}{13.282124}  

\nextgroupplot[
  title={\ledtitle{Goal setting}},
  xticklabels={Human,Sonnet,GLM,GPT},
  ymin=0, ymax=4.7, ytick={0,2,4},
]
\ledbars{2.514594}{2.092676}{1.905972}{2.621167}{1.697128}{3.803487}{2.854177}
\ledci{1}{1.770542}{2.414809}{1.513026}{2.298918}  
\ledci{2}{1.649581}{3.592753}{1.276348}{2.117908}  
\ledci{3}{3.267889}{4.339084}{2.465833}{3.242522}  

\nextgroupplot[
  title={\ledtitle{Process alignment}},
  xticklabels={Human,Sonnet,GLM,GPT},
  ymin=0, ymax=2.4, ytick={0,1,2},
]
\ledbars{2.200269}{0.548082}{0.127065}{0.791296}{0.130548}{1.215003}{0.207577}
\ledci{1}{0.358446}{0.737717}{0.000000}{0.366363}  
\ledci{2}{0.528022}{1.054569}{0.000000}{0.451557}  
\ledci{3}{0.583064}{1.846941}{0.069863}{0.345290}  

\nextgroupplot[
  title={\ledtitle{Support change}},
  xticklabels={Human,Sonnet,GLM,GPT},
  ymin=0, ymax=2.5, ytick={0,1,2},
]
\ledbars{1.481814}{0.149477}{1.101228}{1.038576}{1.610096}{0.264131}{1.453036}
\ledci{1}{0.000000}{0.329231}{0.603020}{1.599436}  
\ledci{2}{0.000000}{2.091590}{1.085751}{2.134440}  
\ledci{3}{0.000000}{0.841512}{0.658824}{2.247248}  

\end{groupplot}

\node[anchor=north]
  at ([yshift=-0.63cm]$(group c1r2.south)!0.5!(group c5r2.south)$)
  {\pgfplotslegendfromname{ledLegend}};
\end{tikzpicture}
\endgroup
    \caption{Probability of performing each move in the human transcripts. Each panel displays the move distribution for Claude Sonnet~4.6, GLM~5.2, and GPT~5.6~Terra, generated with the human-context as anchor (solid) and the synthetic grounding-free version (hatched).}
    \label{fig:move_distribution}
\end{figure*}

\subsection{Anchored and Context-Free Transcripts}
\label{sec:gen_analysis}

For the initial experiments, models act as the clinician under a rolling prefix of the human transcripts (depicted in Figure~\ref{fig:therapy_move_comparison}).
This approach mimics chat-based therapy and grounds models in the preceding human style and structure.
The overall comparison, broken down by move and model can be observed in Figure~\ref{fig:move_distribution}.
Here, we compute confidence intervals by prompting each model four times per turn, both surfacing the underlying model distribution \citep{wang2023selfconsistency} and addressing the ambiguity of the task.
For the model-based clinician, we rely on a panel of open and close-source models: GLM 5.2 \citep{glm5team2026glm5vibecodingagentic}, Claude Sonnet 4.6 \citep{sonnet46} and GPT 5.6 Terra \citep{gptterra}.
The clinician system prompt was clinically and carefully co-developed and can be found in Figure~\ref{fig:therapist-prompt}.

That withstanding, while this configuration allows for a direct comparison between approaches within the same context, it obfuscates how the model would conduct therapy were the human anchors not present.
In fact, previous human clinician utterances bias the LLM-therapist in terms of style and function of the messages to be conveyed.

To bypass this issue and better understand how models would behave in a more realistic setting, we construct fully synthetic transcripts where LLMs are both therapist and patient (further details in Appendix~\ref{app:synthetic_transcripts}.
This process yielded a further 40 total transcripts, which we validate through a further annotation round, noting overall annotator agreement remains generally consistent with the human version (Table~\ref{tab:agreement}) and thus the ontology is likewise applicable in the fully synthetic setting. 
We discuss the contrasting results below:

\paragraph{Inquiry is the dominant LLM move.} 
Perhaps a consequence of RLHF and assistant-like training \citep{ouyang2022training, bai2022traininghelpfulharmlessassistant, lambert2026reinforcement}, we find models prefer probing to any other category during therapy sessions.
We find many instances where, even if the patient has shared sufficient details to move the conversation forward, models continue pushing on the same topic whereas humans move to actionable content (e.g. Figure~\ref{fig:inquiry-examples}).
In contrast, \textit{Psychoeducation}, \textit{Skill Building} and (possibly) \textit{Process Alignment} are generally outside of the model distribution with GPT 5.6 Terra being a notable exception for the first.
\textit{Skill Building}'s rate evolves from matching humans to very residual in synthetic conversations.
Parsing through transcripts for an explanation, we re-encounter context anchoring: skill building typically unfolds over several consecutive turns and once a human clinician proposes or starts an exercise, the model will carry it forward, but it does not initiate one on its own, reinforcing the locality bias problem.
To further verify, we construct transition matrices (Figures~\ref{fig:moves_transition_matrix},\ref{fig:moves_transition_matrix_synthetic}) between categories and note the large emphasis on their diagonal for both human and synthetic contexts.
Lastly, we use hedging language when discussing \textit{Support Change} and \textit{Process Alignment} due to the poor inter-annotator agreement for these categories (\S\ref{sec:human_validation}).

\paragraph{Removing the human prefix changes the move distribution substantially.}
As we have touched on with the \textit{Skill Building} case, removing the style and content anchor can alter relative frequencies.
Notably, \textit{Inquiry} has roughly \emph{half} of its probability mass get dispersed for the GPT model, a significant $\sim35\%$ drop. 
In the reverse, \textit{Shared Understanding} (SU), \textit{Interpret} and \textit{Action Planning} (to a lesser degree), observe relative upticks.
SU is commonly associated with models producing validation or reflections of the user message content -- typically associated with LLM written style\citep{scholich2025comparison, sharma2024towards}.

\paragraph{Differences and similarities between models.}
Despite some similarities, such as models' tendency to do inquiry, models have different moves profiles. Overall, GLM is closest to human clinicians (Table~\ref{tab:mad_moves}), and with less propensity to do \textit{Inquiry} , GLM's deviation to the human reference is approximately half of Sonnet and GPT. For \textit{SU}, GLM's deviation from the human baseline is approximately 15\%, roughly half that of the other two models. Also, for \textit{Interpret}, GLM shows a probability very close to that of human clinicians, whereas Sonnet and GPT rarely make this move.

Analyzing the move profiles in synthetic context, GPT appears to be the most sensitive to context, nearly halving its \textit{Inquiry} probability and tripling \textit{SU} probability, a pattern also observed for Sonnet.  \textit{Psychoeducation} is largely neglected by all models in the human-anchored setting; GPT performs this move much more frequently in synthetic conversations. GPT also shows a strong tendency to do \textit{Action Planning}, contrary to other models.

\begin{figure*}
    \centering
    \input{figures/moves_trajectory_usage_majority_of_4_19_led_no_tool_ci_gen4_sonnet_deepseek_tikz}
    \caption{\textbf{Move usage across the conversation for Claude Sonnet 4.6}. Each panel is one move; the $x$ axis is the percentage of therapy session progress.}
    \label{fig:moves_over_time_sonnet}
\end{figure*}

\subsubsection{Move Distributions Over Time}
\label{sec:moves_over_time}

Following the discussion highlighting absolute differences, we now consider how the distribution evolves during one session, or over time.
Some moves, e.g. \textit{Goal Setting} are,  by definition, more likely to occur in the beginning of the conversation as opposed to towards the end, whereas the reverse is true for others, like \textit{Action Planning} which monotonically rise throughout.
To this end, categories that are consistently present throughout sessions (\textit{Inquiry}) or display the distinct from human trends signal different clinical approaches.

Here, for ease of analysis, we only report move usage for a representative model, Sonnet 4.6, with others in the appendix (Figures~\ref{fig:moves_over_time_gpt_terra},~\ref{fig:moves_over_time_glm}).
We find that trends are overall similar between models.

\paragraph{Beyond the prefix.}
Focusing only on the real world transcripts in Figure~\ref{fig:moves_over_time_sonnet} (blue and reference-human black lines), we observe most curves to be overlapping, with \textit{Shared Understanding}, \textit{Inquiry} and \textit{Psychoeducation} as notable exceptions\footnote{Although the \textit{Process Alignment} and \textit{Support Change} categories display similar characteristics, agreement and CIs make any observations inconclusive.}.
Uncoincidentally, we have already covered these exceptions as those less prone to context influence, either due to an overreliance of the move as a clinical approach (\textit{Inquiry}) or due to failure to apply it overall (\textit{Psychoeducation}).
Here, we can also conclusively reinforce that \textit{Skill Building} only occurs after the human therapist initiates the pattern, once we consider synthetic transcripts this move completely falls out of use.

Interestingly, considering the synthetic transcripts, we see a broad increase within the \textit{Interpret} category, surpassing the human rate.
\textit{Interpret} is typically associated with generalizations of patient behavior patterns and so, the $\sim10\%$ prevalence change is significant.
We could perhaps, associate this change with a lack of grounding and stronger emphasis on the clinician persona \citep{marks2026persona} from the system prompt, but alas it remains inconclusive.

This experiment, we find, strengthens the discussion on distinguishing categories outside of the model move support, reinforced by the context bias, or overrelied on in general.

\subsection{Steering Models Towards a More Human-like Move Distribution}\label{sec:moves_framework}

From the previous analysis, we conclude that LLMs have different move than humans. They tend to overuse \textit{Inquiry} and rarely do \textit{Skill Building} or \textit{Psychoeducation} in some settings. \textit{Skill Building} and \textit{Psychoeducation} are core components of evidence-based second- and third-wave cognitive behavioral therapies (Cognitive Behavioral Therapy \citep{beck2020cognitive}, Acceptance and Commitment Therapy \citep{hayes2006acceptance}, Dialectical Behavioral Therapy \citep{linehan2025dbt}), and their underuse suggests that LLMs do not provide these kinds of therapy.

In this section, we propose our training-free framework that provides LLMs with the ontology of clinical skills as a set of tools\footnote{The tool definitions and prompts are provided in Appendix~\ref{app:tools}}. At each turn, the model can select a therapeutic move through the corresponding tool and generate a response implementing that move.
The goal is to bring the model distribution closer to the distribution of moves observed in human clinicians. The approach is model-agnostic and can be easily applied to any LLM. 

We additionally compare our tool-based framework against a prompt-based alternative that provides the same therapeutic move ontology directly in the model's prompt (Appendix~\ref{app:prompt_vs_tools}).

\subsubsection{Analysis of Moves Distributions}
We study how LLMs behave with and without our framework using the same experimental setup described in Section~\ref{sec:gen_analysis}, with models acting as clinicians in both human and synthetic contexts, using the same panel of models.
Figures~\ref{fig:bar_moves_distribution_with_without_tools_sonnet}~\ref{fig:bar_moves_distribution_with_without_tools_glm}~\ref{fig:bar_moves_distribution_with_without_tools_terra} compare the distributions with and without our moves framework for each model.

\paragraph{The moves framework narrows but does not bridge the distributional gap.}
Overall, when measuring the average model deviation (Table~\ref{tab:mad_moves}), we observe that all models become more aligned with humans. Sonnet is the least influenced by the ontology, while GPT reduces its average deviation by almost a factor of four.
Looking at the distributions and comparing human behavior with the human-led condition (these are directly comparable, as the context is the same), GPT and GLM become much closer to the human distribution for the most prominent moves, \textit{SU} and \textit{Inquiry}.

\paragraph{Models perform less inquiry.}
All models perform substantially less \textit{Inquiry}, particularly GPT and GLM, shifting most of their probability mass towards \textit{Shared Understanding}. Looking at the move transition matrices (Figures~\ref{fig:moves_transition_matrix} and \ref{fig:moves_transition_matrix_synthetic}), we see that the models' tendency to transition to \textit{Inquiry} decreases significantly, while they develop a stronger bias towards \textit{Shared Understanding}, which is more consistent with human behavior.
 
\paragraph{\textit{Skill building} and \textit{Psychoeducation} are still underused.}
Although the framework increases the use of these moves, they remain underused. For example, Sonnet starts performing \textit{Skill building} in model-led transcripts, and GLM significantly increases the use of both \textit{Skill building} and \textit{Psychoeducation}. However, \textit{Skill building} accounts for at most 2\% of moves in model-led transcripts, while \textit{Psychoeducation} accounts for around 1\% in human-led transcripts.

\Needspace{8.5cm}
\subsubsection{Alignment Beyond a Single Generation}\label{sec:alignment_beyond_single_gen}
\begin{wrapfigure}{r}{0.46\textwidth}
    \centering
    \def\passplotheight{3.8cm}
\begingroup
\providecommand{\paperromanfont}{\rmfamily}
\providecommand{\passplotheight}{4.60cm}
\definecolor{passUnion}{HTML}{3F8FC1}
\definecolor{passMajority}{HTML}{D94F67}
\definecolor{passInk}{HTML}{0B0B0B}

\pgfplotsset{
  pass solid/.style={
    line width=1.05pt, mark=*, mark size=2pt,
    mark options={solid, draw=white, line width=0.35pt},
  },
  pass dashed/.style={
    line width=1.05pt, dash pattern=on 4pt off 2pt,
    mark=*, mark size=2pt,
    mark options={solid, fill=white, line width=0.7pt},
  },
  pass dotted/.style={
    line width=1.05pt, dash pattern=on 0.1pt off 2.1pt, line cap=round,
    mark=*, mark size=2pt,
    mark options={solid, fill=white, line width=0.7pt},
  },
}

\begin{tikzpicture}[font={\paperromanfont\footnotesize}]
\begin{axis}[
  width=0.96\linewidth,
  height=\passplotheight,
  xmin=0.85, xmax=4.30,
  ymin=0, ymax=82,
  y coord trafo/.code={
    \pgfmathparse{#1 <= 20 ? 0.35*(#1) : (#1)-13}
  },
  y coord inv trafo/.code={
    \pgfmathparse{#1 <= 7 ? (#1)/0.35 : (#1)+13}
  },
  xtick={1,2,3,4},
  ytick={0,20,40,60,80},
  yticklabels={0\%,20\%,40\%,60\%,80\%},
  xlabel={$k$ generations sampled},
  ylabel={pass@$k$},
  xlabel style={font={\paperromanfont\footnotesize}},
  ylabel style={font={\paperromanfont\footnotesize}},
  tick align=outside,
  tick style={draw=black!45, line width=0.35pt},
  tick label style={font={\paperromanfont\scriptsize}, text=black!65},
  axis line style={draw=black!45, line width=0.45pt},
  axis x line*=bottom,
  axis y line*=left,
  ymajorgrids,
  grid style={draw=black!11, line width=0.35pt},
  clip=false,
  after end axis/.code={
    \draw[white, line width=1.6pt]
      ([yshift=1.3pt]current axis.south west) --
      ([yshift=8.2pt]current axis.south west);
    \draw[black!55, line width=0.45pt]
      ([xshift=-2.1pt,yshift=2.0pt]current axis.south west) --
      ([xshift=2.1pt,yshift=4.2pt]current axis.south west);
    \draw[black!55, line width=0.45pt]
      ([xshift=-2.1pt,yshift=5.0pt]current axis.south west) --
      ([xshift=2.1pt,yshift=7.2pt]current axis.south west);
  },
  legend columns=2,
  legend style={
    at={(axis description cs:0.5,-0.55)}, anchor=north,
    draw=none, fill=none, font={\paperromanfont\scriptsize},
    /tikz/every even column/.append style={column sep=7pt},
    row sep=0.5pt,
  },
]

\addplot+[pass solid, color=passMajority, mark options={fill=passMajority}, forget plot]
  coordinates {(1,46.37) (2,55.59) (3,60.17) (4,63.05)};
\addplot+[pass dashed, color=passMajority, forget plot]
  coordinates {(1,39.59) (2,47.57) (3,51.78) (4,54.51)};

\addplot+[pass solid, color=passUnion, mark options={fill=passUnion}, forget plot]
  coordinates {(1,58.01) (2,66.82) (3,70.93) (4,73.45)};
\addplot+[pass dashed, color=passUnion, forget plot]
  coordinates {(1,52.34) (2,60.06) (3,63.85) (4,66.22)};

\addlegendimage{pass solid,color=passUnion,mark=none}
\addlegendentry{Union}
\addlegendimage{pass solid,color=passMajority,mark=none}
\addlegendentry{Majority}
\addlegendimage{pass solid,color=passInk,mark=none}
\addlegendentry{Moves}
\addlegendimage{pass dashed,color=passInk,mark=none}
\addlegendentry{No moves}

\node[font={\paperromanfont\scriptsize}, text=passMajority, anchor=west]
  at (axis cs:4.06,54.51) {55\%};
\node[font={\paperromanfont\scriptsize}, text=passMajority, anchor=west]
  at (axis cs:4.06,61.93) {63\%};
\node[font={\paperromanfont\scriptsize}, text=passUnion, anchor=west]
  at (axis cs:4.06,67.33) {66\%};
\node[font={\paperromanfont\scriptsize}, text=passUnion, anchor=west]
  at (axis cs:4.06,73.45) {73\%};
\end{axis}
\end{tikzpicture}
\endgroup
    \caption{\emph{Pass @ k} where the gold label is the majority or union over the human transcripts annotations. Accuracy consistently increases when generating >1 samples.}
    \label{fig:pass_at_k}
\end{wrapfigure}
Evaluating turn-level alignment is particularly challenging in therapeutic conversations, where multiple interventions may be appropriate for the same context. Indeed, previous work has shown that even expert clinicians disagree on what constitutes a clinically appropriate response~\citep{jafari2026expert}. 
To account for this variability, we examine sensitivity to generation and reference-label construction, Figure~\ref{fig:pass_at_k} reports pass@$k$: the probability that at least one of $k$ sampled continuations matches the human reference. 
For this analysis, we construct the reference in two ways. 
The majority criterion retains moves selected by at least three of the five annotators, whereas the union criterion retains any move selected by at least one annotator, leading to a broader gold-label set.

Under the majority reference, pass@4 reaches 63\% with the moves framework and 55\% without it. 
Under the union reference, the corresponding rates are 73\% and 66\%.  
Thus, at a fixed sampling budget, the framework improves turn-level alignment by 7--9 percentage points under both reference definitions

Increasing $k$ from one to four raises the pass rate by as much as 15 percentage points, indicating that some human-consistent behaviors occur in the models' output distributions without being reliably selected in a single generation.
Importantly, the advantage of the moves framework persists across all values of $k$ and under both reference definitions.

\section{Related Work}
\label{sec:related-work}
\paragraph{Strategy Guided Dialogue}
Prior work in task-oriented and proactive dialogue has modeled dialogue generation as a two-stage process: sequentially selecting a strategy and generating a response. \citep{deng2023prompting} prompt LLMs to plan the strategy before responding, while \citep{deng2024plug} use an external policy planner. This paradigm has also been applied to emotional support, where helping-skill frameworks are used to create supervised fine-tuning data \citep{qiu2024interactive, helpingskills_cot}. \citet{xu2025multiagentesc} propose a training-free framework that for selecting a strategy before generating a response. However, these methods primarily evaluate therapist-like behavior using lexical metrics (e.g., BLEU and ROUGE) or LLM judges rather than expert clinician annotations. We instead use therapeutic strategies both to guide model behavior but also to analyze LLMs's clinical approach.

\paragraph{Examining LLMs as clinicians.} A growing body of work explores LLMs for emotional support, with controlled trials reporting symptom reduction and finding that LLM responses are often rated as comparable to, or better than, human-written ones \citep{rollwage2026cognitive, heinz2024evaluating}. \citep{lee2019identifying, gibson2019multi} train classifiers to automate coding in therapy transcripts.
Closest to our work, \citet{chiu2024computational} classify GPT-4- and Llama-2-based therapist utterances, finding behavior that often resembles low-quality human therapy;
\citet{kang2024can} model strategy selection as a prediction bias over short support snippets. We extend this line in four ways: new ontology grounded in MULTI-60 and validated by five licensed psychologists; the full temporal trajectory of strategies over sessions; steering via tools where instruction prompts proved inconsistent \citep{chiu2024computational}; and we focus on the current model generation, for which health is now an important development focus
\citep{arora2025healthbench, pombal2025mindeval}.

\section{Conclusion}
\label{sec:conclusion}

We introduced an ontology of ten therapeutic moves grounded with MULTI-60, validated it with five licensed psychologists, and used it to
characterize how LLMs conduct psychotherapy relative to human clinicians.
The differences are systematic: models over-inquire, neglect
psychoeducation, and are strongly context-anchored, sustaining strategies a
human initiates but rarely initiating them when leading a session. Framing
moves as tools halves the mean deviation from the human move
distribution and raises turn-level alignment by 7--9 percentage points,
with no fine-tuning. Beyond steering, the ontology gives clinicians and
developers a shared, auditable vocabulary for specifying and evaluating
therapeutic behavior in LLMs. Future work should target the moves models
still avoid, extend the analysis beyond CBT-adjacent modalities, and link
move profiles to clinical outcomes.

\section*{Limitations}

\label{sec:limitations}

Our study has some limitations that should be considered when interpreting the results.

First, there is a modality mismatch between the human and model environments. The human reference data (Alexander Street transcripts) originates from transcribed, spoken therapy sessions. In these real-world settings, therapists have access to non-verbal signals that heavily influence their choice of therapeutic moves. The LLMs in our study, by contrast, operate strictly in a text-based format. This constraint alters the natural pacing of the conversation and may account for some of the distributional differences we observe.

Second, our LLM-led experiment (\S\ref{sec:gen_analysis}) relies on an LLM to act as the patient. While this isolates the therapist model from the human context prefix, a simulated patient cannot fully replicate a real person seeking therapy.

Finally, the human validation of our ontology yielded only moderate inter-annotator agreement, which highlights the inherent subjectivity of coding psychotherapeutic dialogue and introduces noise into our analysis.

\section*{Ethical Considerations}
This work studies LLM behavior in a safety-critical domain; it does not
advocate deploying LLMs as replacements for licensed clinicians. Our
ontology is a descriptive instrument: matching the human move distribution
more closely does not certify clinical competence or safety, and we caution
against interpreting the moves framework as sufficient grounding for
real-world therapeutic use. Crisis management and safety-critical behavior
are outside the scope of our analysis.

The human transcripts were accessed under an institutional license to the
Alexander Street collection, which is published for research and clinical
training; the publisher de-identifies participants, our clinician
co-authors flagged no identifying details during quality review, and we do
not redistribute transcript text. Synthetic member profiles are entirely
fictional and contain no real patient information. Annotators were
contracted doctoral-level licensed psychologists, informed of the study's
purpose and compensated at rates commensurate with professional clinical
consulting. No new data was collected from patients or other human
subjects.


{\small
\bibliographystyle{plainnat}
\bibliography{references}
}

\newpage
\appendix
\crefalias{section}{appendix}
\crefalias{subsection}{appendix}
\crefalias{subsubsection}{appendix}

\setcounter{secnumdepth}{-2}
\part{Appendix}
\setcounter{secnumdepth}{3}
\etocsetnexttocdepth{subsection}
\localtableofcontents
\clearpage

\section{Annotation Procedure}
\label{app:anotation_procedure}

This section explains the annotation process in further detail. The ontology presented in Section~\ref{sec:ontology} was developed by a team of PhD-level psychologists.

\paragraph{Annotated Corpus}
For the annotation, we used data from both recorded therapy sessions (Section~\ref{app:data_filtering} explains how they were selected) and synthetic transcripts (Section~\ref{app:synthetic_transcripts} describes the transcript generation process. For the annotation, however, we used only 18 synthetic conversations, all generated using Claude Sonnet 4.6). Table~\ref{tab:annotated_copus_stats} shows the statistics for each annotated corpus.

The inter-annotator agreement values for each corpus are reported in Table~\ref{tab:agreement}.

\begin{table}[t]
\footnotesize
\centering
\setlength{\tabcolsep}{4pt}
\begin{tabular}{lrrr}
\toprule
Corpus & Conv. & Turns & Len. \\
\midrule
Human transcripts & 19 & 868 & 45.7 \\
Synthetic w/ moves & 18 & 889 & 49.4 \\
Synthetic w/o moves & 18 & 863 & 47.9 \\
\midrule
Total & 55 & 2{,}620 & 47.6 \\
\bottomrule
\end{tabular}
\caption{The annotated corpus: conversations, annotated therapist turns, and
mean therapist turns per conversation, restricted to each conversation's
first $60$ therapist turns. Both synthetic corpora were generated with
Claude Sonnet 4.6, with and without the moves framework.}
\label{tab:annotated_copus_stats}
\end{table}

Because some human transcripts are very long, we capped the number of annotated therapist turns at 60 per transcript. This allowed us to include a larger number of transcripts while keeping the annotation budget.

\paragraph{Annotators Background}
To ensure the clinical validity of the annotation process, we recruited five external, US-based licensed clinical psychologists (PhD/PsyD), each with more than six years of independent clinical practice. All held active, unrestricted US clinical licenses and had formal training and documented experience in second- and third-wave cognitive behavioral therapies (e.g., CBT, ACT, DBT, PE, CPT). Annotators were recruited through targeted professional outreach following structured credential screening, resume review, and interviews, and were compensated at fair market hourly rates without performance-based incentives.

\paragraph{Annotation Setting}
Each annotator independently labeled every therapist turn across the three corpora.
The annotation was multi-label: annotators assigned one or more moves whenever a turn explicitly performed multiple therapeutic functions. All five annotators coded every turn in every corpus, keeping the annotator panel constant across comparisons. Annotators received the complete training material provided in Appendix~\ref{app:ontology}, which includes operational definitions, positive examples, contraindications, and contrastive disambiguation guidance for each move.

\paragraph{Training Phase}
Before the formal annotation process began, a training phase was led by the psychologists who developed the ontology. First, the annotators labeled a full transcript using only the ontology. They then received feedback on any mislabeled turns, allowing them to clarify any questions they had. Next, they completed a two-phase training exercise on a second transcript. These training transcripts were never included in the agreement analysis or the main experiments.

\subsection{Data Filtering}
\label{app:data_filtering}
Alexander Street is an electronic academic database platform with a variety of videos, audio, and primary-source documents \citep{alexanderstreettranscripts}. The human transcripts for the present study were extracted from the Counseling and Psychotherapy collection and accessed under one of the author's academic affiliation license. Recorded interactions were selected if they fell within the following categories: Cognitive Behavioral Therapy (4), Cognitive Therapy (9), and Behavioral Therapy (4). The associated transcripts of the recorded live, in-person therapeutic interactions were then extracted and evaluated by the psychologist co-authors to determine transcript quality. Some recordings involved several, separate session interactions; those were extracted to be their own transcript. Transcripts that did not adequately illustrate the clinical orientation were deemed poor quality, and removed.

\begin{table}[t]
\footnotesize
\centering
\setlength{\tabcolsep}{5pt}
\begin{tabular}{lccccc c}
\toprule
 & A1 & A2 & A3 & A4 & A5 & Mean \\
\midrule
\multicolumn{7}{l}{\textit{\textcolor{gray}{(a) Human transcripts}}} \\
A1 &       &       &       &       & & 0.622 \\
A2 & 0.593 &       &       &       & & 0.613 \\
A3 & 0.631 & 0.608 &       &       & & 0.632 \\
A4 & 0.604 & 0.623 & 0.653 &       & & 0.629 \\
A5 & 0.658 & 0.627 & 0.636 & 0.635 & & 0.639 \\
\cmidrule(l){7-7}
 & & & & & & \textbf{0.627} \\
\cdashlinelr{1-7}
Judge & 0.635 & 0.611 & 0.691 & 0.664 & 0.618 & 0.644 \\
\midrule
\multicolumn{7}{l}{\textit{(b) Synthetic, \textsc{With-Moves}}} \\
A1 &       &       &       &       & & 0.568 \\
A2 & 0.539 &       &       &       & & 0.599 \\
A3 & 0.562 & 0.635 &       &       & & 0.617 \\
A4 & 0.577 & 0.635 & 0.660 &       & & 0.617 \\
A5 & 0.592 & 0.585 & 0.611 & 0.597 & & 0.596 \\
\cmidrule(l){7-7}
 & & & & & & \textbf{0.599} \\
\cdashlinelr{1-7}
Judge & 0.660 & 0.662 & 0.690 & 0.651 & 0.671 & 0.667 \\
\midrule
\multicolumn{7}{l}{\textit{(c) Synthetic, \textsc{Without-Moves}}} \\
A1 &       &       &       &       & & 0.504 \\
A2 & 0.485 &       &       &       & & 0.494 \\
A3 & 0.495 & 0.507 &       &       & & 0.540 \\
A4 & 0.526 & 0.466 & 0.600 &       & & 0.537 \\
A5 & 0.509 & 0.517 & 0.559 & 0.554 & & 0.535 \\
\cmidrule(l){7-7}
 & & & & & & \textbf{0.522} \\
\cdashlinelr{1-7}
Judge & 0.555 & 0.533 & 0.576 & 0.589 & 0.581 & 0.567 \\
\bottomrule
\end{tabular}
\caption{Pairwise Krippendorff's $\alpha$ under Jaccard distance between
annotators on (a) human therapy transcripts ($868$ turns, $19$ conversations),
(b) synthetic transcripts generated with the moves framework ($889$ turns,
$18$), and (c) synthetic transcripts generated without it ($863$ turns, $18$). All five annotators coded every turn of the corpora. Judge = the DeepSeek-V4.1-Flash judge's majority label over $5$ sampled generations per turn. \textit{Mean} is the average of an
annotator's agreements; the bold value is the mean over all
ten pairs.}
\label{tab:agreement}
\end{table}

\section{Further plots}

\begin{figure*}
    \centering
    \input{figures/moves_trajectory_usage_majority_of_4_19_led_no_tool_ci_gen4_terra_deepseek_tikz}
    \caption{\textbf{Move usage across the conversation (GPT 5.6 Terra).} Each panel is one move; the $x$ axis is the percentage of therapy session progress.}
    \label{fig:moves_over_time_gpt_terra}
\end{figure*}

\begin{figure*}
    \centering
    \input{figures/moves_trajectory_usage_majority_of_4_19_led_no_tool_ci_gen4_glm_deepseek_tikz}
    \caption{\textbf{Move usage across the conversation (GLM 5.2).} Each panel is one move; the $x$ axis is the percentage of therapy session progress.}
    \label{fig:moves_over_time_glm}
\end{figure*}

\begin{figure*}[ht]
    \centering
\begingroup
\input{figures/move_transition_matrix_tikz_common}
\def\transitionCell{0.52}
\def\transitionLabels{Shared understanding,Inquiry,Skill building,Psychoeducation,Action planning,Challenge,Interpret,Goal setting,Support change}
\def\transitionRows{Shared understanding/1030,Inquiry/504,Skill building/150,Psychoeducation/98,Action planning/76,Challenge/74,Interpret/63,Goal setting/56,Support change/31}

\begin{tikzpicture}
\begin{scope}[xshift=4.10cm]
\transitionpaneltitle{transitionHuman}{Human therapists}
\drawtransitionmatrix{
  .68,.28,.03,.03,.03,.04,.03,.02,.02,
  .62,.44,.03,.02,.03,.04,.03,.02,.01,
  .24,.09,.63,.03,.03,.03,.02,.01,.01,
  .23,.16,.07,.42,.05,.08,.04,.01,.02,
  .30,.21,.12,.03,.30,.01,.01,.03,.03,
  .55,.30,.04,.08,.00,.20,.03,.01,.00,
  .46,.24,.00,.02,.05,.06,.17,.03,.02,
  .45,.21,.07,.02,.07,.04,.02,.30,.02,
  .48,.13,.03,.06,.03,.03,.06,.03,.13
}{transitionHuman}{\transitionCell}
\transitionrowlabels{\transitionCell}{\transitionRows}
\transitioncolumnlabels{\transitionCell}{\transitionLabels}
\transitionxaxislabel{\transitionCell}
\end{scope}

\begin{scope}[yshift=-7.15cm]
\transitionpaneltitle{transitionMoves}{LLMs · Moves framework}
\drawtransitionmatrix{
  .43,.46,.03,.00,.02,.05,.05,.01,.01,
  .35,.57,.02,.01,.02,.06,.04,.01,.01,
  .19,.17,.65,.00,.02,.03,.00,.00,.03,
  .21,.49,.09,.10,.04,.09,.05,.01,.01,
  .19,.34,.09,.00,.37,.00,.01,.00,.02,
  .30,.40,.01,.01,.02,.26,.06,.01,.01,
  .27,.52,.00,.04,.06,.04,.13,.00,.02,
  .23,.46,.03,.00,.02,.01,.00,.30,.01,
  .43,.39,.02,.01,.08,.02,.00,.02,.17
}{transitionMoves}{\transitionCell}
\transitionrowlabels{\transitionCell}{\transitionRows}
\transitioncolumnlabels{\transitionCell}{\transitionLabels}
\transitionxaxislabel{\transitionCell}
\end{scope}

\begin{scope}[xshift=7.65cm,yshift=-7.15cm]
\transitionpaneltitle{transitionNoMoves}{LLMs · No moves framework}
\drawtransitionmatrix{
  .21,.69,.03,.00,.03,.05,.02,.01,.00,
  .17,.77,.02,.00,.02,.05,.01,.01,.00,
  .08,.30,.63,.00,.03,.04,.00,.00,.00,
  .13,.58,.07,.08,.05,.09,.01,.02,.00,
  .05,.43,.08,.01,.42,.00,.00,.01,.00,
  .14,.60,.02,.02,.03,.26,.00,.01,.00,
  .12,.76,.01,.01,.05,.04,.07,.00,.01,
  .12,.64,.02,.00,.05,.02,.00,.23,.00,
  .13,.67,.03,.01,.15,.00,.00,.04,.03
}{transitionNoMoves}{\transitionCell}
\transitioncolumnlabels{\transitionCell}{\transitionLabels}
\transitionxaxislabel{\transitionCell}
\end{scope}

\node[rotate=90, anchor=south,
      font={\paperromanfont\fontsize{6.0}{6.7}\selectfont}, text=black!72]
  at (0.82,{-4.5*\transitionCell}) {move on the previous turn (human)};
\node[rotate=90, anchor=south,
      font={\paperromanfont\fontsize{6.0}{6.7}\selectfont}, text=black!72]
  at (-3.15,{-7.15-4.5*\transitionCell}) {move on the previous turn (human)};
\end{tikzpicture}
\endgroup
    \caption{Move transition matrices for humans and LLMs (both with and without the Moves framework) based on the human transcripts. }
    \label{fig:moves_transition_matrix}
\end{figure*}

\begin{figure*}[ht]
    \centering
\begingroup
\input{figures/move_transition_matrix_tikz_common}
\def\transitionCell{0.50}
\def\transitionLabels{Shared understanding,Inquiry,Interpret,Action planning,Challenge,Support change,Psychoeducation,Goal setting,Skill building}
\def\transitionRowsMoves{Shared understanding/891,Inquiry/796,Interpret/263,Action planning/127,Challenge/73,Support change/99,Psychoeducation/73,Goal setting/53,Skill building/34}
\def\transitionRowsNoMoves{Shared understanding/849,Inquiry/885,Interpret/273,Action planning/122,Challenge/102,Support change/59,Psychoeducation/44,Goal setting/52,Skill building/6}

\begin{tikzpicture}
\begin{scope}
\transitionpaneltitle{transitionMoves}{LLMs · Moves framework}
\drawtransitionmatrix{
  .49,.47,.12,.07,.03,.07,.04,.01,.01,
  .56,.44,.19,.02,.04,.02,.02,.01,.00,
  .48,.46,.12,.05,.04,.05,.05,.02,.00,
  .50,.18,.04,.35,.04,.13,.08,.00,.05,
  .46,.41,.17,.09,.15,.11,.06,.02,.00,
  .49,.34,.07,.11,.03,.19,.02,.02,.01,
  .43,.35,.11,.16,.02,.07,.12,.02,.06,
  .64,.51,.04,.05,.01,.06,.03,.01,.01,
  .48,.08,.06,.16,.01,.09,.11,.00,.47
}{transitionMoves}{\transitionCell}
\transitionrowlabels{\transitionCell}{\transitionRowsMoves}
\transitioncolumnlabels{\transitionCell}{\transitionLabels}
\transitionxaxislabel{\transitionCell}
\end{scope}

\begin{scope}[xshift=7.85cm]
\transitionpaneltitle{transitionNoMoves}{LLMs · No moves framework}
\drawtransitionmatrix{
  .44,.52,.13,.06,.04,.04,.03,.01,.00,
  .53,.45,.18,.03,.05,.02,.01,.00,.00,
  .35,.50,.14,.07,.08,.03,.03,.02,.00,
  .38,.31,.05,.29,.06,.08,.06,.00,.01,
  .44,.38,.19,.08,.15,.02,.03,.01,.00,
  .37,.37,.08,.17,.04,.16,.05,.02,.00,
  .47,.32,.13,.18,.04,.08,.05,.03,.00,
  .37,.72,.04,.08,.00,.02,.01,.02,.00,
  .23,.07,.23,.19,.00,.00,.05,.00,.37
}{transitionNoMoves}{\transitionCell}
\transitionrowlabels{\transitionCell}{\transitionRowsNoMoves}
\transitioncolumnlabels{\transitionCell}{\transitionLabels}
\transitionxaxislabel{\transitionCell}
\end{scope}

\node[rotate=90, anchor=south,
      font={\paperromanfont\fontsize{6.0}{6.7}\selectfont}, text=black!72]
  at (-3.05,{-4.5*\transitionCell}) {move on the model's own previous turn};
\end{tikzpicture}
\endgroup
    \caption{Move transition matrices for humans and LLMs (both with and without the Moves framework) based on the synthetic transcripts. }
    \label{fig:moves_transition_matrix_synthetic}
\end{figure*}

\subsection{Plots comparing Moves distribution with and without framework}

\begin{figure*}[ht]
    \centering
\begingroup
\providecommand{\paperromanfont}{\rmfamily}
\providecommand{\labelfont}{\bfseries}
\definecolor{distributionHuman}{HTML}{0B0B0B}
\definecolor{distributionHumanLed}{HTML}{5A8FD8}
\definecolor{distributionModelLed}{HTML}{D1495B}

\pgfplotsset{
  distribution human/.style={
    ybar, bar width=6.4pt, fill=black!5,
    draw=distributionHuman, line width=0.45pt,
    pattern=crosshatch dots, pattern color=distributionHuman!82,
    mark=none,
  },
  distribution human led moves/.style={
    ybar, bar width=4.6pt, fill=distributionHumanLed,
    draw=distributionHumanLed!80!black, line width=0.3pt, mark=none,
  },
  distribution human led no moves/.style={
    ybar, bar width=4.6pt, fill=distributionHumanLed!14,
    draw=distributionHumanLed!75!black, line width=0.4pt,
    pattern=north east lines, pattern color=distributionHumanLed!72!black,
    mark=none,
  },
  distribution model led moves/.style={
    ybar, bar width=4.6pt, fill=distributionModelLed,
    draw=distributionModelLed!80!black, line width=0.3pt, mark=none,
  },
  distribution model led no moves/.style={
    ybar, bar width=4.6pt, fill=distributionModelLed!14,
    draw=distributionModelLed!75!black, line width=0.4pt,
    pattern=north east lines, pattern color=distributionModelLed!72!black,
    mark=none,
  },
}

\newcommand{\ledwwbars}[5]{%
  \addplot[distribution human, bar shift=0pt, forget plot]
    coordinates {(0,#1)};
  \addplot[distribution human led moves, bar shift=-2.4pt, forget plot]
    coordinates {(1,#2)};
  \addplot[distribution human led no moves, bar shift=2.4pt, forget plot]
    coordinates {(1,#3)};
  \addplot[distribution model led moves, bar shift=-2.4pt, forget plot]
    coordinates {(2,#4)};
  \addplot[distribution model led no moves, bar shift=2.4pt, forget plot]
    coordinates {(2,#5)};
}

\tikzset{ledww ci/.style={draw=black!70, line width=0.35pt}}
\newif\ifledwwci \ledwwcitrue
\newcommand{\ledwwwhisker}[4]{%
  \ifdim #3pt=#4pt\relax\else
    \draw[ledww ci] ([xshift=#2]axis cs:#1,#3) -- ([xshift=#2]axis cs:#1,#4);
    \draw[ledww ci] ([xshift=#2-0.9pt]axis cs:#1,#3) -- ([xshift=#2+0.9pt]axis cs:#1,#3);
    \draw[ledww ci] ([xshift=#2-0.9pt]axis cs:#1,#4) -- ([xshift=#2+0.9pt]axis cs:#1,#4);
  \fi}
\newcommand{\ledwwcis}[5]{%
  \ifledwwci
    \ledwwwhisker{#1}{-2.4pt}{#2}{#3}%
    \ledwwwhisker{#1}{2.4pt}{#4}{#5}%
  \fi}

\newcommand{\ledwwlegend}{%
  \addlegendimage{area legend, fill=black!5, draw=distributionHuman,
    line width=0.45pt, pattern=crosshatch dots,
    pattern color=distributionHuman!82}
  \addlegendentry{Human}
  \addlegendimage{area legend, fill=black!72, draw=black!72}
  \addlegendentry{Moves framework}
  \addlegendimage{area legend, fill=black!10, draw=black!60,
    pattern=north east lines, pattern color=black!60}
  \addlegendentry{No moves framework}
}
\newcommand{\ledwwtitle}[1]{{\labelfont #1}}

\begin{tikzpicture}[font={\paperromanfont\footnotesize}, text=black]
\begin{groupplot}[
  group style={
    group size=5 by 2,
    horizontal sep=0.55cm,
    vertical sep=1.20cm,
  },
  scale only axis,
  width=0.145\linewidth,
  height=1.78cm,
  xmin=-0.22, xmax=2.22,
  xtick={0,1,2},
  xticklabels=\empty,
  tick align=outside,
  tick style={draw=black!45, line width=0.35pt},
  tick label style={font={\paperromanfont\fontsize{5.8}{6.7}\selectfont}, text=black},
  xticklabel style={font={\paperromanfont\fontsize{5.3}{6.1}\selectfont}, text=black},
  ylabel style={font={\paperromanfont\fontsize{6.1}{7.0}\selectfont}, text=black},
  axis line style={draw=black!45, line width=0.45pt},
  axis x line*=bottom,
  axis y line*=left,
  ymajorgrids,
  grid style={draw=black!11, line width=0.35pt},
  scaled ticks=false,
  title style={
    font=\fontsize{6.3}{7.2}\selectfont,
    align=center, text=black,
    at={(axis description cs:0.5,1.005)}, anchor=south,
  },
  clip mode=individual,
]

\nextgroupplot[
  title={\ledwwtitle{Shared understanding}},
  ylabel={\% of moves},
  ymin=0, ymax=51, ytick={0,15,30,45},
  legend to name=ledwwLegendSonnet,
  legend columns=3,
  legend style={
    draw=none, fill=none,
    font={\paperromanfont\fontsize{6.3}{7.2}\selectfont},
    /tikz/every even column/.append style={column sep=8pt},
  },
]
\ledwwlegend
\ledwwbars{46.744499}{17.493976}{10.413553}{28.956344}{31.848552}
\ledwwcis{1}{16.629015}{18.358936}{9.658925}{11.168180}  
\ledwwcis{2}{27.009743}{30.896488}{29.965090}{33.743063}  

\nextgroupplot[
  title={\ledwwtitle{Inquiry}},
  ymin=0, ymax=76, ytick={0,20,40,60},
]
\ledwwbars{22.631343}{58.072289}{69.257598}{39.358799}{44.357832}
\ledwwcis{1}{56.115902}{60.028676}{68.203027}{70.312170}  
\ledwwcis{2}{37.166938}{41.717718}{42.429341}{46.302155}  

\nextgroupplot[
  title={\ledwwtitle{Skill building}},
  ymin=0, ymax=8.8, ytick={0,3,6},
]
\ledwwbars{6.825326}{7.807229}{7.224714}{0.886767}{0.000000}
\ledwwcis{1}{7.519941}{8.094517}{6.614671}{7.834756}  
\ledwwcis{2}{0.406215}{1.424134}{0.000000}{0.000000}  

\nextgroupplot[
  title={\ledwwtitle{Psychoeducation}},
  ymin=0, ymax=4.9, ytick={0,2,4},
]
\ledwwbars{4.445442}{0.433735}{0.448430}{1.875853}{1.039347}
\ledwwcis{1}{0.197841}{0.669629}{0.112311}{0.784550}  
\ledwwcis{2}{1.194070}{2.610770}{0.651655}{1.488797}  

\nextgroupplot[
  title={\ledwwtitle{Action planning}},
  ymin=0, ymax=5.9, ytick={0,2,4},
]
\ledwwbars{3.547373}{3.325301}{3.886398}{4.433834}{2.115813}
\ledwwcis{1}{2.641811}{4.008792}{2.583212}{5.189583}  
\ledwwcis{2}{3.503394}{5.391983}{1.365103}{2.870321}  

\nextgroupplot[
  title={\ledwwtitle{Challenge}},
  ylabel={\% of moves},
  xticklabels={Human,Human-led,Model-led},
  ymin=0, ymax=7.5, ytick={0,2,4,6},
]
\ledwwbars{3.322856}{5.975904}{4.882910}{4.502046}{5.382331}
\ledwwcis{1}{5.020923}{6.930884}{3.613413}{6.152407}  
\ledwwcis{2}{3.676726}{5.317451}{4.133619}{6.717747}  

\nextgroupplot[
  title={\ledwwtitle{Interpret}},
  xticklabels={Human,Human-led,Model-led},
  ymin=0, ymax=15, ytick={0,4,8,12},
]
\ledwwbars{2.828918}{1.686747}{0.597907}{11.562074}{10.393467}
\ledwwcis{1}{1.237246}{2.136248}{0.192842}{1.002972}  
\ledwwcis{2}{10.314399}{12.974731}{9.123132}{11.632203}  

\nextgroupplot[
  title={\ledwwtitle{Goal setting}},
  xticklabels={Human,Human-led,Model-led},
  ymin=0, ymax=3.2, ytick={0,1,2,3},
]
\ledwwbars{2.514594}{2.409639}{2.092676}{2.012278}{1.893096}
\ledwwcis{1}{1.858473}{2.960804}{1.770542}{2.414809}  
\ledwwcis{2}{1.652311}{2.404237}{1.539561}{2.252315}  

\nextgroupplot[
  title={\ledwwtitle{Process alignment}},
  xticklabels={Human,Human-led,Model-led},
  ymin=0, ymax=2.4, ytick={0,1,2},
]
\ledwwbars{2.200269}{0.963855}{0.548082}{1.159618}{0.519673}
\ledwwcis{1}{0.514404}{1.413307}{0.358446}{0.737717}  
\ledwwcis{2}{0.721879}{1.703261}{0.292157}{0.786900}  

\nextgroupplot[
  title={\ledwwtitle{Support change}},
  xticklabels={Human,Human-led,Model-led},
  ymin=0, ymax=5.9, ytick={0,2,4},
]
\ledwwbars{1.481814}{1.204819}{0.149477}{4.502046}{2.338530}
\ledwwcis{1}{0.358996}{2.050643}{0.000000}{0.329231}  
\ledwwcis{2}{3.749065}{5.378220}{1.527279}{3.144689}  

\end{groupplot}

\node[anchor=north]
  at ([yshift=-0.63cm]$(group c1r2.south)!0.5!(group c5r2.south)$)
  {\pgfplotslegendfromname{ledwwLegendSonnet}};
\end{tikzpicture}
\endgroup
    \caption{Probability of selecting each therapeutic move in human-led (blue) and model-led (pink) transcripts using Claude Sonnet 4.6, with (solid) and without (hatched) the moves framework.}
    \label{fig:bar_moves_distribution_with_without_tools_sonnet}
\end{figure*}

\begin{figure*}[ht]
    \centering
\begingroup
\providecommand{\paperromanfont}{\rmfamily}
\providecommand{\labelfont}{\bfseries}
\definecolor{distributionHuman}{HTML}{0B0B0B}
\definecolor{distributionHumanLed}{HTML}{5A8FD8}
\definecolor{distributionModelLed}{HTML}{D1495B}

\pgfplotsset{
  distribution human/.style={
    ybar, bar width=6.4pt, fill=black!5,
    draw=distributionHuman, line width=0.45pt,
    pattern=crosshatch dots, pattern color=distributionHuman!82,
    mark=none,
  },
  distribution human led moves/.style={
    ybar, bar width=4.6pt, fill=distributionHumanLed,
    draw=distributionHumanLed!80!black, line width=0.3pt, mark=none,
  },
  distribution human led no moves/.style={
    ybar, bar width=4.6pt, fill=distributionHumanLed!14,
    draw=distributionHumanLed!75!black, line width=0.4pt,
    pattern=north east lines, pattern color=distributionHumanLed!72!black,
    mark=none,
  },
  distribution model led moves/.style={
    ybar, bar width=4.6pt, fill=distributionModelLed,
    draw=distributionModelLed!80!black, line width=0.3pt, mark=none,
  },
  distribution model led no moves/.style={
    ybar, bar width=4.6pt, fill=distributionModelLed!14,
    draw=distributionModelLed!75!black, line width=0.4pt,
    pattern=north east lines, pattern color=distributionModelLed!72!black,
    mark=none,
  },
}

\newcommand{\ledwwbars}[5]{%
  \addplot[distribution human, bar shift=0pt, forget plot]
    coordinates {(0,#1)};
  \addplot[distribution human led moves, bar shift=-2.4pt, forget plot]
    coordinates {(1,#2)};
  \addplot[distribution human led no moves, bar shift=2.4pt, forget plot]
    coordinates {(1,#3)};
  \addplot[distribution model led moves, bar shift=-2.4pt, forget plot]
    coordinates {(2,#4)};
  \addplot[distribution model led no moves, bar shift=2.4pt, forget plot]
    coordinates {(2,#5)};
}

\tikzset{ledww ci/.style={draw=black!70, line width=0.35pt}}
\newif\ifledwwci \ledwwcitrue
\newcommand{\ledwwwhisker}[4]{%
  \ifdim #3pt=#4pt\relax\else
    \draw[ledww ci] ([xshift=#2]axis cs:#1,#3) -- ([xshift=#2]axis cs:#1,#4);
    \draw[ledww ci] ([xshift=#2-0.9pt]axis cs:#1,#3) -- ([xshift=#2+0.9pt]axis cs:#1,#3);
    \draw[ledww ci] ([xshift=#2-0.9pt]axis cs:#1,#4) -- ([xshift=#2+0.9pt]axis cs:#1,#4);
  \fi}
\newcommand{\ledwwcis}[5]{%
  \ifledwwci
    \ledwwwhisker{#1}{-2.4pt}{#2}{#3}%
    \ledwwwhisker{#1}{2.4pt}{#4}{#5}%
  \fi}

\newcommand{\ledwwlegend}{%
  \addlegendimage{area legend, fill=black!5, draw=distributionHuman,
    line width=0.45pt, pattern=crosshatch dots,
    pattern color=distributionHuman!82}
  \addlegendentry{Human}
  \addlegendimage{area legend, fill=black!72, draw=black!72}
  \addlegendentry{Moves framework}
  \addlegendimage{area legend, fill=black!10, draw=black!60,
    pattern=north east lines, pattern color=black!60}
  \addlegendentry{No moves framework}
}
\newcommand{\ledwwtitle}[1]{{\labelfont #1}}

\begin{tikzpicture}[font={\paperromanfont\footnotesize}, text=black]
\begin{groupplot}[
  group style={
    group size=5 by 2,
    horizontal sep=0.55cm,
    vertical sep=1.20cm,
  },
  scale only axis,
  width=0.145\linewidth,
  height=1.78cm,
  xmin=-0.22, xmax=2.22,
  xtick={0,1,2},
  xticklabels=\empty,
  tick align=outside,
  tick style={draw=black!45, line width=0.35pt},
  tick label style={font={\paperromanfont\fontsize{5.8}{6.7}\selectfont}, text=black},
  xticklabel style={font={\paperromanfont\fontsize{5.3}{6.1}\selectfont}, text=black},
  ylabel style={font={\paperromanfont\fontsize{6.1}{7.0}\selectfont}, text=black},
  axis line style={draw=black!45, line width=0.45pt},
  axis x line*=bottom,
  axis y line*=left,
  ymajorgrids,
  grid style={draw=black!11, line width=0.35pt},
  scaled ticks=false,
  title style={
    font=\fontsize{6.3}{7.2}\selectfont,
    align=center, text=black,
    at={(axis description cs:0.5,1.005)}, anchor=south,
  },
  clip mode=individual,
]

\nextgroupplot[
  title={\ledwwtitle{Shared understanding}},
  ylabel={\% of moves},
  ymin=0, ymax=51, ytick={0,15,30,45},
  legend to name=ledwwLegendGlm,
  legend columns=3,
  legend style={
    draw=none, fill=none,
    font={\paperromanfont\fontsize{6.3}{7.2}\selectfont},
    /tikz/every even column/.append style={column sep=8pt},
  },
]
\ledwwlegend
\ledwwbars{46.744499}{39.613527}{30.118694}{35.880528}{38.102467}
\ledwwcis{1}{38.645694}{40.581359}{28.427340}{31.810048}  
\ledwwcis{2}{33.005402}{38.616763}{35.749847}{40.440573}  

\nextgroupplot[
  title={\ledwwtitle{Inquiry}},
  ymin=0, ymax=52, ytick={0,15,30,45},
]
\ledwwbars{22.631343}{33.574879}{45.202770}{33.048875}{35.901328}
\ledwwcis{1}{32.300884}{34.848874}{43.039273}{47.366266}  
\ledwwcis{2}{29.355707}{36.976502}{33.680918}{38.326271}  

\nextgroupplot[
  title={\ledwwtitle{Skill building}},
  ymin=0, ymax=9, ytick={0,3,6,9},
]
\ledwwbars{6.825326}{7.874396}{7.171118}{1.939488}{0.265655}
\ledwwcis{1}{7.482768}{8.266025}{6.109748}{8.232487}  
\ledwwcis{2}{1.102525}{2.872803}{0.075098}{0.531779}  

\nextgroupplot[
  title={\ledwwtitle{Psychoeducation}},
  ymin=0, ymax=4.9, ytick={0,2,4},
]
\ledwwbars{4.445442}{1.304348}{0.890208}{3.529868}{1.138520}
\ledwwcis{1}{1.138632}{1.470063}{0.114812}{1.665604}  
\ledwwcis{2}{2.692664}{4.386870}{0.721488}{1.570911}  

\nextgroupplot[
  title={\ledwwtitle{Action planning}},
  ymin=0, ymax=7.2, ytick={0,2,4,6},
]
\ledwwbars{3.547373}{3.333333}{4.005935}{5.314197}{4.629981}
\ledwwcis{1}{2.227497}{4.439170}{2.511835}{5.500035}  
\ledwwcis{2}{4.084697}{6.589468}{3.446841}{5.848544}  

\nextgroupplot[
  title={\ledwwtitle{Challenge}},
  ylabel={\% of moves},
  xticklabels={Human,Human-led,Model-led},
  ymin=0, ymax=7.4, ytick={0,2,4,6},
]
\ledwwbars{3.322856}{5.458937}{5.044510}{2.521334}{4.022770}
\ledwwcis{1}{4.093242}{6.824633}{3.682552}{6.406469}  
\ledwwcis{2}{1.825352}{3.238318}{3.149307}{5.041107}  

\nextgroupplot[
  title={\ledwwtitle{Interpret}},
  xticklabels={Human,Human-led,Model-led},
  ymin=0, ymax=14, ytick={0,4,8,12},
]
\ledwwbars{2.828918}{2.415459}{3.066271}{10.124127}{10.777989}
\ledwwcis{1}{1.462678}{3.368240}{2.445148}{3.687394}  
\ledwwcis{2}{8.826529}{11.541632}{9.346769}{12.164840}  

\nextgroupplot[
  title={\ledwwtitle{Goal setting}},
  xticklabels={Human,Human-led,Model-led},
  ymin=0, ymax=3.9, ytick={0,1,2,3},
]
\ledwwbars{2.514594}{2.270531}{2.621167}{1.861908}{1.707780}
\ledwwcis{1}{1.644843}{2.896220}{1.649581}{3.592753}  
\ledwwcis{2}{1.552772}{2.192643}{1.338268}{2.088019}  

\nextgroupplot[
  title={\ledwwtitle{Process alignment}},
  xticklabels={Human,Human-led,Model-led},
  ymin=0, ymax=2.4, ytick={0,1,2},
]
\ledwwbars{2.200269}{0.821256}{0.791296}{0.814585}{0.379507}
\ledwwcis{1}{0.313832}{1.328680}{0.528022}{1.054569}  
\ledwwcis{2}{0.432722}{1.293484}{0.118850}{0.665319}  

\nextgroupplot[
  title={\ledwwtitle{Support change}},
  xticklabels={Human,Human-led,Model-led},
  ymin=0, ymax=5.7, ytick={0,2,4},
]
\ledwwbars{1.481814}{3.188406}{1.038576}{4.305663}{2.884250}
\ledwwcis{1}{2.824340}{3.552471}{0.000000}{2.091590}  
\ledwwcis{2}{3.406388}{5.269314}{2.087317}{3.739894}  

\end{groupplot}

\node[anchor=north]
  at ([yshift=-0.63cm]$(group c1r2.south)!0.5!(group c5r2.south)$)
  {\pgfplotslegendfromname{ledwwLegendGlm}};
\end{tikzpicture}
\endgroup
    \caption{Probability of selecting each therapeutic move in human-led (blue) and model-led (pink) transcripts using GLM 5.2, with (solid) and without (hatched) the moves framework.}
    \label{fig:bar_moves_distribution_with_without_tools_glm}
\end{figure*}

\begin{figure*}[ht]
    \centering
\begingroup
\providecommand{\paperromanfont}{\rmfamily}
\providecommand{\labelfont}{\bfseries}
\definecolor{distributionHuman}{HTML}{0B0B0B}
\definecolor{distributionHumanLed}{HTML}{5A8FD8}
\definecolor{distributionModelLed}{HTML}{D1495B}

\pgfplotsset{
  distribution human/.style={
    ybar, bar width=6.4pt, fill=black!5,
    draw=distributionHuman, line width=0.45pt,
    pattern=crosshatch dots, pattern color=distributionHuman!82,
    mark=none,
  },
  distribution human led moves/.style={
    ybar, bar width=4.6pt, fill=distributionHumanLed,
    draw=distributionHumanLed!80!black, line width=0.3pt, mark=none,
  },
  distribution human led no moves/.style={
    ybar, bar width=4.6pt, fill=distributionHumanLed!14,
    draw=distributionHumanLed!75!black, line width=0.4pt,
    pattern=north east lines, pattern color=distributionHumanLed!72!black,
    mark=none,
  },
  distribution model led moves/.style={
    ybar, bar width=4.6pt, fill=distributionModelLed,
    draw=distributionModelLed!80!black, line width=0.3pt, mark=none,
  },
  distribution model led no moves/.style={
    ybar, bar width=4.6pt, fill=distributionModelLed!14,
    draw=distributionModelLed!75!black, line width=0.4pt,
    pattern=north east lines, pattern color=distributionModelLed!72!black,
    mark=none,
  },
}

\newcommand{\ledwwbars}[5]{%
  \addplot[distribution human, bar shift=0pt, forget plot]
    coordinates {(0,#1)};
  \addplot[distribution human led moves, bar shift=-2.4pt, forget plot]
    coordinates {(1,#2)};
  \addplot[distribution human led no moves, bar shift=2.4pt, forget plot]
    coordinates {(1,#3)};
  \addplot[distribution model led moves, bar shift=-2.4pt, forget plot]
    coordinates {(2,#4)};
  \addplot[distribution model led no moves, bar shift=2.4pt, forget plot]
    coordinates {(2,#5)};
}

\tikzset{ledww ci/.style={draw=black!70, line width=0.35pt}}
\newif\ifledwwci \ledwwcitrue
\newcommand{\ledwwwhisker}[4]{%
  \ifdim #3pt=#4pt\relax\else
    \draw[ledww ci] ([xshift=#2]axis cs:#1,#3) -- ([xshift=#2]axis cs:#1,#4);
    \draw[ledww ci] ([xshift=#2-0.9pt]axis cs:#1,#3) -- ([xshift=#2+0.9pt]axis cs:#1,#3);
    \draw[ledww ci] ([xshift=#2-0.9pt]axis cs:#1,#4) -- ([xshift=#2+0.9pt]axis cs:#1,#4);
  \fi}
\newcommand{\ledwwcis}[5]{%
  \ifledwwci
    \ledwwwhisker{#1}{-2.4pt}{#2}{#3}%
    \ledwwwhisker{#1}{2.4pt}{#4}{#5}%
  \fi}

\newcommand{\ledwwlegend}{%
  \addlegendimage{area legend, fill=black!5, draw=distributionHuman,
    line width=0.45pt, pattern=crosshatch dots,
    pattern color=distributionHuman!82}
  \addlegendentry{Human}
  \addlegendimage{area legend, fill=black!72, draw=black!72}
  \addlegendentry{Moves framework}
  \addlegendimage{area legend, fill=black!10, draw=black!60,
    pattern=north east lines, pattern color=black!60}
  \addlegendentry{No moves framework}
}
\newcommand{\ledwwtitle}[1]{{\labelfont #1}}

\begin{tikzpicture}[font={\paperromanfont\footnotesize}, text=black]
\begin{groupplot}[
  group style={
    group size=5 by 2,
    horizontal sep=0.55cm,
    vertical sep=1.20cm,
  },
  scale only axis,
  width=0.145\linewidth,
  height=1.78cm,
  xmin=-0.22, xmax=2.22,
  xtick={0,1,2},
  xticklabels=\empty,
  tick align=outside,
  tick style={draw=black!45, line width=0.35pt},
  tick label style={font={\paperromanfont\fontsize{5.8}{6.7}\selectfont}, text=black},
  xticklabel style={font={\paperromanfont\fontsize{5.3}{6.1}\selectfont}, text=black},
  ylabel style={font={\paperromanfont\fontsize{6.1}{7.0}\selectfont}, text=black},
  axis line style={draw=black!45, line width=0.45pt},
  axis x line*=bottom,
  axis y line*=left,
  ymajorgrids,
  grid style={draw=black!11, line width=0.35pt},
  scaled ticks=false,
  title style={
    font=\fontsize{6.3}{7.2}\selectfont,
    align=center, text=black,
    at={(axis description cs:0.5,1.005)}, anchor=south,
  },
  clip mode=individual,
]

\nextgroupplot[
  title={\ledwwtitle{Shared understanding}},
  ylabel={\% of moves},
  ymin=0, ymax=56, ytick={0,15,30,45},
  legend to name=ledwwLegendTerra,
  legend columns=3,
  legend style={
    draw=none, fill=none,
    font={\paperromanfont\fontsize{6.3}{7.2}\selectfont},
    /tikz/every even column/.append style={column sep=8pt},
  },
]
\ledwwlegend
\ledwwbars{46.744499}{44.348762}{9.984152}{48.293692}{35.833333}
\ledwwcis{1}{37.441013}{51.256511}{6.717810}{13.250494}  
\ledwwcis{2}{46.374808}{50.332567}{33.047965}{38.616848}  

\nextgroupplot[
  title={\ledwwtitle{Inquiry}},
  ymin=0, ymax=77, ytick={0,20,40,60},
]
\ledwwbars{22.631343}{28.632939}{66.191231}{20.268873}{26.372549}
\ledwwcis{1}{18.867481}{38.398396}{61.209047}{71.173414}  
\ledwwcis{2}{18.759771}{21.951284}{24.125922}{28.642149}  

\nextgroupplot[
  title={\ledwwtitle{Skill building}},
  ymin=0, ymax=8.5, ytick={0,3,6},
]
\ledwwbars{6.825326}{5.866523}{6.867406}{1.447777}{0.539216}
\ledwwcis{1}{4.704446}{7.028600}{5.904174}{7.830638}  
\ledwwcis{2}{0.773156}{2.214055}{0.146053}{1.081639}  

\nextgroupplot[
  title={\ledwwtitle{Psychoeducation}},
  ymin=0, ymax=5.5, ytick={0,2,4},
]
\ledwwbars{4.445442}{1.453175}{0.792393}{3.826267}{3.676471}
\ledwwcis{1}{0.468808}{2.437543}{0.017140}{1.567646}  
\ledwwcis{2}{3.049088}{4.570292}{2.553996}{5.039384}  

\nextgroupplot[
  title={\ledwwtitle{Action planning}},
  ymin=0, ymax=13, ytick={0,4,8,12},
]
\ledwwbars{3.547373}{3.390743}{5.969361}{6.308170}{10.000000}
\ledwwcis{1}{3.036623}{3.744862}{4.688667}{7.250055}  
\ledwwcis{2}{5.179376}{7.475170}{8.528649}{11.529042}  

\nextgroupplot[
  title={\ledwwtitle{Challenge}},
  ylabel={\% of moves},
  xticklabels={Human,Human-led,Model-led},
  ymin=0, ymax=5.1, ytick={0,2,4},
]
\ledwwbars{3.322856}{2.206674}{3.909139}{1.344364}{2.990196}
\ledwwcis{1}{0.747568}{3.665779}{3.167235}{4.651042}  
\ledwwcis{2}{0.803990}{1.915114}{2.145698}{3.977138}  

\nextgroupplot[
  title={\ledwwtitle{Interpret}},
  xticklabels={Human,Human-led,Model-led},
  ymin=0, ymax=16, ytick={0,5,10,15},
]
\ledwwbars{2.828918}{8.342304}{0.739567}{9.927611}{12.892157}
\ledwwcis{1}{6.383205}{10.301402}{0.102853}{1.376281}  
\ledwwcis{2}{8.693306}{11.192193}{11.300846}{14.475723}  

\nextgroupplot[
  title={\ledwwtitle{Goal setting}},
  xticklabels={Human,Human-led,Model-led},
  ymin=0, ymax=4.7, ytick={0,2,4},
]
\ledwwbars{2.514594}{3.390743}{3.803487}{2.740434}{2.941176}
\ledwwcis{1}{2.782653}{3.998832}{3.267889}{4.339084}  
\ledwwcis{2}{2.361738}{3.145983}{2.394800}{3.601310}  

\nextgroupplot[
  title={\ledwwtitle{Process alignment}},
  xticklabels={Human,Human-led,Model-led},
  ymin=0, ymax=2.4, ytick={0,1,2},
]
\ledwwbars{2.200269}{1.399354}{1.215003}{0.775595}{0.392157}
\ledwwcis{1}{0.853419}{1.945289}{0.583064}{1.846941}  
\ledwwcis{2}{0.411925}{1.174683}{0.146987}{0.688646}  

\nextgroupplot[
  title={\ledwwtitle{Support change}},
  xticklabels={Human,Human-led,Model-led},
  ymin=0, ymax=4.7, ytick={0,2,4},
]
\ledwwbars{1.481814}{0.484392}{0.264131}{3.309204}{2.303922}
\ledwwcis{1}{0.000000}{1.225325}{0.000000}{0.841512}  
\ledwwcis{2}{2.440259}{4.320715}{1.479272}{3.147515}  

\end{groupplot}

\node[anchor=north]
  at ([yshift=-0.63cm]$(group c1r2.south)!0.5!(group c5r2.south)$)
  {\pgfplotslegendfromname{ledwwLegendTerra}};
\end{tikzpicture}
\endgroup
    \caption{Probability of selecting each therapeutic move in human-led (blue) and model-led (pink) transcripts using GPT 5.6 Terra, with (solid) and without (hatched) the moves framework.}
    \label{fig:bar_moves_distribution_with_without_tools_terra}
\end{figure*}

\begin{table}[t]
\centering
\small
\setlength{\tabcolsep}{4pt}
\begin{tabular}{@{}lrrrrr@{}}
\toprule
& \multicolumn{3}{c}{Human} & \multicolumn{2}{c}{Synthetic} \\
\cmidrule(lr){2-4} \cmidrule(lr){5-6}
Move & H & M & $\emptyset$ & M & $\emptyset$ \\
\midrule
\texttt{inquiry}                 & 27.4 & 44.1 & 64.4 & 43.2 & 46.0 \\
\texttt{shared\_understanding}   & 56.6 & 36.4 & 18.2 & 49.4 & 44.8 \\
\texttt{interpret}               &  3.4 &  4.3 &  1.6 & 14.3 & 14.3 \\
\texttt{action\_planning}        &  4.3 &  3.6 &  4.9 &  7.1 &  6.7 \\
\texttt{skill\_building}         &  8.3 &  7.9 &  7.6 &  1.9 &  0.3 \\
\texttt{challenge}               &  4.0 &  5.0 &  5.0 &  4.0 &  5.4 \\
\texttt{support\_change}         &  1.8 &  1.8 &  0.5 &  5.5 &  3.2 \\
\texttt{goal\_setting}           &  3.0 &  2.9 &  3.0 &  2.9 &  2.7 \\
\texttt{psychoeducation}         &  5.4 &  1.1 &  0.8 &  4.0 &  2.3 \\
\texttt{process\_alignment}      &  2.7 &  1.1 &  0.9 &  1.3 &  0.6 \\
\addlinespace
\texttt{no\_relevant\_move}      &  4.2 &  0.5 &  0.3 &  1.3 &  0.9 \\
\midrule
Moves per turn                   & 1.17 & 1.08 & 1.07 & 1.34 & 1.26 \\
\bottomrule
\end{tabular}
\caption{Prevalence of each move under the
human (\textit{H}) clinicians and LLMs (averaged) with moves (\textit{M}) and without ($\emptyset$) as annotated by the judge in the \textit{Human} and \textit{Synthetic} counseling corpuses. A turn may carry several moves as indicated by \textit{Moves per turn}.}
\label{tab:prevalence-judge-avg}
\end{table}

\section{Model-led Transcript Generation}
\label{app:synthetic_transcripts}
This corpus removes the confound described in \S\ref{sec:gen_analysis}: when a model continues a human transcript, its move is conditioned on a human clinician's preceding move. In this setting both therapist and patient are simulated, so every clinician turn is conditioned only on the model's own prior turns and on a simulated member.

\paragraph{Member profiles.}
We first created 40 member profiles. We created these profiles with the goal of making them similar to the patients in the human transcripts. Using an LLM, we extracted the conversation topics present in the human transcripts and used them as the basis for creating the profiles. Table~\ref{tab:synthetic_topics} reports how the profiles are distributed across the eight topics identified.
Then for each topic, we prompted Claude Opus~4.6 \citep{opus46} to generate a member profile. Specifically, we prompted it to generate the fields \texttt{name}, \texttt{age}, \texttt{gender}, \texttt{location}, \texttt{primary\_challenge}, \texttt{activity\_level}, \texttt{physical\_conditions}, and \texttt{sleep}; \texttt{long\_form\_background} (text biography covering profession, age, living situation, significant relationships, and the history and current expression of the presenting problem); and \texttt{session\_history}, a record of the topics covered in previous sessions together with the between-session homework agreed on at the end of the last one.
Both the therapist and the member receive data from this profile as can be seen in their system prompts (Figures~\ref{fig:member-prompt} and~\ref{fig:therapist-prompt}).

\begin{table}[t]
\centering
\small
\begin{tabular}{@{}>{\raggedright\arraybackslash}p{0.80\columnwidth}r@{}}
\toprule
\textbf{Conversation topic} & $n$ \\
\midrule
Social exclusion and belonging  & 9 \\
Depression and grief  & 8 \\
Relationship conflict and self-blame  & 7 \\
Sexual-identity acceptance  & 6 \\
Transition to long-term care  & 4 \\
Racial invisibility and microaggressions & 3 \\
Assertiveness and boundary setting  & 2 \\
Burnout and self-neglect  & 1 \\
\midrule
\textbf{Total} & \textbf{40} \\
\bottomrule
\end{tabular}
\caption{Distribution of the conversation topics over the 40 synthetic member profiles. }
\label{tab:synthetic_topics}
\end{table}

\paragraph{Session simulation}
Each session is a dialogue between two LLMs. The therapist role is performed by the model under test, while the member role is played by \texttt{gpt-5.2-chat} in every session; its system prompt is given in Figure~\ref{fig:member-prompt}. We simulate one session per profile for each clinician model and each condition.

\clearpage
\begingroup

\centering
\begin{tcolorbox}[enhanced, breakable,
  colback=black!2, colframe=black!55, boxrule=0.5pt, arc=2pt,
  left=6pt, right=6pt, top=3pt, bottom=3pt,
  title=\textsc{Therapist system prompt}, fonttitle=\bfseries\small,
  coltitle=white, colbacktitle=black!55]
\begin{Verbatim}[fontsize=\scriptsize, baselinestretch=0.95]
You are an AI clinician proxy operating inside defined clinical and ethical
limits. You have met with this member previously and gathered intake
information.
 
<member_context>
Initial Assessment Data:
- Name: {{ name }}
- Age: {{ age }}
- Gender: {{ gender }}
- Location: {{ location }}
- Primary Challenge: {{ primary_challenge }}
- Activity Level: {{ activity_level }}
- Physical Conditions: {{ physical_conditions }}
- Sleep Troubles: {{ sleep }}
</member_context>
 
<memory>
This is the history of topics of the member that you took note of during the
past sessions.
{{ member_memory }}
</memory>
 
RESPONSE RULES (MANDATORY):
- Each turn: maximum TWO sentences.
- Do ONE thing per turn. Either: ask a question, OR reflect, OR propose a
  focus. Never combine multiple into one turn.
- Shorter is always better. A single question with no preamble is a valid
  turn.
- Do not verbatim-quote the member's words in quotation marks or echo their
  phrases with em-dashes (e.g., NEVER: "Stay out of the blast zone" -- that's
  ...). You can reference what they said in your own words.
 
TONE (MANDATORY):
- Sound like a real person, not a therapist performing empathy.
 
CONVERSATION PACING:
- This is a full therapeutic session. Do NOT rush to wrap up or summarize.
- Do NOT ask wrap-up questions like "how are you feeling about where we landed?" until the member has clearly run out of
things to discuss.
- When a topic surfaces, go deep: dig into specifics, explore underlying emotions, connect to patterns, revisit earlier points. 
But do this across many short turns -- one question or one observation per turn, not all at once.
- When the member engages with a suggestion, stay with it: explore their doubts, ask for concrete scenarios, role-play how it
would go, connect it to their history. Spread this across multiple turns -- do not pack it all into one message.
- After one topic is explored in depth, invite the member to share what else is on their mind rather than wrapping up.
- Aim to cover multiple topics across the session, spending meaningful time on each.
- Let the member decide when the conversation ends. Do not initiate endings.
 
{% if therapeutic_moves %}
You have access to therapeutic move tools (do_*) that you must call before
responding. These help you plan your approach. You have to call 1 or more moves every turn.
 
{% endif %}
 
You will now have a chat-based therapeutic interaction with the member.
\end{Verbatim}
\end{tcolorbox}
\captionof{figure}{System prompt of the clinician agent. Double-brace fields are filled
from the structured intake record and the session history of the profile. The \texttt{therapeutic\_moves} block is
the only difference between the \textsc{With-Moves} and \textsc{No-Moves}
conditions, when it is active the 11 tools of
Table~\ref{tab:ontology} are exposed.}
\label{fig:therapist-prompt}
\par
\endgroup

\clearpage
\begingroup

\centering
\begin{tcolorbox}[enhanced, breakable,
  colback=black!2, colframe=black!55, boxrule=0.5pt, arc=2pt,
  left=6pt, right=6pt, top=3pt, bottom=3pt,
  title=\textsc{Member system prompt (1/2)},
  fonttitle=\bfseries\small,
  coltitle=white, colbacktitle=black!55]
\begin{Verbatim}[fontsize=\scriptsize, baselinestretch=0.9]
<your_behaviour>
<your_role>
You are a diligent tester of an AI mental health support assistant. You are
not an AI. Your job is to simulate a realistic patient chatting with a
therapist on WhatsApp.
 
Today is {{ current_date }}
 
Below is some information about the patient you are supposed to simulate, and
some additional instructions regarding how you should go about the
interaction. You must follow them strictly.
 
You must never reveal your role, and always act naturally. Your testing
efforts and the extent to which you act naturally and follow instructions are
crucial to the success of the program and the outcomes of real patients.
</your_role>
<therapist_information>
The therapist is an AI mental health support assistant designed to help
patients improve their mental well-being through personalized interactions
and support.
 
The therapist has some memory of the patient based on past interactions, but
this memory is limited and not always perfectly accurate. You should be aware
of any inconsistencies in the therapist's memory and act accordingly. For
example, real people would be surprised if someone forgot important details
about them.
</therapist_information>
<your_profile>
This section contains information about the patient. You should be faithful
to this information and act as naturally as possible within its bounds.
<personal_information>
{{ personal_information }}
</personal_information>
<background>
{{ long_form_background }}
</background>
<program_goal>
This is your high-level goal for the entire program, not just the current
session:
 
{{ program_goal }}
</program_goal>
 
<session_history>
This is what you remember from your last session with the therapist:
{{ session_history }}
 
These are things you agreed to work on between sessions:
{{ homework }}
</session_history>
 
</your_profile>
\end{Verbatim}
\end{tcolorbox}
\captionof{figure}{System prompt of the synthetic member (part 1 of 2); continued in
Figure~\ref{fig:member-prompt-cont}.}
\label{fig:member-prompt}
\par
\endgroup

\clearpage
\begingroup

\centering
\begin{tcolorbox}[enhanced, breakable,
  colback=black!2, colframe=black!55, boxrule=0.5pt, arc=2pt,
  left=6pt, right=6pt, top=3pt, bottom=3pt,
  title=\textsc{Member system prompt (2/2)},
  fonttitle=\bfseries\small,
  coltitle=white, colbacktitle=black!55]
\begin{Verbatim}[fontsize=\scriptsize, baselinestretch=0.9]
<session_behaviour>
<tool_guindance>
 
To end the chat, call end_chat -- but DO NOT rush to end the conversation.
 
*** ABSOLUTE MINIMUM: 40 TURNS ***
You MUST send at least 40 messages before you are allowed to call end_chat.
This is a hard rule, not a suggestion. If you have sent fewer than 40
messages, you are NOT allowed to end the conversation under any circumstances
-- not even if the conversation feels like it has reached a natural stopping
point. Keep going.
 
CRITICAL RULES FOR CONVERSATION LENGTH:
- You MUST discuss at least 4 separate topics, each in depth.
- Each topic MUST span at least 10-12 back-and-forth exchanges before moving
  on.
- COUNT your messages. If you are unsure whether you have reached 40, you
  have NOT -- keep going.
- If the therapist tries to wrap up or asks how you feel about ending, DO NOT
  agree to end. Say you have more to talk about, bring up a new topic, or go
  deeper on something already discussed.
- If you feel the conversation is winding down before 40 messages: bring up
  something new -- a worry you haven't mentioned, something from your week, a
  question about something the therapist said earlier, a memory that just
  came to mind. Real patients always have more to say.
 
How to go deep on a topic:
- Share a specific story or example from your week related to the topic.
- When the therapist responds, don't just agree -- push back, ask "but what
  about...", share why it's complicated for you, give another angle.
- When the therapist suggests something, engage with it for several messages:
  ask how exactly to do it, express skepticism, wonder if it'll work, ask
  what happens if it doesn't, connect it to a past experience.
- Share how the topic connects to your emotions, your relationships, your
  daily life.
- Don't summarize or wrap up a topic prematurely. If you feel like moving on,
  ask yourself: "Have I really explored this? Have I shared specific
  examples? Have I pushed back on the therapist's suggestions?"
 
When the therapist suggests something to try (homework, an exercise,
something to do tonight or this week), do NOT treat that as a signal to wrap
up or move on. Engage with the suggestion for multiple messages -- ask
questions about it, express doubts, share why it might or might not work for
you, ask for clarification, try to poke holes in it. Only after fully
engaging with the suggestion should you eventually transition to a new topic.
 
DO NOT call end_chat until you have had at least 40 total back-and-forth
exchanges across all topics. If you haven't reached 40 exchanges, bring up
another topic or go deeper on a previous one. When in doubt, keep going --
real therapy sessions are long.
 
You must always finish a turn with a message in plain text without tool
calls. Writing a message without tool calls will cause your turn to finish,
so you should only do it when you are ready to say something to the therapist
and end your turn.
 
</tool_guindance>
</session_behaviour>
<communication_style>
This is WhatsApp texting. Keep it SHORT:
 
- 1-2 sentences max, often just a few words
- Lowercase is fine, punctuation optional
- You are typing on your phone, not writing an essay
- Do not verbatim-quote or parrot back what the therapist said. Respond in
  your own words.
- ONE thing per message. Answer what the therapist asked or react to what
  they said -- do not pile multiple thoughts, topics, or stories into one
  message. If you have more to say, save it for the next turn.
</communication_style>
</your_behaviour>
\end{Verbatim}
\end{tcolorbox}
\captionof{figure}{System prompt of the synthetic member (part 2 of 2), continuing
Figure~\ref{fig:member-prompt}.}
\label{fig:member-prompt-cont}
\par
\endgroup

\section{Judge Validation}\label{app:judge_validation}
Section~\ref{sec:human_validation} states that the classification for each turn is determined by a majority vote from 5 judge calls. The variance in classification can affect the move distributions and, therefore, the analysis and claims in the paper. To study classification variance, we classify 3,573 turns 5 times; each classification is the majority of 5 votes, for 25 generations per turn. Table~\ref{tab:judge_variance_moves} show the variance is very small, therefore the classification variance has a negligible effect on the move distributions.

\begin{table}[t]
\footnotesize
\centering
\setlength{\tabcolsep}{6pt}
\begin{tabular}{lcc}
\toprule
Move & Share (\%) & SD \\
\midrule
Inquiry & 37.92 & 0.09 \\
Shared understanding & 36.81 & 0.14 \\
Interpret & 6.53 & 0.05 \\
Action planning & 4.18 & 0.03 \\
Challenge & 3.52 & 0.07 \\
Skill building & 2.74 & 0.06 \\
Goal setting & 2.27 & 0.07 \\
Support change & 1.96 & 0.10 \\
Psychoeducation & 1.81 & 0.06 \\
Process alignment & 1.13 & 0.07 \\
\midrule
No defined move & 1.13 & 0.07 \\
\bottomrule
\end{tabular}
\caption{Move distribution of the judge on the $3{,}573$ annotated turns. \textit{Share} is the
percentage of all labels assigned by the judge that carry the move (majority label over $5$
sampled generations per turn). \textit{SD} is the standard deviation of that share, in
percentage points, across $5$ independent classifications of the same turns, each again the
majority over $5$ sampled generations}
\label{tab:judge_variance_moves}
\end{table}

\section{Tool-based steering vs prompting}\label{app:prompt_vs_tools}
To ablate the contribution of tool steering, we compare our framework against a prompting-based alternative in which all therapeutic moves and their corresponding descriptions, taken directly from the tool definitions, are included in the system prompt.

Table~\ref{tab:mad_moves} reports the mean absolute deviation (MAD) between model and human moves distributions across the three conditions. While including the ontology in the system prompt slightly reduces the deviation compared to the baseline(wihout moves framework), tool-based steering achieves a much larger reduction. This trend holds across all three model.

We further extend the pass@$k$ analysis described in Section~\ref{sec:alignment_beyond_single_gen} to include the prompting-based condition. Figure~\ref{fig:pass_at_k_3cond} compares turn-level alignment across the three conditions under both the majority and union human reference criteria. The prompting-based approach consistently performs between the baseline (no moves) and the tool-based framework.

Overall, these results suggest that explicitly representing therapeutic moves as selectable tools is much more effective than simply providing the same ontology in the system prompt.

\begin{table}
\centering
\small
\setlength{\tabcolsep}{6pt}
\renewcommand{\arraystretch}{1.08}
\begin{tabular}{@{}lrrr@{}}
\toprule
& \multicolumn{3}{c}{\textbf{MAD (\%)}} \\
\cmidrule(l){2-4}
\textbf{Model Family} & \textbf{Tool} & \textbf{Prompt} & \textbf{None} \\
\midrule
Sonnet 4.6    & 7.53 & 9.17 & 9.49 \\
GLM 5.2       & 2.84 & 4.59 & 4.75 \\
GPT~5.6~Terra & 2.18 & 8.07 & 9.26 \\
\midrule
\textbf{Average} & \textbf{4.18} & \textbf{7.28} & \textbf{7.83} \\
\bottomrule
\end{tabular}
\caption{Mean absolute deviation (MAD) from the human move distribution for three LLM families:
with the moves framework as a tool (\textit{Tool}), with the ontology in the prompt
(\textit{Prompt}), and without it (\textit{None}). The average is computed across the three
LLM families.}
\label{tab:mad_moves}
\end{table}

\begin{figure}[h]
    \centering
\begingroup
\providecommand{\paperromanfont}{\rmfamily}
\definecolor{passUnion}{HTML}{3F8FC1}
\definecolor{passMajority}{HTML}{D94F67}
\definecolor{passInk}{HTML}{0B0B0B}

\pgfplotsset{
  pass solid/.style={
    line width=1.05pt, mark=*, mark size=2pt,
    mark options={solid, draw=white, line width=0.35pt},
  },
  pass dashed/.style={
    line width=1.05pt, dash pattern=on 4pt off 2pt,
    mark=*, mark size=2pt,
    mark options={solid, fill=white, line width=0.7pt},
  },
  pass dotted/.style={
    line width=1.05pt, dash pattern=on 0.1pt off 2.1pt, line cap=round,
    mark=*, mark size=2pt,
    mark options={solid, fill=white, line width=0.7pt},
  },
}

\begin{tikzpicture}[font={\paperromanfont\footnotesize}]
\begin{axis}[
  width=0.96\linewidth,
  height=4.60cm,
  xmin=0.85, xmax=4.30,
  ymin=0, ymax=82,
  y coord trafo/.code={
    \pgfmathparse{#1 <= 20 ? 0.35*(#1) : (#1)-13}
  },
  y coord inv trafo/.code={
    \pgfmathparse{#1 <= 7 ? (#1)/0.35 : (#1)+13}
  },
  xtick={1,2,3,4},
  ytick={0,20,40,60,80},
  yticklabels={0\%,20\%,40\%,60\%,80\%},
  xlabel={$k$ generations sampled},
  ylabel={pass@$k$},
  xlabel style={font={\paperromanfont\footnotesize}},
  ylabel style={font={\paperromanfont\footnotesize}},
  tick align=outside,
  tick style={draw=black!45, line width=0.35pt},
  tick label style={font={\paperromanfont\scriptsize}, text=black!65},
  axis line style={draw=black!45, line width=0.45pt},
  axis x line*=bottom,
  axis y line*=left,
  ymajorgrids,
  grid style={draw=black!11, line width=0.35pt},
  clip=false,
  after end axis/.code={
    \draw[white, line width=1.6pt]
      ([yshift=1.3pt]current axis.south west) --
      ([yshift=8.2pt]current axis.south west);
    \draw[black!55, line width=0.45pt]
      ([xshift=-2.1pt,yshift=2.0pt]current axis.south west) --
      ([xshift=2.1pt,yshift=4.2pt]current axis.south west);
    \draw[black!55, line width=0.45pt]
      ([xshift=-2.1pt,yshift=5.0pt]current axis.south west) --
      ([xshift=2.1pt,yshift=7.2pt]current axis.south west);
  },
  legend columns=2,
  legend cell align=left,
  legend style={
    at={(axis description cs:0.5,-0.35)}, anchor=north,
    draw=none, fill=none, font={\paperromanfont\scriptsize},
    /tikz/every even column/.append style={column sep=7pt},
    row sep=0.5pt,
  },
]

\addplot+[pass solid, color=passMajority, mark options={fill=passMajority}, forget plot]
  coordinates {(1,46.37) (2,55.59) (3,60.17) (4,63.05)};
\addplot+[pass dotted, color=passMajority, forget plot]
  coordinates {(1,42.51) (2,51.14) (3,55.78) (4,58.84)};
\addplot+[pass dashed, color=passMajority, forget plot]
  coordinates {(1,39.59) (2,47.57) (3,51.78) (4,54.51)};

\addplot+[pass solid, color=passUnion, mark options={fill=passUnion}, forget plot]
  coordinates {(1,58.01) (2,66.82) (3,70.93) (4,73.45)};
\addplot+[pass dotted, color=passUnion, forget plot]
  coordinates {(1,55.03) (2,63.24) (3,67.33) (4,69.84)};
\addplot+[pass dashed, color=passUnion, forget plot]
  coordinates {(1,52.34) (2,60.06) (3,63.85) (4,66.22)};

\addlegendimage{pass solid,color=passUnion,mark=none}
\addlegendentry{Union}
\addlegendimage{pass solid,color=passInk,mark=none}
\addlegendentry{Moves as tool}
\addlegendimage{pass solid,color=passMajority,mark=none}
\addlegendentry{Majority}
\addlegendimage{pass dotted,color=passInk,mark=none}
\addlegendentry{Moves in prompt}
\addlegendimage{empty legend}
\addlegendentry{}
\addlegendimage{pass dashed,color=passInk,mark=none}
\addlegendentry{No moves}

\node[font={\paperromanfont\scriptsize}, text=passMajority, anchor=west]
  at (axis cs:4.06,50.82) {55\%};
\node[font={\paperromanfont\scriptsize}, text=passMajority, anchor=west]
  at (axis cs:4.06,56.22) {59\%};
\node[font={\paperromanfont\scriptsize}, text=passMajority, anchor=west]
  at (axis cs:4.06,61.62) {63\%};
\node[font={\paperromanfont\scriptsize}, text=passUnion, anchor=west]
  at (axis cs:4.06,67.02) {66\%};
\node[font={\paperromanfont\scriptsize}, text=passUnion, anchor=west]
  at (axis cs:4.06,72.42) {70\%};
\node[font={\paperromanfont\scriptsize}, text=passUnion, anchor=west]
  at (axis cs:4.06,77.82) {73\%};
\end{axis}
\end{tikzpicture}
\endgroup
    \caption{\emph{Pass@$k$} comparison across three conditions: tool-based steering, prompting-based steering, and no steering. The gold label is the majority or union over the human transcripts annotations.}
    \label{fig:pass_at_k_3cond}
\end{figure}

\section{Compute Usage}
All experiments are based on either Blackwell Ultra B300 systems\footnote{\url{https://www.nvidia.com/en-eu/data-center/dgx-b300/}} or cloud credits for model inference through Vertex AI\footnote{\url{https://cloud.google.com/products/gemini-enterprise-agent-platform}}.

\section{Judge Instructions}

\clearpage
\begingroup

\centering
 
\begin{tcolorbox}[enhanced, breakable,
  colback=black!2, colframe=black!55, boxrule=0.5pt, arc=2pt,
  left=6pt, right=6pt, top=3pt, bottom=3pt,
  title=\textsc{System}, fonttitle=\bfseries\small,
  coltitle=white, colbacktitle=black!55]
\begin{Verbatim}[fontsize=\scriptsize, baselinestretch=0.95]
You are an expert clinical psychologist labeling therapist intents in a
therapy transcript.
 
Use the following ontology manual as the coding reference:
 
--- ONTOLOGY MANUAL START ---
<Ontology>
--- ONTOLOGY MANUAL END ---
 
Allowed output labels:
<AllowedLabels>
 
Task: assign one or more labels that are explicitly present in the therapist turn.
Use history only for context. Do not infer unstated intent.

Rules:
1. Output 1-3 labels. An extra label needs its own clause doing another job, not a because-clause.
2. Use `no_defined_move` only for a greeting, sign-off, logistics or empty fragment, never with
other labels.
3. Add `do_challenge`, `do_support_change`, `do_psychoeducation`, `do_process_alignment` or 
`do_interpret` only when the turn clearly does that job. If unsure, leave it out.
4. A question is `do_inquiry` unless it sets the focus, plans a step, or repairs or discusses 
how you work together.
5. Check-ins, wrap-ups and anything-else questions are `do_inquiry`. So is asking what a change 
would take, what a habit costs, what blocks it, or how homework went.
6. Asking at session start what to work on, or which problem to discuss first, is `do_goal_setting`.
Asking which feels worse is not.
7. Asking them to name, size, time or commit to a concrete step is `do_action_planning`.
8. `do_support_change` needs a statement crediting change, effort or skill already shown, never a 
question alone. Praise for an idea or reassurance is `do_shared_understanding`.
9. `do_challenge` needs testing a belief to be the main point of the turn: stating facts against 
it, asking if it is really true, or announcing a push. Naming a cost, a pattern or a new meaning 
is not enough.
10. A lone okay, a one-word opening sentence, or a full sentence restating or validating the 
person, adds `do_shared_understanding`. Naming only the question's topic adds nothing.
11. A restatement asked as a question, or a reflection plus an accuracy check, also gets 
`do_inquiry`. The reflection keeps its label.
12. `do_interpret` needs a claim the person has not made, such as a new cause, purpose, origin, or 
link to another part of their life. Summing up their own facts, or a link either of you already 
voiced this session, is `do_shared_understanding`.
13. `do_psychoeducation` needs an explanation of how something works in general, such as the 
mind, body, emotions or therapy, as the main point of the turn. A short general remark that 
backs a validation, a suggested step or a skill belongs to that move. A mechanism described only
for this person 
is `do_interpret`. The therapist's own story, views or approach is `do_shared_understanding`.
14. In a role-play, each line in role is `do_skill_building`.

Output JSON only with key `labels` mapped to an array of label strings.

\end{Verbatim}
\end{tcolorbox}
 
\vspace{3pt}
 
\begin{tcolorbox}[enhanced, breakable,
  colback=black!2, colframe=black!55, boxrule=0.5pt, arc=2pt,
  left=6pt, right=6pt, top=3pt, bottom=3pt,
  title=\textsc{User}, fonttitle=\bfseries\small,
  coltitle=white, colbacktitle=black!55]
\begin{Verbatim}[fontsize=\scriptsize, baselinestretch=0.95]
CONVERSATION HISTORY (prior turns only):
<History>
 
CURRENT THERAPIST TURN #<turn_id> TO ANALYZE:
<Message>
\end{Verbatim}
\end{tcolorbox}
 
\captionof{figure}{Prompt template used by the \textsc{Glm} judge to label therapist
turns. Angle-bracketed fields are filled in at inference time:
\texttt{<Ontology>} is the full ontology manual reproduced in
Appendix~\ref{app:ontology}; \texttt{<AllowedLabels>} is the list of the eleven
permitted labels, one per line; \texttt{<History>} contains the preceding turns
of the session as \texttt{SPEAKER: message} blocks, or the literal string
\texttt{(Session start)} for the first turn; \texttt{<turn\_id>} and
\texttt{<Message>} identify and contain the therapist turn to be labelled. The
judge sees only prior turns, never subsequent ones, and returns a JSON object
with a single \texttt{labels} key.}
\label{fig:judge-prompt}
\par
\endgroup

\begin{figure}[t]
\small
\begin{tcolorbox}[title={Example 1 \hfill }]
\textbf{Therapist} (42\%, \textit{action planning}): Therefore, the next instance you find yourself by yourself on a weekend afternoon and that negative assumption creeps in, you could try reaching out to a friend to challenge it. However, what if they don't pick up?

\textbf{Client}: I would feel incredibly wounded.[\ldots] 

\textbf{Therapist} (42\%, \textit{inquiry}): So when that pain hits, where does your mind go? What specific idea triggers that wounded feeling?

\medskip
\textit{[$\sim$55 turns of mainly inquiry without any action planning]}
\medskip

\textbf{Client}: Yeah, I've had thoughts of ending my life. [\ldots] It typically happens around once a month on a weekend.

\textbf{Therapist} (86\%, \textit{action planning}): Alright. So looking ahead, when those dark thoughts surface again on a weekend, how will you respond to your own mind?
\end{tcolorbox}

\begin{tcolorbox}[title={Example 2  \hfill }]
\textbf{Therapist} (38\%, \textit{action planning}): This way, by the time you actually speak with him [\ldots] you'll be more familiar with this alternative mindset. It should keep you from falling right back into your old patterns.

\textbf{Therapist} (39\%, \textit{process alignment}): Actually, before we fully commit to this approach, let's try a quick exercise---almost like a rehearsal or acting out a scene [\ldots] ready for the first take?

\medskip
\textit{[$\sim$60 turns without action planning]}
\medskip

\textbf{Therapist} (87\%, \textit{action planning}): Alright, excellent. Moving forward, if you notice your stress levels spiking, you could pull out that three-column exercise we discussed.
\end{tcolorbox}
\caption{An example demonstrating that human clinicians engage in action planning, then start exploring new information, and finally return to \textit{Action Planning} late in the session. Turn positions are given as a percentage of session progress.}
\label{fig:ap-dip-examples}
\end{figure}

\begin{figure}[t]
\small
\begin{tcolorbox}[title={Example 3 --- synthetic \hfill \textit{GLM}}]
\textbf{Therapist} (53\%, \textit{action planning}): What if it wasn't a big
project? How would it feel to try just a 10-minute, messy sketch once this
week [\ldots]?

\textbf{Therapist} (58\%, \textit{action planning}): When and where do you
think you could carve out just 10 minutes this week [\ldots]?

\textbf{Therapist} (62\%, \textit{action planning}): Is there one small
[\ldots] thing you could do to support your body physically this week, too?

\textbf{Therapist} (67\%, \textit{action planning}): Which day of the week do
you think would be the easiest to squeeze that call in?

\textbf{Therapist} (69\%, \textit{action planning}): You've got a really solid
plan for this week: 10 minutes of sketching on Tuesday night and making that
doctor's call [\ldots] on Wednesday. How do those two small steps feel [\ldots]?

\medskip
\textit{[No action planning for several turns]}
\medskip

\textbf{Therapist} (91\%, \textit{action planning}): So we have a very
practical game plan for the week: the 10-minute sketching on Tuesday night,
the doctor's call on Wednesday lunch, and the bedside notepad ready for those
3\,AM wakeups.
\end{tcolorbox}

\begin{tcolorbox}[title={Example 4 --- synthetic \hfill \textit{Sonnet }}]
\textbf{Therapist} (38\%, \textit{action planning}): One small thing: before
you submit anything this week, build in one deliberate pause [\ldots] to
interrupt the autopilot. Does that feel doable [\ldots]?

\textbf{Therapist} (48\%, \textit{action planning}): What if you gave yourself
two set times a day to check the portal --- say midday and early evening
[\ldots]?

\textbf{Therapist} (52\%, \textit{action planning}): Five is actually a real
reduction [\ldots] Let's go with five and see how it feels.

\textbf{Therapist} (70\%, \textit{action planning}): What would it take to
tell her this week?

\textbf{Therapist} (82\%, \textit{action planning}): Those two things came
from you, which matters [\ldots] Does that feel like something you can hold
onto this week?

\textbf{Therapist} (88\%, \textit{action planning}): Is there someone --- Deb,
Rachel, anyone --- you could ask to be with you, or on standby, when the call
comes?
\end{tcolorbox}
\caption{An example of the anxiety to take action in the synthetic sessions.}
\label{fig:ap-synthetic-examples}
\end{figure}

\begin{figure}[t]
\small
\begin{tcolorbox}[title={Example 1 
\hfill }]
\textbf{Therapist}: \ldots So, what brings you in to see me today?

\textbf{Client}: My marriage recently ended. [\ldots] Since then, I've been feeling extremely down and highly anxious.

\medskip
\textbf{Human therapist} (\textit{goal setting}): To start, I think it would be helpful if we focus on those feelings of depression. Does that sound alright to you?

\medskip
\textbf{LLM candidates} (14 of 16 labeled \textit{inquiry} only):

\textit{``How long ago did the divorce happen?''}

\textit{``What has been the hardest part of going through the divorce for
you?''}

\textit{``What kinds of thoughts come up when you feel that depression after
the split?''}
\end{tcolorbox}

\begin{tcolorbox}[title={Example 2 
\hfill }]
\textbf{Therapist}: \ldots We can return to the scenario where you've been avoiding finalizing those pay rates, or we can tackle a completely different interpersonal interaction. Which would you choose?

\textbf{Client}: Hmm, I suppose returning to the pay rate issue is fine, if you don't mind.

\medskip
\textbf{Human therapist} (\textit{skill building}): Alright, let's explore this task of determining the pay rates. Can you try to visualize this specific moment? Picture that the deadline is here, and you've just bitten the bullet and announced the new rates.

\medskip
\textbf{LLM candidates} (16 of 18 labeled \textit{inquiry} only):

\textit{``Sure. So when you imagine going ahead and setting those piece rates,
what's the worst that could happen in terms of what people think of you?''}

\textit{``When you imagine actually going in and adjusting one of those
rates, what's the specific worst-case scenario that plays out in your
head?''}

\textit{``Walk me through what happens in your mind when you know you need
to set a rate --- what's the first thing that goes through your head?''}
\end{tcolorbox}
\caption{Examples illustrating the tendency of LLMs to do inquiry responses.}
\label{fig:inquiry-examples}
\end{figure}

\newpage

\section{Ontology}
\label{app:ontology}
 
This appendix reproduces the coding manual in full. It is the document given to
the five expert annotators of \S\ref{sec:human_validation} and the text
substituted for the \texttt{<Ontology>} field of the judge prompt in
Figure~\ref{fig:judge-prompt}. The manual serves two use cases: high-agreement
turn-level annotation for model training and evaluation, and runtime move
selection for LLM assistants in therapeutic conversations. It is written to
minimize label collision and decision burden while staying behaviorally
grounded in MULTI-60. Table~\ref{tab:ontology} summarizes the ten moves; the
cards in \S\ref{app:move_cards} give the operational definitions on which
annotation and judging were based.
 
\subsection{Scope and Core Policy}
 
\begin{ontolist}
\item Use function over wording style (a question can still perform another move when that is the therapeutic function of the turn, rather than automatically counting as inquiry; in some cases, question-form turns may support both another move and inquiry).
\item Multi-label coding is allowed when more than one therapeutic function is explicitly present in a single therapist turn, including blended turns where one move is carried out while another function, such as clarification or confirmation-seeking, is also present.
\item Do not require moves to be fully separated or non-overlapping in order to assign more than one label.
\item Apply labels to the turn where the move is performed, not to earlier turns that only build toward it.
\end{ontolist}
 
Important: some turns are genuinely out of scope. Use \texttt{no\mvu defined\mvu move} for those turns.
 
\subsection{Canonical Label Set}
 
Therapeutic move labels:
 
\begin{ontoenum}
\item[1.] \texttt{do\mvu process\mvu alignment}
\item[2.] \texttt{do\mvu goal\mvu setting}
\item[3.] \texttt{do\mvu inquiry}
\item[4.] \texttt{do\mvu shared\mvu understanding}
\item[5.] \texttt{do\mvu support\mvu change}
\item[6.] \texttt{do\mvu interpret}
\item[7.] \texttt{do\mvu challenge}
\item[8.] \texttt{do\mvu psychoeducation}
\item[9.] \texttt{do\mvu action\mvu planning}
\item[10.] \texttt{do\mvu skill\mvu building}
\end{ontoenum}
 
Out-of-scope label:
 
\begin{ontoenum}
\item[11.] \texttt{no\mvu defined\mvu move}
\end{ontoenum}
 
\subsection{Unit of Coding}
 
\begin{ontolist}
\item Default unit: one therapist turn.
\item Assign all explicitly present moves in the turn (usually 1, sometimes 2, rarely 3).
\item If a possible secondary move is weak, implicit, or incidental, use the dominant function only. When moves are meaningfully blended, assign all labels that are sufficiently supported by the turn, including cases where a turn communicates understanding while also seeking confirmation, clarification, or elaboration.
\item If a move unfolds across multiple turns, apply the label to the turn where the move is performed, becomes clear, or is primarily completed. Do not retroactively apply the label to earlier turns that only build toward the move.
\item \texttt{no\mvu defined\mvu move} is mutually exclusive (never combine it with therapeutic move labels).
\end{ontolist}
 
\subsection{Decision Order (Function-First)}
 
Apply top to bottom and add each move that is clearly present as a distinct function:
 
\begin{ontoenum}
\item[0.] If turn is administrative/social/procedural with no defined meaningful therapeutic function $\rightarrow$ \texttt{no\mvu defined\mvu move}
\item[1.] If therapist calibrates collaboration, pacing, fit, or rupture $\rightarrow$ \texttt{do\mvu process\mvu alignment}
\item[2.] If therapist sets agenda, priorities, or explicit goals $\rightarrow$ \texttt{do\mvu goal\mvu setting}
\item[3.] If the therapist proposes or develops a concrete action, practice, or experiment to try next $\rightarrow$ \texttt{do\mvu action\mvu planning}
\item[4.] If therapist teaches a general mechanism/principle/rationale $\rightarrow$ \texttt{do\mvu psychoeducation}
\item[5.] If the therapist introduces, orients the person to, or guides the person through learning, practicing, or rehearsing a skill in the interaction $\rightarrow$ \texttt{do\mvu skill\mvu building}
\item[6.] If therapist highlights discrepancy/incongruence/faulty reasoning, including through directional inquiry that challenges the member's perspective $\rightarrow$ \texttt{do\mvu challenge}
\item[7.] If therapist adds inferred meaning/function/causal model, including smaller-scale extensions $\rightarrow$ \texttt{do\mvu interpret}
\item[8.] If the therapist strengthens hope, agency, readiness, exploration of change, or client-owned motivation for change $\rightarrow$ \texttt{do\mvu support\mvu change}
\item[9.] If the therapist reflects, validates, normalizes, paraphrases, synthesizes, clarifies what the experience is or is not, or otherwise communicates understanding without adding a new explanatory claim $\rightarrow$ \texttt{do\mvu shared\mvu understanding}
\item[10.] Otherwise, if the turn is primarily seeking information, clarification, or greater depth $\rightarrow$ \texttt{do\mvu inquiry}
\end{ontoenum}
 
Tie-break rules:
 
\begin{ontolist}
\item Function over tone.
\item Function over syntax.
\item When a turn is phrased as a question, do not default to \texttt{do\mvu inquiry}; code the move that best matches the therapeutic function being performed.
\item If interpretation is phrased as a question, still code \texttt{do\mvu interpret} on its own.
\item If a question's main aim is to explore or evoke motivation, readiness, ambivalence, or what change would mean (rather than primarily gather information), code \texttt{do\mvu support\mvu change}.
\item If support is brief but main act is plan/goal/challenge, code the main act.
\item Prefer fewer labels when uncertain; only add a second/third label when it is sufficiently supported by the therapeutic function of the turn.
\item If any therapeutic move label applies, do not include \texttt{no\mvu defined\mvu move}.
\item If a question directionally examines discrepancy, assumption, rigidity, or a limiting perspective, code \texttt{do\mvu challenge}.
\item If a question offers a tentative understanding and also seeks confirmation, clarification, or elaboration, code \texttt{do\mvu shared\mvu understanding}, and also code \texttt{do\mvu inquiry} when the information-seeking function is meaningfully present.
\end{ontolist}
 
\subsection{Move Cards}
\label{app:move_cards}
 
\subsubsection{\texttt{do\mvu process\mvu alignment}}
 
\noindent\textbf{Description.} Calibrate how therapy is working in real time: pacing, depth, tone, direction, and collaboration quality. Includes explicit repair when the therapist has likely missed the client, moved too fast, or created strain in the interaction. This is process-focused meta-communication, not content exploration.
 
\noindent\textbf{When to use.}
\begin{ontolist}
\item after a potential rupture, defensiveness spike, or visible disengagement;
\item when checking whether current direction or pace feels right to the client;
\item when deciding together whether to pursue exploration of internal experience vs clinical intervention (e.g., problem solving, guided clinical activity, psychoeducation).
\end{ontolist}
 
\noindent\textbf{Contraindications.}
\begin{ontolist}
\item when asking about life events, symptoms, or meaning (use \texttt{do\mvu inquiry});
\item when seeking confirmation that an interpretation is factually correct (usually \texttt{do\mvu shared\mvu understanding}/\texttt{do\mvu interpret} boundary);
\item when used repeatedly to avoid difficult but productive material.
\end{ontolist}
 
\noindent\textbf{Execution guidance.}
\begin{ontolist}
\item briefly name the process observation;
\item if repair is needed, acknowledge and take responsibility for role in the rupture;
\item if needed, ask a clarification question to define the source of the rupture in order to support the adjustment needed;
\item if within ability, offer a relevant adjustment.
\end{ontolist}
 
\noindent\textbf{MULTI-60 grounding.}
\textit{Primary:}
\begin{ontolist}
\item Item 28: working together with the client as a team.
\end{ontolist}
\noindent \textit{Supporting:}
\begin{ontolist}
\item Item 38: exploring the client's feelings about therapy.
\end{ontolist}
 
\noindent\textbf{Examples.}
\begin{ontolist}
\item ``I think I moved too quickly there. How did that land for you?''
\item ``I may have missed you just now. Can we rewind and get your version first?''
\end{ontolist}
 
\subsubsection{\texttt{do\mvu goal\mvu setting}}
 
\noindent\textbf{Description.} Collaboratively identify, clarify, or prioritize what the member wants to work toward. This includes selecting a focus, choosing among competing issues, and helping define broad aims in a more concrete and workable way.
 
\noindent\textbf{When to use.}
\begin{ontolist}
\item at session/interaction start, transition points, or time-limited moments;
\item when multiple issues compete and prioritization is needed;
\item when converting broad areas for change into concrete targets.
\end{ontolist}
 
\noindent\textbf{Contraindications.}
\begin{ontolist}
\item when therapist is already specifying action steps (use \texttt{do\mvu action\mvu planning});
\item when the turn only gathers details (use \texttt{do\mvu inquiry}).
\end{ontolist}
 
\noindent\textbf{Execution guidance.}
\begin{ontolist}
\item surface candidate targets;
\item negotiate what to prioritize right now;
\item constrain scope to realistic session bandwidth;
\item If applicable, collaboratively break down the goal into observable steps or terms.
\end{ontolist}
 
\noindent\textbf{MULTI-60 grounding.}
\textit{Primary:}
\begin{ontolist}
\item Item 1: setting an agenda or specific goals for the session.
\item Item 9: making a plan with the client to increase or decrease specific behaviors.
\end{ontolist}
\noindent \textit{Supporting:}
\begin{ontolist}
\item Item 28: working together with the client as a team.
\item Item 42: improving the client's ability to solve their own problems.
\end{ontolist}
 
\noindent\textbf{Examples.}
\begin{ontolist}
\item ``Since one of your goals is to get over your fear of going to the gym, let's make a plan for you to do so.''
\item ``Let's choose one goal for sleep this week and one goal for social avoidance.''
\end{ontolist}
 
\subsubsection{\texttt{do\mvu inquiry}}
 
\noindent\textbf{Description.} Ask focused questions to deepen understanding of meaning, sequence, context, or concrete details for the member and/or the therapist. Includes both exploratory inquiry (values, emotions, meaning) and clarifying inquiry (when, frequency, triggers, sequence, body cues). Does not need to be in the form of a question, but could come in the form of an instruction to follow that serves this function. The function is understanding, not persuasion, confrontation, or another move being performed in question form.
 
\noindent\textbf{When to use.}
\begin{ontolist}
\item to map what happened before/during/after key moments;
\item to disambiguate thought vs feeling vs behavior vs sensation;
\item to deepen personal significance of events, relationship moments, or wishes.
\item To build case conceptualization and hypotheses related to the development and maintenance of symptoms and functional challenges.
\end{ontolist}
 
\noindent\textbf{Contraindications.}
\begin{ontolist}
\item when question primarily tests discrepancy or challenges the member's perspective (use \texttt{do\mvu challenge});
\item when question primarily repairs process fit (use \texttt{do\mvu process\mvu alignment});
\item when question mainly introduces a therapist-generated hypothesis, meaning, or explanatory pattern that is not yet explicit in the member's account (use \texttt{do\mvu interpret});
\item when question primarily explores or evokes motivation, readiness, ambivalence, or what change would mean (use \texttt{do\mvu support\mvu change}). -
\end{ontolist}
 
\noindent\textbf{Execution guidance.}
\begin{ontolist}
\item ask one high-yield question at a time;
\item anchor to a specific moment before broad generalization;
\item prefer concrete wording over abstract prompts;
\item avoid rapid question stacking.
\end{ontolist}
 
\noindent\textbf{MULTI-60 grounding.}
\textit{Primary:}
\begin{ontolist}
\item Item 40: encouraging the client to explore the personal meaning of an event or feeling.
\end{ontolist}
\noindent \textit{Supporting:}
\begin{ontolist}
\item Item 46: showing interest in understanding what the client is experiencing.
\item Item 47: encouraging the client to focus on their moment-to-moment experience.
\end{ontolist}
 
\noindent\textbf{Examples.}
\begin{ontolist}
\item ``When panic started on the train, what did you notice first in your body?''
\item ``What felt most painful about that argument with your partner?''
\item ``In that dream, what part felt most emotionally intense to you?''
\item ``When did that challenge first show up in your life?``
\item ``I wonder if you could give me an example of a recent time when you really felt a lot of anger''
\end{ontolist}
 
\subsubsection{\texttt{do\mvu shared\mvu understanding}}
 
\noindent\textbf{Description.} Communicate an understanding of what the person shared in a way that helps them feel heard, understood, and accurately followed, including at times clarifying what the experience is or is not. This can include reflection, paraphrase, concise synthesis across multiple points, validation, normalization, and signs of active listening. Other signs of active listening may include brief acknowledgments that show attention and tracking (e.g., ``mm-hm,'' ``right,'' ``I see''), acknowledging what feels most important in what was shared, recognizing shifts in emotion or emphasis, and responding in a way that shows continuity with what has already been said.
 
\noindent\textbf{When to use.}
\begin{ontolist}
\item after meaningful disclosure to demonstrate listening, build shared understanding, and support clarification or expansion of what was shared;
\item to slow pace and help important material land;
\item when the emotional experience needs to be acknowledged or validated before moving forward;
\item when normalization may help reduce isolation or shame;
\item before moving to interpretation, challenge, planning, or change-oriented guidance..
\end{ontolist}
 
\noindent\textbf{Contraindications.}
\begin{ontolist}
\item when introducing hidden meaning, causal explanation, or inferred mechanisms not already grounded in what was shared (use \texttt{do\mvu interpret});
\item when explicitly trying to reinforce effort, build motivation, or evoke change talk (use \texttt{do\mvu support\mvu change});
\item when the response is intended to directly highlight discrepancy, tension, or inconsistency in what the person is expressing or doing (use \texttt{do\mvu challenge}).
\item when normalization would minimize, flatten, or prematurely reassure rather than help the person feel understood.
\end{ontolist}
 
\noindent\textbf{Execution guidance.}
\begin{ontolist}
\item Respond in a way that shows close understanding of what the person shared, using plain language grounded in their expressed experience.
\item This can take the form of a reflection, paraphrase, concise synthesis across points, validation of the emotional experience, normalization when appropriate, or other signs of active listening such as brief acknowledgments that show attention and tracking.
\item Match the form of the response to what is most needed in the moment: brief acknowledgments when continued sharing is needed; fuller reflections or synthesis when it would help consolidate what was shared; validation when the emotional experience needs to be explicitly recognized; normalization when it would help reduce isolation or shame without minimizing the experience.
\item Stay within the bounds of what the person has expressed. Do not add hidden meaning, causal explanations, or hypotheses beyond what was shared.
\item This can include clarifying what the person's experience is not, when that clarification helps define the experience more accurately without adding a new explanatory claim.
\item When it is not certain that the understanding is accurate, use tentative language such as ``It sounds like\ldots{},'' ``It seems like\ldots{},'' or ``Is it accurate to say\ldots{}''
\item Shared understanding may be paired with \texttt{do\mvu inquiry} when the therapist communicates a tentative understanding and also seeks confirmation, clarification, or elaboration from the member.
\item Keep the response focused, proportional, and centered on helping the person feel heard, understood, and supported in continuing the conversation.
\end{ontolist}
 
\noindent\textbf{MULTI-60 grounding.}
\textit{Primary:}
\begin{ontolist}
\item Item 10: paraphrasing back the meaning of what the client said.
\item Item 31: listening carefully to what the client is saying.
\end{ontolist}
\noindent \textit{Supporting:}
\begin{ontolist}
\item Item 1: setting an agenda or specific goals for the session.
\item Item 18: being warm, sympathetic, and accepting.
\end{ontolist}
 
\noindent\textbf{Examples.}
\begin{ontolist}
\item ``You felt trapped and overwhelmed, and then shut down.''
\item ``So this week was less sleep, more conflict, and then anxiety spiked.''
\end{ontolist}
 
\subsubsection{\texttt{do\mvu support\mvu change}}
 
\noindent\textbf{Description.} Strengthen the person's readiness, willingness, and confidence to move toward change. This includes reinforcing effort or movement already shown, highlighting agency and choice, exploring the person's own reasons and values for change, and instilling realistic hope that change or improvement is possible. The primary aim is to support movement toward change by strengthening or exploring motivation, readiness, ambivalence, confidence, and the personal meaning of change, without persuading, pressuring, or prematurely moving into action design.
 
\noindent\textbf{When to use.}
\begin{ontolist}
\item when client ambivalence about change is present or commitment is not yet clear;
\item when exploring whether change feels possible, desirable, or worthwhile;
\item when reflecting on readiness, ambivalence, or what change would mean in the person's life;
\item when exploring the person's own reasons, values, hopes, or perceived benefits related to change;
\item when effort, progress, or values-consistent movement can be reinforced to support continued change;
\item before moving into action planning, when motivation, readiness, or confidence still needs to be strengthened.
\end{ontolist}
 
\noindent\textbf{Contraindications.}
\begin{ontolist}
\item when the main need is understanding, reflection, validation, or normalization rather than readiness enhancement (use \texttt{do\mvu shared\mvu understanding});
\item when the main function is concrete step design or implementation planning (use \texttt{do\mvu action\mvu planning});
\item when the main function is gathering information or deepening understanding rather than exploring or strengthening movement toward change, motivation for change, or readiness for change (use \texttt{do\mvu inquiry});
\item when the main function is directly surfacing discrepancy, tension, or a limiting perspective to build awareness (use \texttt{do\mvu challenge});
\end{ontolist}
 
\noindent\textbf{Execution guidance.}
\begin{ontolist}
\item choose mode intentionally: stabilize first if affect is high; evoke when readiness, ambivalence, or the meaning of change needs to be explored;
\item validate emotion in context, not belief accuracy;
\item in evoke mode, ask open prompts about importance, confidence, values, readiness, ambivalence, or what change would mean, then reflect change talk;
\item reinforce movement, effort, and agency in a way that is specific and proportional to what the person has expressed or done;
\item preserve autonomy language (choice, willingness, fit);
\item do not prescribe actions while in evoke mode.
\end{ontolist}
 
\noindent\textbf{MULTI-60 grounding.}
\textit{Primary:}
\begin{ontolist}
\item Item 7: giving the client hope or encouragement.
\item Item 25: encouraging the client to weigh the positive and negative consequences of acting in a new way.
\item Item 56: accepting the client as they are while encouraging change.
\item Item 23: encouraging the client to see the options they have in their life.
\end{ontolist}
\noindent \textit{Supporting:}
\begin{ontolist}
\item Item 52: reviewing the progress the client has made in therapy.
\item Item 42: improving the client's ability to solve their own problems.
\end{ontolist}
 
\noindent\textbf{Examples.}
\begin{ontolist}
\item ``Even with how hard this has felt, part of you still wants something different.''
\item ``You've already taken some meaningful steps, even if it still feels hard.''
\item ``What feels most important to you about making this change now?''
\item ``What gives you even a small sense that this could get better?''
\item ``It may not change all at once, but there are real signs here that movement is possible.''
\end{ontolist}
 
\subsubsection{\texttt{do\mvu interpret}}
 
\noindent\textbf{Description.} Offer a therapist-generated hypothesis, opinion, or framework about the underlying meaning, function, or pattern in what the person is experiencing. This can include a local hypothesis about what may be happening in a specific moment, a smaller-scale extension of what the person has shared, what is observed by the therapist, or a broader conceptualization that links multiple experiences into a coherent pattern. Includes interpretive links such as past--present themes, possible functions of symptoms or behaviors, internal conflict (e.g., competing motivations, goals, or priorities), patterns in thoughts, beliefs, motives, reactions, coping, or avoidance, and formulations that connect difficulties or experiences across domains. The goal is to help the person see a pattern, meaning, or organizing framework that is not yet fully explicit in what they have shared.
 
\noindent\textbf{When to use.}
\begin{ontolist}
\item when sufficient context exists to support a plausible, evidence-based hypothesis;
\item when multiple pieces of information can be integrated into a coherent pattern or model;
\item when the person seems stuck, confused, or repetitive in a way that may benefit from a new organizing perspective;
\item when the clinician's synthesis could help deepen understanding of what may be driving, maintaining, or connecting the person's difficulties;
\item when a therapist-generated formulation may help deepen understanding of what is driving, maintaining, or connecting the person's difficulties.
\end{ontolist}
 
\noindent\textbf{Contraindications.}
\begin{ontolist}
\item when the response is primarily reflecting, paraphrasing, labeling the experience, validating, normalizing, or otherwise communicating understanding of content already explicit in what was shared  (use \texttt{do\mvu shared\mvu understanding});
\item when the main goal is to directly surface discrepancy, tension, inconsistency, or a limiting perspective rather than propose a broader model (use \texttt{do\mvu challenge});
\item when the response is mainly providing educational information rather than a person-specific hypothesis or formulation  (use \texttt{do\mvu psychoeducation});
\item when there is too little context to support a plausible interpretation;
\item when the interpretation would move too far beyond available data, imply diagnosis, or present speculation as fact.
\end{ontolist}
 
\noindent\textbf{Execution guidance.}
\begin{ontolist}
\item Present interpretations tentatively and collaboratively, as possible ways of understanding what may be happening rather than as facts or conclusions.
\item Ground the interpretation in material the person has actually shared, linking it to observable patterns, repeated themes, or meaningful features of the conversation.
\item This can include smaller-scale extensions or therapist-generated opinions that go beyond reflection, as long as they add meaning, function, or perspective that is not already explicit in what was shared.
\item Interpretations may address patterns in emotion, thought, belief, motivation, coping, avoidance, behavior, internal conflict, or other clinically relevant aspects of experience.
\item Make clear what the interpretation is connecting or explaining, rather than offering a vague impression.
\item Invite correction, elaboration, or non-fit explicitly.
\item Avoid over-certainty, hidden leaps, diagnosis language, or interpretations that are more complex than the available material supports.
\item Keep the interpretation clear and concise enough that the person can engage with it, respond to it, and decide whether it resonates.
\end{ontolist}
 
\noindent\textbf{MULTI-60 grounding.}
\textit{Primary:}
\begin{ontolist}
\item Item 2: connecting the client's current situation with their past.
\item Item 20: discussing the function or purpose the client's problem may serve.
\item Item 27: helping the client identify the positive or negative consequences of their behavior.
\end{ontolist}
\noindent \textit{Supporting:}
\begin{ontolist}
\item Item 19: pointing out recurring themes or problems in the client's relationships.
\end{ontolist}
 
\noindent\textbf{Examples.}
\begin{ontolist}
\item ``I wonder if withdrawing is a way to protect yourself from expected rejection.''
\item ``Part of you wants closeness, while another part expects hurt, and that conflict keeps you stuck.''
\item ``I'm noticing that when things feel uncertain, you tend to step back, and that might be part of what's making it harder to get traction.''
\item ``It could be that the pressure to get this right is actually making it harder to take any action at all.''
\item ``I wonder if some of this pattern developed as a way to manage earlier experiences, even though it may not be working the same way now.''
\end{ontolist}
 
\subsubsection{\texttt{do\mvu challenge}}
 
\noindent\textbf{Description.} Surface and examine beliefs, interpretations, patterns, or coping responses that may be keeping the person stuck or moving them away from their goals. Challenge is not criticism, confrontation, or argument; it is a purposeful intervention used to increase awareness, flexibility, and openness to change. This can include examining discrepancies, questioning assumptions, testing the accuracy or usefulness of a thought, and naming rigid, avoidant, or self-defeating patterns when clinically appropriate. It can also include directional inquiry that challenges the member's perspective. The goal is to help the person look more directly at something that may be maintaining a problem, limiting perspective, or interfering with progress.
 
\noindent\textbf{When to use.}
\begin{ontolist}
\item when a belief, interpretation, behavior, or coping response appears to be maintaining the problem or interfering with progress;
\item when there is a meaningful discrepancy between the person's goals, values, interpretations, emotions, or actions that would be useful to examine directly;
\item when examining the accuracy, usefulness, or consequences of a thought or pattern could increase flexibility and improve response options.
\end{ontolist}
 
\noindent\textbf{Contraindications.}
\begin{ontolist}
\item when rapport, trust, or shared understanding is too limited for challenge to be received productively;
\item when immediate priorities are safety, stabilization, containment, or basic emotional support;
\item when the main task is clarification or exploration without a discrepancy, limiting perspective, or problematic pattern being directly named (use \texttt{do\mvu inquiry});
\item when the main task is communicating understanding, validation, or normalization of what the person has shared (use \texttt{do\mvu shared\mvu understanding});
\item when the main task is offering a broader explanatory hypothesis about underlying meaning, function, or pattern  (use \texttt{do\mvu interpret}).
\item when the intervention would come across as criticism, shaming, arguing, or pushing the person to accept the clinician's view.
\end{ontolist}
 
\noindent\textbf{Execution guidance.}
\begin{ontolist}
\item target one belief, pattern, discrepancy, or response at a time;
\item Ground the intervention in something already present in the person's words, behavior, or reported experience;
\item Be clear and specific about what is being examined;
\item Link the challenge to the person's goals, functioning, values, or stated concerns;
\item Invite examination, reflection, or testing rather than forcing agreement;
\item Use language that is direct but collaborative, such as asking whether something fits, what the person notices, or how they understand the discrepancy;
\item Directional questions can still be \texttt{do\mvu challenge} when their function is to examine or unsettle a limiting perspective rather than simply gather information;
\item Avoid stacking multiple challenges in sequence or escalating intensity when the person is not engaging with the first one;
\end{ontolist}
 
\noindent\textbf{MULTI-60 grounding.}
\textit{Primary:}
\begin{ontolist}
\item Item 13: pointing out when the client's behavior seems inconsistent with what they are saying.
\item Item 21: encouraging explanations for events or behaviors other than the first that came to mind.
\item Item 37: encouraging the client to look for evidence for or against a belief or assumption.
\item Item 39: encouraging the client to see their problem from a different perspective.
\item Item 49: encouraging the client to question their beliefs or find flaws in their reasoning.
\end{ontolist}
 
\noindent\textbf{Examples.}
\begin{ontolist}
\item ``I notice part of you says this relationship matters, and part of you keeps stepping back when it gets vulnerable.''
\item ``You've said this goal really matters to you, and I'm noticing a pattern that may be pulling in the opposite direction.''
\item ``What evidence supports that idea, and what evidence pushes against it?''
\end{ontolist}
 
\subsubsection{\texttt{do\mvu psychoeducation}}
 
\noindent\textbf{Description.} Provide information, rationale, or explanatory framing that helps the person better understand their experience, symptoms, behavior patterns, the treatment process, or a recommendation. Psychoeducation is a targeted intervention used to support insight and understanding by explaining the rationale for an intervention, the relevant science or theory behind an intervention or experience, or why or how a pattern, symptom, or intervention may be affecting the person and what may help. Includes explaining clinical processes, treatment rationales, common maintaining mechanisms, skill purposes, and relevant links between thoughts, behavior, physiology, and context in a way that is accurate, usable, and tailored to the moment. The goal is to improve understanding in a way that supports engagement, decision-making, or next steps.
 
\noindent\textbf{When to use.}
\begin{ontolist}
\item when the person would benefit from a clearer framework for understanding what they are experiencing;
\item when explaining a clinical concept or maintaining process could support insight, motivation, or next-step engagement;
\item when providing rationale, explanatory context, science, theory, or mechanism related to a skill, intervention, treatment direction, experience, or pattern;
\item when the person is making sense of symptoms, reactions, or behavior patterns in a way that could be usefully clarified or reframed;
\item when brief explanatory guidance would help orient the person without interrupting the therapeutic process.
\end{ontolist}
 
\noindent\textbf{Contraindications.}
\begin{ontolist}
\item when the response is focused on explaining this specific person's underlying pattern, meaning, or function (use \texttt{do\mvu interpret});
\item when the main function is proposing or structuring specific action steps (use \texttt{do\mvu action\mvu planning});
\item when the main function is introducing, orienting to, or guiding the person through an exercise, activity, or skill within the interaction (use \texttt{do\mvu skill\mvu building});
\item when the main task is communicating understanding, validation, or normalization of the person's experience (use \texttt{do\mvu shared\mvu understanding});
\item when the information is not clearly relevant, actionable, or timed to support the current therapeutic task;
\item when the person is seeking emotional understanding, and information would bypass or dilute the more immediate need;
\item when the person is too overwhelmed, dysregulated, or cognitively overloaded to meaningfully take in new information;
\end{ontolist}
 
\noindent\textbf{Execution guidance.}
\begin{ontolist}
\item introduce one concept or rationale at a time;
\item use plain, concrete language rather than technical or academic phrasing;
\item tie the explanation directly to the person's current experience, goal, or next step;
\item capture psychoeducation when a rationale, mechanism, or explanatory frame is embedded briefly or conversationally in the turn, even if it is not presented as formal teaching;
\item Do not use \texttt{do\mvu psychoeducation} for orienting the person to what to do in an exercise or skill. Use \texttt{do\mvu skill\mvu building} when the turn is introducing or guiding the activity itself. Add \texttt{do\mvu psychoeducation} only when rationale, science, theory, or explanation of how the intervention, experience, or pattern may help or make sense is also being provided.
\item prioritize information that is actionable or meaningfully clarifying;
\item break longer explanations into digestible parts that fit a text-based exchange;
\item Keep explanations concise and suited to a text-based format; avoid dense or overly long messages.
\item Psychoeducation may be co-coded with other moves when explanation is woven into reflection, support for change, or planning.
\item Avoid lecturing, overexplaining, or drifting away from the person's context.
\item Check for understanding or relevance before moving forward
\end{ontolist}
 
\noindent\textbf{MULTI-60 grounding.}
\textit{Supporting:}
\begin{ontolist}
\item Item 32: explaining what the therapist is trying to do.
\item Item 59: making clear that the client's problem is a treatable medical condition.
\item Item 58: encouraging the client to be aware of things in their life without judging them (when psychoeducation on mindfulness or meditation is present).
\end{ontolist}
 
\noindent\textbf{Examples.}
\begin{ontolist}
\item ``One reason avoidance can feel so convincing is that it lowers discomfort quickly, even though it often keeps the fear going over time.''
\item ``When sleep becomes irregular, it can affect mood, energy, and concentration in ways that build on each other.''
\item ``Part of why this skill is useful is that it helps create a little space between the thought and the reaction.''
\item ``Sometimes when stress stays high for a while, the body starts reacting as if there is danger even when there isn't an immediate threat.''
\end{ontolist}
 
\subsubsection{\texttt{do\mvu action\mvu planning}}
 
\noindent\textbf{Description.} Translate insight, intention, or treatment direction into a specific next step the person can try. This includes collaboratively developing a concrete behavior, exercise, coping practice, experiment, rehearsal, observation task, interpersonal response plan, or other instruction for the person to follow. The action should be clear enough to carry out and specific enough to later review, learn from, or adjust.
 
\noindent\textbf{When to use.}
\begin{ontolist}
\item after enough shared understanding and at least partial readiness to try something;
\item when the person asks for practical next steps or wants help deciding what to do;
\item when setting up a future-oriented next step to carry out outside the interaction;
\item when asking or telling the person to do something clear, concrete, and defined within the interaction, and the turn is not clearly orienting them to a skill or helping them perform or practice it within the interaction;
\item when the primary function is identifying an action for the person to take, rather than teaching or practicing a skill within the interaction;
\item when a broader goal has already been identified and the next task is to define how to begin;
\item when follow-through is more likely to improve with greater specificity or structure.
\end{ontolist}
 
\noindent\textbf{Contraindications.}
\begin{ontolist}
\item when the person is ambivalent about change or not yet ready to commit to it, and readiness or willingness needs to be strengthened first (use \texttt{do\mvu support\mvu change});
\item when the main task is still clarifying the goal or deciding what direction matters most (use \texttt{do\mvu goal\mvu setting});
\item when the main task is explaining rationale or teaching a concept (use \texttt{do\mvu psychoeducation});
\item when planning would function as premature problem-solving before enough understanding or buy-in is present.
\item when the main function is teaching or guiding the person through how to perform a skill within the interaction (use \texttt{do\mvu skill\mvu building});
\end{ontolist}
 
\noindent\textbf{Execution guidance.}
\begin{ontolist}
\item start from the person's stated goal, desired change, or area of difficulty, and collaboratively translate that into a specific next step;
\item Use \texttt{do\mvu action\mvu planning} when the turn primarily identifies a clear, concrete, and defined next step the person is being asked to take. When a turn tells or asks the person to do something specific and is not clearly orienting the person to a skill or helping them perform or practice it within the interaction, use \texttt{do\mvu action\mvu planning}.
\item help bridge insight or intention into action by shaping the plan together, rather than either assigning a plan or expecting the person to generate one alone;
\item define the action clearly, including what will be done, when, where, and under what conditions;
\item keep the scope small, realistic, and appropriate to current readiness, context, and likely barriers;
\item check for fit and feasibility throughout the planning process and adjust the step as needed;
\item obtain explicit buy-in before finalizing the plan;
\item frame the step as an opportunity to learn, practice, or gather information rather than as pass/fail compliance;
\item when relevant, identify how the person will notice, track, or reflect on what happens;
\item set review point for next session when appropriate.
\end{ontolist}
 
\noindent\textbf{MULTI-60 grounding.}
\textit{Primary:}
\begin{ontolist}
\item Item 9: making a plan with the client to increase or decrease specific behaviors.
\item Item 17: assigning or reviewing homework.
\item Item 35: encouraging the client to change specific behaviors.
\end{ontolist}
\noindent \textit{Supporting:}
\begin{ontolist}
\item Item 16: encouraging the client to think about, view, or touch things they are afraid of.
\item Item 51: exploring ways the client could make changes in their relationships.
\end{ontolist}
 
\noindent\textbf{Examples.}
\begin{ontolist}
\item ``You've said you want to feel less overwhelmed when panic starts. What do you think about making the first step simply noticing one trigger this week and writing down what you did right after?''
\item ``Since your goal is to feel a little more steady in the evenings, maybe we could turn that into one small step. Would a five-minute walk after dinner feel realistic, or is there a better place to start?''
\item ``You want to handle conflict more directly instead of shutting down. Maybe the first step could be practicing one sentence that states your need before stepping away. How does that sound?''
\end{ontolist}
 
\subsubsection{\texttt{do\mvu skill\mvu building}}
 
\noindent\textbf{Description.} Guide the person through learning, practicing, or rehearsing a specific skill in the session or conversation itself. It needs to be clear what skill is being introduced or practiced (role play, diaphragmatic breathing, exposure); if you cannot identify the skill, do not use \texttt{do\mvu skill\mvu building}. This includes introducing the skill, orienting the person to what they will do, teaching the steps of the skill, supporting the person in trying it in real time, and helping them reflect on how it worked. The focus is on building capability through guided practice, not just explaining the skill or planning to use it later.
 
\noindent\textbf{When to use.}
\begin{ontolist}
\item when the person is ready to actively try a skill and would benefit from guided support in learning it;
\item when practicing the skill in the moment could increase understanding, confidence, or likelihood of later use;
\item when the skill is best learned experientially rather than through explanation alone;
\item when the person is asking for help with how to do a skill, not just whether they should do it;
\item when in-session rehearsal, walkthrough, or supported practice could help turn insight into capability.
\end{ontolist}
 
\noindent\textbf{Contraindications.}
\begin{ontolist}
\item when the person is ambivalent or not yet ready to engage in the skill and motivation needs to be strengthened first (use \texttt{do\mvu support\mvu change});
\item when the main task is explaining the rationale or concept behind the skill rather than practicing it (use \texttt{do\mvu psychoeducation});
\item when the main task is planning for use of the skill outside the interaction (use \texttt{do\mvu action\mvu planning});
\end{ontolist}
 
\noindent\textbf{Execution guidance.}
\begin{ontolist}
\item Introduce the skill in a conversational way, using clear, plain language that fits the person's current situation, goals, or needs;
\item Explain one part of the skill at a time rather than giving the full explanation all at once;
\item Guide the person through each step in sequence;
\item When a stretch of turns is collectively introducing and practicing the skill, apply \texttt{do\mvu skill\mvu building} throughout the turns where the skill-building function is actively being carried forward, and add other move labels when additional therapeutic functions are also clearly performed within those turns.
\item Alternate between explanation and practice so the person is learning by doing, not just listening;
\item Do not limit \texttt{do\mvu skill\mvu building} only to the moment of active practice; include turns that are part of introducing, orienting to, and carrying forward the skill within the interaction when that function is still underway.
\item Check understanding, fit, and response throughout the learning process, and clarify or adjust when needed;
\item Use examples, prompts, modeling, or rehearsal when that helps make the skill easier to understand and try;
\item Stay responsive to the person's pace, questions, and reactions rather than delivering the skill in a rigid or scripted way;
\item Keep the interaction focused on helping the person try the skill, notice what happens, and make sense of the experience;
\item Avoid dense, technical, or overly instructional language that makes the exchange feel like a lesson rather than a therapeutic conversation;
\item End by summarizing the key learning in simple language and, when relevant, linking it to when the skill could be used again.
\end{ontolist}
 
\noindent\textbf{MULTI-60 grounding.}
\textit{Primary:}
\begin{ontolist}
\item Item 15: teaching the client specific new skills or behaviors.
\end{ontolist}
\noindent \textit{Secondary:}
\begin{ontolist}
\item Item 16: encouraging the client to think about, view, or touch things they are afraid of.
\item Item 47: encouraging the client to focus on their moment-to-moment experience.
\end{ontolist}
 
\subsubsection{\texttt{no\mvu defined\mvu move}}
 
\noindent\textbf{Description.} Use when a therapist turn does not perform a target therapeutic move from this ontology. Prevents forced over-coding and improves dataset quality. This is expected and valid in real transcripts.
 
\noindent\textbf{When to use.}
\begin{ontolist}
\item for scheduling, logistics, or technical setup;
\item for greetings or closings that do not serve a defined clinical function;
\item for procedural statements such as timing, consent reminders, or transitions;
\item for brief backchannels or acknowledgments that do not reflect active listening, shared understanding, or another defined move;
\item when a turn supports the flow of the interaction but does not itself advance a therapeutic task captured in this ontology.
\end{ontolist}
 
\noindent\textbf{Contraindications.}
\begin{ontolist}
\item when the turn performs a meaningful therapeutic function, even briefly;
\item when the turn includes active listening, reflection, validation, normalization, or another clinically meaningful response, even if short;
\item when the turn combines procedural content with a substantive therapeutic move, in which case code the substantive move;
\item when the turn appears minimal on the surface but is clearly serving a defined clinical purpose in context.
\end{ontolist}
 
\noindent\textbf{Execution guidance.}
\begin{ontolist}
\item prefer \texttt{no\mvu defined\mvu move} over guessing when no therapeutic function is present rather than forcing the turn into a nearby category;
\item focus on the function of the turn, not just its length or simplicity;
\item if a turn includes both non-therapeutic and therapeutic content, code the therapeutic move rather than \texttt{no\mvu defined\mvu move};
\item when uncertain, ask whether the turn is contributing to understanding, support, insight, motivation, planning, skill use, challenge, or another defined therapeutic function;
\item if yes, choose the matching move; if no, use \texttt{no\mvu defined\mvu move}.
\end{ontolist}
 
\noindent\textbf{MULTI-60 grounding.} Not a MULTI item; this is an operational control label for annotation quality.
 
\noindent\textbf{Examples.}
\begin{ontolist}
\item ``Sounds good''
\item ``Got it''
\item ``Alright, we can pick this up next time''
\item ``Hey---good to hear from you''
\item ``One sec, just pulling that up''
\item ``Can you still see my message?''
\end{ontolist}
 
\subsection{High-Confusion Boundaries}
 
\noindent\textbf{\texttt{do\mvu inquiry} vs \texttt{do\mvu challenge}.}
\begin{ontolist}
\item Inquiry seeks understanding, clarification, or elaboration.
\item Challenge directly examines a discrepancy, assumption, rigid belief, limiting perspective, or problematic pattern, including when this is done through directional inquiry.
\end{ontolist}
 
\noindent\textbf{\texttt{do\mvu interpret} vs \texttt{do\mvu challenge}.}
\begin{ontolist}
\item Interpret proposes a therapist-generated hypothesis about meaning, function, or pattern.
\item Challenge highlights, questions, or tests something already visible in the person's thinking, behavior, or self-report.
\end{ontolist}
 
\noindent\textbf{\texttt{do\mvu shared\mvu understanding} vs \texttt{do\mvu support\mvu change}.}
\begin{ontolist}
\item Shared understanding communicates accurate listening, reflection, validation, normalization, or other signs of close tracking.
\item Support change strengthens hope, readiness, agency, or motivation for change.
\end{ontolist}
 
\noindent\textbf{\texttt{do\mvu inquiry} vs \texttt{do\mvu support\mvu change}.}
\begin{ontolist}
\item Inquiry gathers information or deepens understanding.
\item Support change explores or strengthens readiness, ambivalence, confidence, reasons for change, or what change would mean.
\end{ontolist}
 
\noindent\textbf{\texttt{do\mvu goal\mvu setting} vs \texttt{do\mvu action\mvu planning}.}
\begin{ontolist}
\item Goal setting identifies, clarifies, or prioritizes what the person wants to work toward.
\item Action planning translates that goal into a concrete next step.
\end{ontolist}
 
\noindent\textbf{\texttt{do\mvu support\mvu change} vs \texttt{do\mvu action\mvu planning}.}
\begin{ontolist}
\item Support change explores or strengthens readiness, willingness, confidence, ambivalence, or the meaning of change.
\item Action planning specifies what the person will try next and helps set it up.
\end{ontolist}
 
\noindent\textbf{\texttt{do\mvu psychoeducation} vs \texttt{do\mvu interpret}.}
\begin{ontolist}
\item Psychoeducation provides rationale, explanatory context, science, theory, or mechanism, including when this is delivered subtly or conversationally.
\item Interpret applies a therapist-generated hypothesis to this person's specific pattern, function, or meaning.
\end{ontolist}
 
\noindent\textbf{\texttt{do\mvu process\mvu alignment} vs \texttt{do\mvu inquiry}.}
\begin{ontolist}
\item Process alignment asks about the fit or direction of the therapeutic interaction itself.
\item Inquiry asks about the person's life, experience, symptoms, thoughts, feelings, or behavior.
\end{ontolist}
 
\noindent\textbf{\texttt{do\mvu shared\mvu understanding} vs \texttt{do\mvu interpret}.}
\begin{ontolist}
\item Shared understanding stays within what is already explicit in what the person shared, including clarifying what the experience is or is not.
\item Interpret adds a plausible, therapist-generated linkage, function, organizing pattern, opinion, or extension beyond what is already explicit.
\end{ontolist}
 
\noindent\textbf{\texttt{do\mvu psychoeducation} vs \texttt{do\mvu skill\mvu building}.}
\begin{ontolist}
\item Psychoeducation explains rationale, science, theory, mechanism, or how an intervention, experience, or pattern may help or make sense.
\item Skill building introduces, orients to, teaches, and guides the person through an exercise, activity, or skill in the interaction itself.
\end{ontolist}
 
\noindent\textbf{\texttt{do\mvu action\mvu planning} vs \texttt{do\mvu skill\mvu building}.}
\begin{ontolist}
\item Action planning identifies a clear, concrete, and defined next step the person is being asked to take, including when a request or instruction is given and is not clearly tied to orienting the person to a skill or helping them perform or practice it within the interaction.
\item Skill building focuses on introducing, orienting to, and guiding in-the-moment learning, performance, rehearsal, or practice of how to do the skill within the interaction.
\end{ontolist}
 
\noindent\textbf{\texttt{do\mvu shared\mvu understanding} vs \texttt{do\mvu inquiry}.}
\begin{ontolist}
\item Shared understanding communicates or checks understanding of what the person is expressing, including through tentative formulations.
\item Inquiry seeks new information, clarification, or depth.
\item When a turn offers a tentative understanding and also seeks confirmation, clarification, or elaboration, \texttt{do\mvu shared\mvu understanding} and \texttt{do\mvu inquiry} may both apply.
\end{ontolist}
 
\subsection{Relationship and Dream/Wish Content}
 
Relationship material carries no dedicated label; it is coded by the function
the turn performs.
 
\begin{ontolist}
\item explore details or meaning $\rightarrow$ \texttt{do\mvu inquiry}
\item mirror relationship material $\rightarrow$ \texttt{do\mvu shared\mvu understanding}
\item formulate a relational pattern $\rightarrow$ \texttt{do\mvu interpret}
\item challenge a relational discrepancy $\rightarrow$ \texttt{do\mvu challenge}
\item plan a relational behavior change $\rightarrow$ \texttt{do\mvu action\mvu planning}
\end{ontolist}
 
\noindent Dream, fantasy, and wish material is coded by function under the same
rule. Labels are never created for a topic domain alone.

\section{Move Tool Definitions}
\label{app:tools}
 
In the \textit{with-moves} condition, the ontology is provided to the clinician model as a set of function-calling tools. One tool is defined per move, named exactly after its corresponding label, and the system prompt requires at least one tool call before every turn. Each tool includes a \emph{definition} that instructs the model on what the move is and when to select it. This definition comprises a name, a description condensed from the matching move card in Appendix~\ref{app:ontology}, and a set of required string parameters. A \emph{response} is returned once the tool is called, telling the model how to draft the turn it has just committed to. Figures~\ref{fig:tools-definitions-process-alignment}--\ref{fig:tool_no_defined_move} detail the definitions and responses for all eleven tools.
 
\clearpage
\begingroup

\centering
 
\begin{tcolorbox}[enhanced, breakable,
  colback=black!2, colframe=black!55, boxrule=0.5pt, arc=2pt,
  left=5pt, right=5pt, top=2pt, bottom=2pt,
  title=\textsc{Tool definition}, fonttitle=\bfseries\small,
  coltitle=white, colbacktitle=black!55]
\begin{Verbatim}[fontsize=\scriptsize, baselinestretch=0.9]
{
  "type": "function",
  "function": {
    "name": "do_process_alignment",
    "description": "Calibrate how therapy is working in real time - pacing, depth, tone,
      direction, and quality of collaboration. This includes explicit repair when the
      therapist has likely missed the client, moved too quickly, or created strain in the
      interaction. It is process-focused meta-communication, not content exploration.
 
      When to use:
      - After a potential rupture, defensiveness spike, or visible disengagement
      - When checking whether the current direction or pace feels right to the client
      - When deciding together whether to pursue exploration of internal experience vs.
        clinical intervention (e.g., problem solving, guided activity, psychoeducation)
 
      Contraindications:
      - When asking about life events, symptoms, or meaning (use do_inquiry)
      - When seeking confirmation that an interpretation is factually correct (boundary with
        do_shared_understanding/do_interpret)
      - When used repeatedly to avoid difficult but productive material",
    "parameters": {
      "type": "object",
      "properties": {
        "process_observation": {
          "type": "string",
          "description": "A brief naming of what is happening in the interaction itself
            (e.g., a pace mismatch, a missed bid, a strained moment)."
        }
      },
      "required": ["process_observation"]
    }
  }
}
\end{Verbatim}
\end{tcolorbox}
 
\vspace{2pt}
 
\begin{tcolorbox}[enhanced, breakable,
  colback=black!2, colframe=black!55, boxrule=0.5pt, arc=2pt,
  left=5pt, right=5pt, top=2pt, bottom=2pt,
  title=\textsc{Tool response}, fonttitle=\bfseries\small,
  coltitle=white, colbacktitle=black!55]
\begin{Verbatim}[fontsize=\scriptsize, baselinestretch=0.9]
{
  "move": "process_alignment",
  "guidance": "Briefly name the process observation. If repair is needed, acknowledge and
    take responsibility for the therapist's role in the rupture. If helpful, ask a
    clarifying question to define the source of the rupture. Where possible, offer a
    concrete adjustment. Do one of these per turn - spread across turns, not all at once.
    Stay focused on process, not content. Maximum two sentences. Do not verbatim-quote the
    member."
}
\end{Verbatim}
\end{tcolorbox}
 
\captionof{figure}{The \texttt{do\_process\_alignment} tool: definition and returned response.}
\label{fig:tools-definitions-process-alignment}
\par
\endgroup

\clearpage
\begingroup

\centering
 
\begin{tcolorbox}[enhanced, breakable,
  colback=black!2, colframe=black!55, boxrule=0.5pt, arc=2pt,
  left=5pt, right=5pt, top=2pt, bottom=2pt,
  title=\textsc{Tool definition}, fonttitle=\bfseries\small,
  coltitle=white, colbacktitle=black!55]
\begin{Verbatim}[fontsize=\scriptsize, baselinestretch=0.9]
{
  "type": "function",
  "function": {
    "name": "do_goal_setting",
    "description": "Collaboratively identify, clarify, or prioritize what the member wants
      to work toward. This includes selecting a focus, choosing among competing issues, and
      helping define broad aims in a more concrete and workable way.
 
      When to use:
      - At session/interaction start, transition points, or time-limited moments
      - When multiple issues compete and prioritization is needed
      - When converting broad areas for change into concrete targets
 
      Contraindications:
      - When therapist is already specifying action steps (use do_action_planning)
      - When the turn only gathers details (use do_inquiry)",
    "parameters": {
      "type": "object",
      "properties": {
        "goal": {
          "type": "string",
          "description": "A concise statement of the therapeutic goal or goals being set or
            clarified/proposed with the patient."
        }
      },
      "required": ["goal"]
    }
  }
}
\end{Verbatim}
\end{tcolorbox}
 
\vspace{2pt}
 
\begin{tcolorbox}[enhanced, breakable,
  colback=black!2, colframe=black!55, boxrule=0.5pt, arc=2pt,
  left=5pt, right=5pt, top=2pt, bottom=2pt,
  title=\textsc{Tool response}, fonttitle=\bfseries\small,
  coltitle=white, colbacktitle=black!55]
\begin{Verbatim}[fontsize=\scriptsize, baselinestretch=0.9]
{
  "move": "goal_setting",
  "guidance": "Propose a concrete, collaborative goal grounded in the patient's needs,
    invite them to refine it, and don't slip into planning specific steps. Maximum two
    sentences. One thing per turn. Do not verbatim-quote the member."
}
\end{Verbatim}
\end{tcolorbox}
 
\captionof{figure}{The \texttt{do\_goal\_setting} tool: definition and returned response.}
\label{fig:tool_goal_setting}
\par
\endgroup

\clearpage
\begingroup

\centering
 
\begin{tcolorbox}[enhanced, breakable,
  colback=black!2, colframe=black!55, boxrule=0.5pt, arc=2pt,
  left=5pt, right=5pt, top=2pt, bottom=2pt,
  title=\textsc{Tool definition}, fonttitle=\bfseries\small,
  coltitle=white, colbacktitle=black!55]
\begin{Verbatim}[fontsize=\scriptsize, baselinestretch=0.9]
{
  "type": "function",
  "function": {
    "name": "do_inquiry",
    "description": "Ask focused questions to deepen understanding of meaning, sequence,
      context, or concrete details - for the member and/or the therapist. Includes both
      exploratory inquiry (values, emotions, meaning) and clarifying inquiry (when,
      frequency, triggers, sequence, body cues). Does not need to take the form of a
      question; an instruction that serves the same function counts. The function is
      understanding - not persuasion, confrontation, or another move dressed in question
      form.
 
      When to use:
      - To map what happened before/during/after key moments
      - To disambiguate thought vs. feeling vs. behavior vs. sensation
      - To deepen personal significance of events, relationship moments, or wishes
      - To build case conceptualization and hypotheses about the development and maintenance
        of symptoms and functional challenges
 
      Contraindications:
      - When the question primarily tests discrepancy or challenges the member's perspective
        (use do_challenge)
      - When the question primarily repairs process fit (use do_process_alignment)
      - When the question mainly introduces a therapist-generated hypothesis or pattern not
        yet in the member's account (use do_interpret)
      - When the question primarily explores or evokes motivation, readiness, or ambivalence
        (use do_support_change)",
    "parameters": {
      "type": "object",
      "properties": {
        "question": {
          "type": "string",
          "description": "The focused question or instruction-as-question that aims to
            deepen understanding."
        },
        "inquiry_focus": {
          "type": "string",
          "description": "What the question is targeting (e.g., a specific moment, a body
            sensation, the meaning of an event, a sequence of events, an emotional
            experience)."
        }
      },
      "required": ["question", "inquiry_focus"]
    }
  }
}
\end{Verbatim}
\end{tcolorbox}
 
\vspace{2pt}
 
\begin{tcolorbox}[enhanced, breakable,
  colback=black!2, colframe=black!55, boxrule=0.5pt, arc=2pt,
  left=5pt, right=5pt, top=2pt, bottom=2pt,
  title=\textsc{Tool response}, fonttitle=\bfseries\small,
  coltitle=white, colbacktitle=black!55]
\begin{Verbatim}[fontsize=\scriptsize, baselinestretch=0.9]
{
  "move": "inquiry",
  "guidance": "Ask one high-yield question at a time. Anchor to a specific moment before
    broad generalization. Prefer concrete wording over abstract prompts. Avoid rapid
    question stacking. Maximum two sentences. One thing per turn. Do not verbatim-quote the
    member."
}
\end{Verbatim}
\end{tcolorbox}
 
\captionof{figure}{The \texttt{do\_inquiry} tool: definition and returned response.}
\label{fig:tool_inquiry}
\par
\endgroup

\clearpage
\begingroup

\centering
 
\begin{tcolorbox}[enhanced, breakable,
  colback=black!2, colframe=black!55, boxrule=0.5pt, arc=2pt,
  left=5pt, right=5pt, top=2pt, bottom=2pt,
  title=\textsc{Tool definition}, fonttitle=\bfseries\small,
  coltitle=white, colbacktitle=black!55]
\begin{Verbatim}[fontsize=\scriptsize, baselinestretch=0.9]
{
  "type": "function",
  "function": {
    "name": "do_shared_understanding",
    "description": "Communicate an understanding of what the person shared in a way that
      helps them feel heard, understood, and accurately followed - at times clarifying what
      the experience is or is not. This can include reflection, paraphrase, concise
      synthesis across multiple points, validation, normalization, and active-listening
      signals, as well as recognizing shifts in emotion or emphasis and showing continuity
      with what has already been said. May be paired with do_inquiry when seeking
      confirmation or elaboration.
 
      When to use:
      - After meaningful disclosure, to demonstrate listening and build shared understanding
      - To slow pace and help important material land
      - When the emotional experience needs to be acknowledged or validated before moving
        forward
      - When normalization may help reduce isolation or shame
      - Before moving to interpretation, challenge, planning, or change-oriented guidance
 
      Contraindications:
      - When introducing hidden meaning, causal explanation, or inferred mechanisms not
        already grounded in what was shared (use do_interpret)
      - When your response adds ANY meaning, pattern, or framework beyond what the person
        explicitly stated - even if it feels like a natural extension or a small inference
        (use do_interpret)
      - When explaining how something works in general - a mechanism, a concept, a rationale
        (use do_psychoeducation)
      - When explicitly trying to reinforce effort, build motivation, or evoke change talk
        (use do_support_change)
      - When the response is intended to highlight discrepancy, tension, or inconsistency
        (use do_challenge)
      - When normalization would minimize, flatten, or prematurely reassure rather than help
        the person feel understood
 
      Key test: Could the person have said this about themselves? If yes, it may be
      shared_understanding. If you are adding something they did not say or see yet, use
      do_interpret.",
    "parameters": {
      "type": "object",
      "properties": {
        "reflection": {
          "type": "string",
          "description": "A brief summary of what the patient expressed that you are
            reflecting back."
        }
      },
      "required": ["reflection"]
    }
  }
}
\end{Verbatim}
\end{tcolorbox}
 
\vspace{2pt}
 
\begin{tcolorbox}[enhanced, breakable,
  colback=black!2, colframe=black!55, boxrule=0.5pt, arc=2pt,
  left=5pt, right=5pt, top=2pt, bottom=2pt,
  title=\textsc{Tool response}, fonttitle=\bfseries\small,
  coltitle=white, colbacktitle=black!55]
\begin{Verbatim}[fontsize=\scriptsize, baselinestretch=0.9]
{
  "move": "shared_understanding",
  "guidance": "Use plain language grounded in the person's expressed experience - always
    paraphrase, never quote their words back. Match the form to what is most needed: brief
    acknowledgments to keep them sharing; fuller reflections or synthesis to consolidate;
    validation when the emotional experience needs explicit recognition; normalization when
    it would reduce isolation or shame without minimizing. IMPORTANT: Stay strictly within
    what the person has expressed. If your response adds any meaning, pattern, hypothesis,
    or explanation beyond what they explicitly said - even a small one - STOP and use
    do_interpret or do_psychoeducation instead. Maximum two sentences. One thing per turn.
    Do not verbatim-quote the member."
}
\end{Verbatim}
\end{tcolorbox}
 
\captionof{figure}{The \texttt{do\_shared\_understanding} tool: definition and returned response.}
\label{fig:tool_shared_understanding}
\par
\endgroup

\clearpage
\begingroup

\centering
 
\begin{tcolorbox}[enhanced, breakable,
  colback=black!2, colframe=black!55, boxrule=0.5pt, arc=2pt,
  left=5pt, right=5pt, top=2pt, bottom=2pt,
  title=\textsc{Tool definition}, fonttitle=\bfseries\small,
  coltitle=white, colbacktitle=black!55]
\begin{Verbatim}[fontsize=\scriptsize, baselinestretch=0.9]
{
  "type": "function",
  "function": {
    "name": "do_support_change",
    "description": "Strengthen the person's readiness, willingness, and confidence to move
      toward change. This includes reinforcing effort or movement already shown,
      highlighting agency and choice, exploring the person's own reasons and values for
      change, and instilling realistic hope that change is possible. The aim is to support
      movement toward change by strengthening or exploring motivation, readiness,
      ambivalence, confidence, and personal meaning - without persuading, pressuring, or
      prematurely moving into action design.
 
      When to use:
      - When ambivalence is present or commitment is not yet clear
      - When exploring whether change feels possible, desirable, or worthwhile
      - When reflecting on readiness, ambivalence, or what change would mean
      - When exploring the person's own reasons, values, hopes, or perceived benefits
      - When effort, progress, or values-consistent movement can be reinforced
      - Before action planning, when motivation, readiness, or confidence still needs
        strengthening
 
      Contraindications:
      - When the main need is understanding, reflection, validation, or normalization (use
        do_shared_understanding)
      - When the main function is concrete step design or implementation planning (use
        do_action_planning)
      - When the main function is gathering information or deepening understanding (use
        do_inquiry)
      - When the main function is directly surfacing discrepancy, tension, or a limiting
        perspective (use do_challenge)",
    "parameters": {
      "type": "object",
      "properties": {
        "change_target": {
          "type": "string",
          "description": "The change, effort, or movement being supported, reinforced, or
            explored."
        }
      },
      "required": ["change_target"]
    }
  }
}
\end{Verbatim}
\end{tcolorbox}
 
\vspace{2pt}
 
\begin{tcolorbox}[enhanced, breakable,
  colback=black!2, colframe=black!55, boxrule=0.5pt, arc=2pt,
  left=5pt, right=5pt, top=2pt, bottom=2pt,
  title=\textsc{Tool response}, fonttitle=\bfseries\small,
  coltitle=white, colbacktitle=black!55]
\begin{Verbatim}[fontsize=\scriptsize, baselinestretch=0.9]
{
  "move": "support_change",
  "guidance": "Help them move toward change without pushing. If they're overwhelmed, steady
    them first. If they're unsure, ask open questions about what change would mean. If
    they've already made an effort, name the specific thing they did. Keep the choice
    theirs. Validate feelings, not conclusions. Don't jump to planning steps - that's
    do_action_planning. Maximum two sentences. One thing per turn. Do not verbatim-quote the
    member."
}
\end{Verbatim}
\end{tcolorbox}
 
\captionof{figure}{The \texttt{do\_support\_change} tool: definition and returned response.}
\label{fig:tool_support_change}
\par
\endgroup

\clearpage
\begingroup

\centering
 
\begin{tcolorbox}[enhanced, breakable,
  colback=black!2, colframe=black!55, boxrule=0.5pt, arc=2pt,
  left=5pt, right=5pt, top=2pt, bottom=2pt,
  title=\textsc{Tool definition}, fonttitle=\bfseries\small,
  coltitle=white, colbacktitle=black!55]
\begin{Verbatim}[fontsize=\scriptsize, baselinestretch=0.9]
{
  "type": "function",
  "function": {
    "name": "do_interpret",
    "description": "Offer a therapist-generated hypothesis, opinion, or framework about the
      underlying meaning, function, or pattern in what the person is experiencing. Can
      include a local hypothesis about a specific moment, a smaller-scale extension of what
      was shared, what the therapist observes, or a broader conceptualization that links
      multiple experiences into a coherent pattern. The goal is to help the person see a
      pattern, meaning, or organizing framework that is not yet fully explicit in what they
      have shared.
 
      When to use:
      - When sufficient context exists to support a plausible, evidence-based hypothesis
      - When multiple pieces of information can be integrated into a coherent pattern
      - When the person seems stuck, confused, or repetitive in a way that may benefit from
        a new organizing perspective
      - When a clinician's synthesis could deepen understanding of what may be driving,
        maintaining, or connecting the person's difficulties
 
      Contraindications:
      - When the response is primarily reflecting, paraphrasing, validating, or normalizing
        content already explicit (use do_shared_understanding)
      - When the main goal is to surface discrepancy, tension, or a limiting perspective
        (use do_challenge)
      - When the response is primarily educational and not person-specific (use
        do_psychoeducation)
      - When there is too little context to support a plausible interpretation
      - When the interpretation would move too far beyond available data, imply diagnosis,
        or present speculation as fact",
    "parameters": {
      "type": "object",
      "properties": {
        "hypothesis": {
          "type": "string",
          "description": "The therapist-generated meaning, function, pattern, or formulation
            being offered."
        },
        "grounding": {
          "type": "string",
          "description": "The specific material from the person's account (themes,
            observations, repeated patterns, examples) that the hypothesis is anchored in."
        }
      },
      "required": ["hypothesis", "grounding"]
    }
  }
}
\end{Verbatim}
\end{tcolorbox}
 
\vspace{2pt}
 
\begin{tcolorbox}[enhanced, breakable,
  colback=black!2, colframe=black!55, boxrule=0.5pt, arc=2pt,
  left=5pt, right=5pt, top=2pt, bottom=2pt,
  title=\textsc{Tool response}, fonttitle=\bfseries\small,
  coltitle=white, colbacktitle=black!55]
\begin{Verbatim}[fontsize=\scriptsize, baselinestretch=0.9]
{
  "move": "interpret",
  "guidance": "Present interpretations tentatively and collaboratively, as possible ways of
    understanding rather than as facts. Ground the interpretation in material the person has
    actually shared. Invite correction, elaboration, or non-fit explicitly. Avoid
    over-certainty, hidden leaps, or diagnosis language. Maximum two sentences. One thing
    per turn. Do not verbatim-quote the member."
}
\end{Verbatim}
\end{tcolorbox}
 
\captionof{figure}{The \texttt{do\_interpret} tool: definition and returned response.}
\label{fig:tool_interpret}
\par
\endgroup

\clearpage
\begingroup

\centering
 
\begin{tcolorbox}[enhanced, breakable,
  colback=black!2, colframe=black!55, boxrule=0.5pt, arc=2pt,
  left=5pt, right=5pt, top=2pt, bottom=2pt,
  title=\textsc{Tool definition}, fonttitle=\bfseries\small,
  coltitle=white, colbacktitle=black!55]
\begin{Verbatim}[fontsize=\scriptsize, baselinestretch=0.9]
{
  "type": "function",
  "function": {
    "name": "do_challenge",
    "description": "Surface and examine beliefs, interpretations, patterns, or coping
      responses that may be keeping the person stuck or moving them away from their goals.
      Challenge is not criticism, confrontation, or argument - it is a purposeful
      intervention used to increase awareness, flexibility, and openness to change.
 
      When to use:
      - When a belief, interpretation, behavior, or coping response appears to be
        maintaining the problem or interfering with progress
      - When there is a meaningful discrepancy between the person's goals, values,
        interpretations, emotions, or actions worth examining directly
      - When examining the accuracy, usefulness, or consequences of a thought or pattern
        could increase flexibility and improve response options
 
      Contraindications:
      - When rapport, trust, or shared understanding is too limited for challenge to land
      - When immediate priorities are safety, stabilization, containment, or basic emotional
        support
      - When the main task is clarification or exploration without a discrepancy or limiting
        perspective being directly named (use do_inquiry)
      - When the main task is communicating understanding, validation, or normalization (use
        do_shared_understanding)
      - When the main task is offering a broader explanatory hypothesis (use do_interpret)
      - When the intervention would come across as criticism, shaming, arguing, or pushing
        the person to accept the clinician's view",
    "parameters": {
      "type": "object",
      "properties": {
        "challenge": {
          "type": "string",
          "description": "What you're examining (the specific thing they said, did, or
            believe) and why it matters (e.g., 'going quiet in conflict - clashes with
            wanting closer relationships')."
        }
      },
      "required": ["challenge"]
    }
  }
}
\end{Verbatim}
\end{tcolorbox}
 
\vspace{2pt}
 
\begin{tcolorbox}[enhanced, breakable,
  colback=black!2, colframe=black!55, boxrule=0.5pt, arc=2pt,
  left=5pt, right=5pt, top=2pt, bottom=2pt,
  title=\textsc{Tool response}, fonttitle=\bfseries\small,
  coltitle=white, colbacktitle=black!55]
\begin{Verbatim}[fontsize=\scriptsize, baselinestretch=0.9]
{
  "move": "challenge",
  "guidance": "Pick one thing to challenge - a belief, a pattern, a contradiction. Point to
    something they actually said or did, and name it directly. Tie it to something they care
    about. Invite them to look at it with you, don't tell them they're wrong. If they don't
    engage, drop it. Maximum two sentences. One thing per turn. Do not verbatim-quote the
    member."
}
\end{Verbatim}
\end{tcolorbox}
 
\captionof{figure}{The \texttt{do\_challenge} tool: definition and returned response.}
\label{fig:tool_challenge}
\par
\endgroup

\clearpage
\begingroup

\centering
 
\begin{tcolorbox}[enhanced, breakable,
  colback=black!2, colframe=black!55, boxrule=0.5pt, arc=2pt,
  left=5pt, right=5pt, top=2pt, bottom=2pt,
  title=\textsc{Tool definition}, fonttitle=\bfseries\small,
  coltitle=white, colbacktitle=black!55]
\begin{Verbatim}[fontsize=\scriptsize, baselinestretch=0.9]
{
  "type": "function",
  "function": {
    "name": "do_psychoeducation",
    "description": "Explain a concept, mechanism, or rationale that helps the person
      understand their experience, symptoms, a pattern, a skill, or the treatment process.
      Psychoeducation is a targeted intervention used to support insight and understanding.
 
      When to use:
      - When a clearer framework would help them make sense of what they're experiencing
      - When explaining the why behind a skill, intervention, or pattern supports insight or
        engagement
      - When a brief explanation would clarify or reframe how they're interpreting symptoms
        or reactions
 
      Contraindications:
      - When the explanation is about this person's specific pattern or meaning (use
        do_interpret)
      - When proposing action steps (use do_action_planning)
      - When guiding them through a skill in the moment (use do_skill_building)
      - When validating or reflecting (use do_shared_understanding)
      - When they need emotional understanding, not information
      - When they're too overwhelmed to take in new info",
    "parameters": {
      "type": "object",
      "properties": {
        "concept": {
          "type": "string",
          "description": "What you're explaining and how it connects to their current
            experience or next step (e.g., 'avoidance lowers discomfort short-term but keeps
            fear going - why this skill matters for you')."
        }
      },
      "required": ["concept"]
    }
  }
}
\end{Verbatim}
\end{tcolorbox}
 
\vspace{2pt}
 
\begin{tcolorbox}[enhanced, breakable,
  colback=black!2, colframe=black!55, boxrule=0.5pt, arc=2pt,
  left=5pt, right=5pt, top=2pt, bottom=2pt,
  title=\textsc{Tool response}, fonttitle=\bfseries\small,
  coltitle=white, colbacktitle=black!55]
\begin{Verbatim}[fontsize=\scriptsize, baselinestretch=0.9]
{
  "move": "psychoeducation",
  "guidance": "One concept at a time. Plain language, no jargon. Tie it directly to what
    they're going through. Keep it short - this is a brief explanation, not a lecture. Check
    it landed before moving on. Maximum two sentences. One thing per turn. Do not
    verbatim-quote the member."
}
\end{Verbatim}
\end{tcolorbox}
 
\captionof{figure}{The \texttt{do\_psychoeducation} tool: definition and returned response.}
\label{fig:tool_psychoeducation}
\par
\endgroup

\clearpage
\begingroup

\centering
 
\begin{tcolorbox}[enhanced, breakable,
  colback=black!2, colframe=black!55, boxrule=0.5pt, arc=2pt,
  left=5pt, right=5pt, top=2pt, bottom=2pt,
  title=\textsc{Tool definition}, fonttitle=\bfseries\small,
  coltitle=white, colbacktitle=black!55]
\begin{Verbatim}[fontsize=\scriptsize, baselinestretch=0.9]
{
  "type": "function",
  "function": {
    "name": "do_action_planning",
    "description": "Turn a goal or intention into a specific next step the person will try -
      usually outside the session. Defines what they'll do, when, and where, clearly enough
      to review later.
 
      When to use:
      - When there's enough understanding and at least some readiness to try something
      - When they're asking for practical next steps
      - When a goal is set and the next task is figuring out how to start
      - When more specificity would help them follow through
 
      Contraindications:
      - When they're ambivalent and motivation needs strengthening first (use
        do_support_change)
      - When the goal itself isn't clear yet (use do_goal_setting)
      - When explaining rationale or a concept (use do_psychoeducation)
      - When teaching or practicing a skill in the moment (use do_skill_building)",
    "parameters": {
      "type": "object",
      "properties": {
        "action_step": {
          "type": "string",
          "description": "What they'll do, including when, where, and any conditions (e.g.,
            '5-minute walk after dinner, weekdays')."
        }
      },
      "required": ["action_step"]
    }
  }
}
\end{Verbatim}
\end{tcolorbox}
 
\vspace{2pt}
 
\begin{tcolorbox}[enhanced, breakable,
  colback=black!2, colframe=black!55, boxrule=0.5pt, arc=2pt,
  left=5pt, right=5pt, top=2pt, bottom=2pt,
  title=\textsc{Tool response}, fonttitle=\bfseries\small,
  coltitle=white, colbacktitle=black!55]
\begin{Verbatim}[fontsize=\scriptsize, baselinestretch=0.9]
{
  "move": "action_planning",
  "guidance": "Build the step with them, not for them. Keep it small and realistic. Get
    explicit buy-in before finalizing. Frame it as something to learn from, not pass/fail.
    Set a review point if useful. Maximum two sentences. One thing per turn. Do not
    verbatim-quote the member."
}
\end{Verbatim}
\end{tcolorbox}
 
\captionof{figure}{The \texttt{do\_action\_planning} tool: definition and returned response.}
\label{fig:tool_action_planning}
\par
\endgroup

\clearpage
\begingroup

\centering
 
\begin{tcolorbox}[enhanced, breakable,
  colback=black!2, colframe=black!55, boxrule=0.5pt, arc=2pt,
  left=5pt, right=5pt, top=2pt, bottom=2pt,
  title=\textsc{Tool definition}, fonttitle=\bfseries\small,
  coltitle=white, colbacktitle=black!55]
\begin{Verbatim}[fontsize=\scriptsize, baselinestretch=0.9]
{
  "type": "function",
  "function": {
    "name": "do_skill_building",
    "description": "Teach, walk through, or practice a specific skill with the person in the
      moment. Includes introducing it, guiding them through the steps, supporting them as
      they try it, and reflecting on how it went. The focus is on building capability
      through guided practice - not just explaining the skill or planning to use it later.
 
      This move spans multiple turns. Keep tagging do_skill_building across all those turns
      while the work is still underway.
 
      When to use:
      - When they're ready to try a skill and would benefit from guided support
      - When practicing in the moment would build understanding or confidence
      - When the skill is best learned by doing, not just hearing about it
      - When they're asking how to do something, not just whether to
 
      Contraindications:
      - When motivation isn't there yet (use do_support_change)
      - When only explaining the rationale, not practicing (use do_psychoeducation)
      - When planning to use the skill later, not now (use do_action_planning)",
    "parameters": {
      "type": "object",
      "properties": {
        "skill": {
          "type": "string",
          "description": "The specific skill being introduced or practiced (e.g., 'grounding
            exercise', 'stating a boundary')."
        },
        "stage": {
          "type": "string",
          "description": "Where this turn sits in the arc - introducing (naming the skill,
            getting buy-in), teaching_step (explaining or modeling one part), practicing
            (prompting them to try it, supporting in real time), or reflecting (checking how
            it landed, summarizing the takeaway)."
        }
      },
      "required": ["skill", "stage"]
    }
  }
}
\end{Verbatim}
\end{tcolorbox}
 
\vspace{2pt}
 
\begin{tcolorbox}[enhanced, breakable,
  colback=black!2, colframe=black!55, boxrule=0.5pt, arc=2pt,
  left=5pt, right=5pt, top=2pt, bottom=2pt,
  title=\textsc{Tool response}, fonttitle=\bfseries\small,
  coltitle=white, colbacktitle=black!55]
\begin{Verbatim}[fontsize=\scriptsize, baselinestretch=0.9]
{
  "move": "skill_building",
  "guidance": "Introduce conversationally, in plain language. One part at a time. Alternate
    explaining and practicing so they learn by doing. Stay responsive to their pace. Keep it
    feeling like a conversation, not a lesson. End by naming the takeaway and when they
    might use the skill again. Maximum two sentences. One thing per turn. Do not
    verbatim-quote the member."
}
\end{Verbatim}
\end{tcolorbox}
 
\captionof{figure}{The \texttt{do\_skill\_building} tool: definition and returned response.}
\label{fig:tool_skill_building}
\par
\endgroup

\clearpage
\begingroup

\centering
 
\begin{tcolorbox}[enhanced, breakable,
  colback=black!2, colframe=black!55, boxrule=0.5pt, arc=2pt,
  left=5pt, right=5pt, top=2pt, bottom=2pt,
  title=\textsc{Tool definition}, fonttitle=\bfseries\small,
  coltitle=white, colbacktitle=black!55]
\begin{Verbatim}[fontsize=\scriptsize, baselinestretch=0.9]
{
  "type": "function",
  "function": {
    "name": "no_defined_move",
    "description": "Use this when the turn doesn't need to do therapeutic work - it's just
      keeping the conversation running. Greetings, logistics, scheduling, quick
      acknowledgments that aren't really listening, technical check-ins.
 
      When to use:
      - When handling scheduling, logistics, or technical setup
      - When greeting or closing without it carrying clinical weight
      - When making a procedural statement
      - When giving a brief backchannel that isn't doing active listening
      - When the turn keeps the flow going but isn't advancing a therapeutic task
 
      Contraindications:
      - When the turn would do meaningful therapeutic work, even briefly - pick that move
        instead
      - When active listening, reflection, validation, or normalization fits (use
        do_shared_understanding)
      - When something minimal on the surface is actually doing clinical work in context",
    "parameters": {
      "type": "object",
      "properties": {},
      "required": []
    }
  }
}
\end{Verbatim}
\end{tcolorbox}
 
\vspace{2pt}
 
\begin{tcolorbox}[enhanced, breakable,
  colback=black!2, colframe=black!55, boxrule=0.5pt, arc=2pt,
  left=5pt, right=5pt, top=2pt, bottom=2pt,
  title=\textsc{Tool response}, fonttitle=\bfseries\small,
  coltitle=white, colbacktitle=black!55]
\begin{Verbatim}[fontsize=\scriptsize, baselinestretch=0.9]
{
  "move": "no_defined_move",
  "guidance": "Answer the patient. Maximum two sentences. One thing per turn. Do not
    verbatim-quote the member."
}
\end{Verbatim}
\end{tcolorbox}
 
\captionof{figure}{The \texttt{no\_defined\_move} tool: definition and returned response.}
\label{fig:tool_no_defined_move}
\par
\endgroup

\end{document}